# Objective, Probabilistic, and Generalized Noise Level Dependent Classifications of sets of more or less 2D Periodic Images into Plane Symmetry Groups

*Andrew Dempsey and Peter Moeck**

Nano-Crystallography Group, Department of Physics, Portland State University, Portland, OR, 97207-0751

Email: pmoeck@pdx.edu*

**Abstract:** Crystallographic symmetry classifications from real-world images with periodicities in two dimensions (2D) are of interest to crystallographers and practitioners of computer vision studies alike. Currently, these classifications are typically made by both communities in a subjective manner that relies heavily on arbitrary thresholds for judgments, and are reported under the pretense of being definitive, which is impossible. Moreover, the computer vision community tends to use direct space methods to make such classifications instead of more powerful and computationally efficient Fourier/reciprocal space (spatial frequency domain) methods, such as those routinely used as part of crystallographic image processing. This is due to the fact that the proper functioning of those methods requires more periodic repeats of a unit cell motif than are commonly present in images analyzed by the computer vision community. We demonstrate a novel, information-theoretic approach to plane symmetry group classifications that is enabled by Kenichi Kanatani's Geometric Akaike Information Criterion and associated Geometric Akaike weights. Our approach leverages the computational efficiency of working in Fourier/reciprocal space, is well suited for handling the hierarchic nature of crystallographic symmetries, and yields probabilistic results that are generalized noise level dependent. The latter feature means crystallographic symmetry classifications can be "updated" when better, i.e. less noisy, image data and more accurate processing algorithms become available. We demonstrate the ability of our approach to objectively estimate the underlying plane symmetry and pseudosymmetries of sets of synthetic 2D-periodic images with varying amounts of "red-green-blue" (converted to gray levels) and "spread" noise. Additionally, we suggest a simple solution to the problem of too few periodic repeats in an input image for practical application of Fourier space methods. In doing so, we effectively solve the decades-old and heretofore intractable problem from computer vision of symmetry detection and classification from images in the presence of noise.

## 1. Introduction

The development of algorithms and practical procedures to automate the detection and classification of various types of crystallographic symmetries present in real-world, more-or-less 2D-periodic, digital images is of mutual interest to a variety of research communities, including those of computer vision [1, 2] and computational symmetry [3, 4], tile and textile manufacturers [5, 6], scholars of Islamic building ornamentation [7-9], and applied crystallographers [10-25]. It stands to reason that open communication and collaboration between these communities will benefit all of them, and a goal of this paper is to encourage such collaborative efforts.

The computer vision/computational symmetry community has been developing symmetry detection algorithms for several decades [1]. Their efforts have yielded a variety of algorithms to detect and classify point symmetries (i.e. rotations and reflections), translation symmetries in one and two dimensions, as well as point symmetries combined with one- and two-dimensional (2D) translations (i.e. frieze and plane/wallpaper symmetries). A thorough review of the state of the art in the computer vision/computational symmetry community up to 2009 can be found in [4].

Generally speaking, the computer vision community tends to produce algorithms that work in direct/physical space (as opposed to reciprocal/Fourier space) with non-transformed input images. It is common practice for the computational symmetry community to conduct performance evaluations of [1] and competitions between [2] different symmetry detection and classification algorithms using data sets with hand-labeled "ground truth" [2] with the goal of ranking their efficacy.

That community's progress is to be lauded, yet there is more work to be done. The abstract of [1] includes a sobering proclamation: *"even after several decades of effort, symmetry detection in real-world images remains a challenging, unsolved problem in computer vision."* A similar statement was made by Yanxi Liu and colleagues in the abstract of [4]: "*Computational symmetry on real world data turns out to be challenging enough that, after decades of effort, a fully automated symmetry–savvy system remains elusive for real world applications*."

While the notion that competitions and performance evaluations such as those mentioned above will eventually produce the best possible symmetry detection algorithm(s) is not entirely misguided in that it stimulates innovation, it must be kept in mind that obtaining *definitive* results from a real symmetry detection algorithm working with real world data is impossible. This observation appeared in print in Kenichi Kanatani's most generally accessible paper in 2004 [26], and while perhaps unsettling to some computer scientists who may believe a perfect algorithm with unlimited accuracy and precision is attainable, it is nonetheless true.

The reason why obtaining definitive results is impossible is that uncertainty is inherent in all image processing tasks. This uncertainty exists independently of, and in addition to, image recording and processing noise as well as any deviations from the idealized structure of the original pattern of which the digital image is taken. No two geometric feature extraction algorithms working with the same input image (or set of images) will produce exactly the same results, as demonstrated extensively by the results of the above-mentioned competitions [1, 2] and in [20]. This is because the manner in which physical properties are translated into image properties is intrinsically heuristic, and all algorithms involved in these competitions rely on internal, arbitrarily set thresholds [26]. As such, the results obtained from any algorithm require statistical treatment, but a traditional statistical approach taken in the limit of infinite observations fails because, for image processing, the number of observations is typically one.

The applied crystallography community is interested in detecting and classifying the full range of crystallographic symmetries present in a more-or-less 2D periodic image and its 2D discrete Fourier transform (DFT). This includes the underlying Bravais lattice type, plane symmetry group, and the 2D point group of the amplitude map of the Fourier transformed image, i.e. the Laue class. In practice, this is accomplished by an integral part of crystallographic image processing (CIP), a Fourier/reciprocal space averaging technique originated by Chemistry Nobel Prize Laureate Sir Aaron Klug (1982) and his various coworkers (including the 2017 Chemistry Nobel Prize winner Richard Henderson) at the MRC Laboratory of Molecular Biology at Cambridge, UK, over 50 years ago [25]. Originally created to improve the signal-to-noise ratio of images of biological macromolecules (that are often arranged more or less regularly in 2D in organic membranes) taken by transmission electron microscopy (TEM), CIP has also proven effective for the analysis of inorganic and organic crystalline arrays imaged by high resolution phase contrast TEM [10, 11], as well as scanning probe and scanning-tunneling microscopy (SPM and STM, respectively) [15, 17, 19, 24].

The ultimate goal of CIP is the extraction of the projected atomic or molecular coordinates within a "characteristic" unit cell that has been averaged over the asymmetric unit; i.e., the minimum amount of information needed to reproduce a pattern by 2D translations and site symmetry operations. This is a key advantage of CIP over traditional Fourier filtering, which averages only over the translation-periodic unit cell. Since up to twelve times more information is being averaged over in the case of CIP, it is much more effective at suppressing noise than Fourier filtering. This, in turn, means CIP facilitates more accurate extractions of atomic or molecular coordinates.

A variety of programs have been developed within the applied crystallography community that implement CIP, all of which work in reciprocal space. One of the most widely used CIP programs is CRISP [12], the popularity of which is thanks in part to the fact that it runs on Windows PCs. Reciprocal space methods have the advantage of condensing all structural information spread out over hundreds of thousands to millions of pixels in the direct space image into a few tens to hundreds of "structure-bearing" complex Fourier coefficients (FCs), which results in a drastic improvement in computational efficiency. Despite this, the computational symmetry community tends to avoid reciprocal space methods due to their requirement of sufficiently large numbers of periodic repeats in an input image to function properly. A review of the state of the art in crystallographic symmetry classifications of real-world images up to 2019 can be found in [21, 22]

In the context of crystallographic symmetry classifications from real-world images, difficulties arise from the fact that symmetries are hierarchic in nature. Hierarchies exist among the full range of crystallographic symmetries, including the five 2D Bravais lattice types, six 2D Laue Classes, and the seventeen 2D space groups (henceforth plane symmetry groups). Symmetry hierarchies render classifications on the basis of traditional residuals (such as those provided by CRISP [1]) as less meaningful as they could possibly be when Kanatani's geometric, information-theoretic approach is embraced [13-15, 21, 22]. This is because any pure distance measure (i.e. traditional residual) will always be smallest for the most general type/class/group in a given symmetry hierarchy branch [28].

This fundamental problem was pointed out (and a solution given) in comments by Kanatani [27] on the earlier work of Zabrodsky and colleagues [32], who classified the most-likely point symmetries of occluded/distorted geometric figures in 2D by defining a *symmetry distance* to be minimized by the method of least squares. Just over two decades later, Kanatani's comments seem to have gone mostly unnoticed by members of the communities to which this paper is addressed, as it remains common practice in both computer vision/computational symmetry and applied crystallography to classify crystallographic symmetries present in an image on the basis of some traditional residual, without embracing information theory, and without giving due consideration to pseudosymmetries. As Zabrodsky et al. introduced the concept of 2D point symmetry as continuous feature (without noticing the symmetry group inclusion/hierarchy problem), our research group generalized the continuous feature idea to Bravais lattice types [15, 23], Laue classes [13], and plane symmetry groups [13], by embracing Kanatani's geometric form of information-theoretic statistics.

Further complications to crystallographic symmetry classifications arise from the presence of Fedorov-type pseudosymmetry [21, 22, 34], motif-based pseudosymmetry, translational pseudosymmetry, and/or metric specialization [20, 35] in addition to generalized noise. As Fedorov-type pseudosymmetry has been discussed elsewhere [13], this paper deals with strong motif-based and translational pseudosymmetries. All of the aforementioned problems lead to the use of subjective "rules of thumb" for interpreting residuals, e.g., the documentation for CRISP states *"the highest possible symmetry should be chosen if you are in doubt"* [36]. Such rules are not rigorously defined, not scientific, and fall far short of a proper statistical treatment.

The solution proposed by Kanatani in [27], Geometric Akaike Information Criterion (G-AIC), is a novel statistical framework for geometric inference from images based on the logic of the traditional, information theory-based, Akaike Information Criterion (AIC). In the context of this study, the information we are interested in extracting is the 2D periodic structure contained in the pixels of a noisy, more-or-less 2D periodic, digital image.

A key difference between traditional AIC and G-AIC is the use of vanishing generalized noise as the asymptotic limit for analysis in the latter, instead of infinite observations as in traditional statistics. G-AICs are essentially geometric bias-corrected sums of squared residuals (SSRs) that are well-suited for geometric model selection in cases where the underlying model classes are frequently non-disjoint, such as crystallographic symmetry classifications. For a given set of possible geometric models, the model with the smallest G-AIC is taken to be the best. This classification as "best" is based on the minimization of the Kullback-Leibler (K-L) divergence [37]. In addition, the K-L-best model will have the largest geometric Akaike weight (G-AW) within any selected model set. Geometric Akaike weights (G-AWs) are individual model probabilities that sum to unity for a given set of geometric models [13]. They can also be interpreted as posterior Bayesian model probabilities [37].

Because the traditional residuals commonly used in CIP software are not SSRs, they are not suitable for calculating G-AICs. In this work, we introduce new SSRs for both full complex FCs and FC amplitudes, which can be used to calculate G-AICs for the 16 higher symmetric plane groups and the 5 higher symmetric Laue classes, respectively. We also present results from a MATLAB program written to calculate our new residuals for sets of example images, as well as an objective and probabilistic method for interpreting these residuals on the basis of a G-AIC and associated G-AWs. Positional SSRs in Fourier space and a corresponding G-AIC implementation for Bravais lattice type classifications have been presented elsewhere [19, 23], and will not be discussed further here.

The primary aim of this paper is two-fold: (*I*) to inform members of the computational symmetry and computer vision communities of the advantages of making crystallographic symmetry classifications in Fourier space, and (*II*) to inform members of the applied crystallography community of recent improvements with respect to crystallographic symmetry classifications that make them objective [13-15]. In the following, we will distinguish between traditional CIP as exemplified by the work of Xiaodong Zou and Sven Hovmöller [10-12] and objective CIP as developed by the authors alongside other members of Portland State University's Nano-Crystallography Group. A secondary aim of this work is to bring to the attention of both communities the problem of genuine pseudosymmetry in the presence of noise and to present our method of objective crystallographic symmetry classifications as a solution to be used in support of a future (non-traditional) CIP.

This paper is organized as follows. In section 2, we discuss in greater detail Kanatani's dictum that no image processing algorithm can yield definitive results; specifically, we discuss how it relates to crystallographic symmetry detection from more-or-less 2D periodic images. Section 3 elaborates upon the above-mentioned obstacles to symmetry detection/classification from images, i.e. symmetry hierarchies, ordinary motif-based and translational pseudosymmetry, and metric specialization. Section 4 contains a detailed discussion of Kanatani's G-AIC concept as applied to the classification of more-or-less 2D periodic images into plane symmetry groups and Laue classes. Section 4 also contains a derivation of an inequality useful for pairwise comparisons of non-disjoint geometric models (e.g. plane symmetry groups or Laue classes that are within minimal supergroup/class and maximal subgroup/class relations) on the same data, as well as ad-hoc defined confidence level for such classifications. Some background on Fourier space methods, with particular focus on plane symmetry group determination in CIP on the basis of up to three traditional figures of merit as implemented in CRISP, is given in section 5. In section 6, we present new SSRs for use in plane symmetry group and Laue class selections by G-AIC, and in section 7 we give results from our MATLAB implementation of these SSRs/G-AICs/G-AWs for plane symmetry groups on sets of test images with systematically-varied amounts of added RGB and pixel spread noise. A discussion of our results is given in section 8, followed by a Summary and Conclusions in the final section.

## 2. Kanatani's Dictum and G-AIC

Symmetry classifications that do not incorporate an information-theoretic approach are often carried out on the basis of visual inspection, or the results of a decision tree based algorithm, which answers a series of dichotomous "yes or no" questions to determine whether or not a given symmetry is present in an input image, as reviewed in [21, 22]. These types of classifications are qualitative in nature, and are reported without any statement of uncertainty which implies, intentionally or otherwise, they are to be considered as definitive. Such a practice should give pause to natural scientists, who are instead interested in probabilistic (i.e. quantitative) classifications with a corresponding measure of uncertainty. Moreover, it completely ignores the following statement made by Kanatani in [26]: "*The reason why there exist so many feature extraction algorithms, none of them being definitive, is that they are aiming at an intrinsically impossible task.*"

It is impossible to obtain definitive results from any geometric feature extraction algorithm because traditional, non-information-theoretic algorithms rely heavily on heuristics and internal, arbitrarily set thresholds to aid judgments. Moreover, all real-world imaging instruments introduce some amount of noise into the image, and image processing algorithms introduce (typically very small) systematic errors.

It must also be observed that crystallographic symmetries are strictly defined, abstract, mathematical entities. The imaged objects are real and as such they are finite and imperfect. Any symmetries they apparently possess can be described by crystallographic group theory only in an approximate sense. For example, real crystals contain defects such as random substitutions of atoms and 2D-periodic Islamic building ornamentation sometimes contains defects placed intentionally by the artisans [8].

In our approach, any deviation from the strictly defined symmetry, regardless of source, contributes to a single term called "generalized noise." The unavoidable presence of generalized noise means that in reality only non-genuine pseudosymmetries are ever observed. An elaboration on the distinction between genuine and non-genuine pseudosymmetries can be found in [20]. As such, an educated guess has to be made as to which strict symmetry would be present in the absence of generalized noise, and the outputs of different algorithms may allow for a different guess [21, 22]. This raises questions as to how the different outputs are to be qualified with respect to one another, and how to make classifications that are generalized noise level dependent.

Kanatani's solution (G-AIC) is an information-theoretic approach based on the same underlying principle as the Akaike Information Criterion of traditional statistics, i.e. the minimization of K-L divergence [37]. Whereas traditional AIC examines the asymptotic behavior of the K-L divergence as the number of observations approaches infinity, such an analysis fails for geometric inference from images because, for a given image (or set of images), the number of observations will always be one [26-31].

Kanatani performed a similar analysis to Akaike, examining instead the asymptotic behavior of the K-L divergence in the limit of vanishing noise, thereby yielding the G-AIC for a geometric model $S$, with $L$ constraints, and $N$ as number of degrees of freedom of the data [29, 30]:

$$G\text{-}AIC(S) = \hat{J} + 2(dN + n)\hat{\epsilon}^2 + O(\hat{\epsilon}^4) + ... \qquad (1),$$

where $\hat{J}$ is a standard sum of squared residuals (SSR) for $S$, $d = N-L$ is the dimension of $S$, $n$ is the number of degrees of freedom of $S$, and $\hat{\epsilon}^2$ is a squared generalized noise term, which obeys a Gaussian distribution to a sufficient approximation. The $O(\hat{\epsilon}^4)$ term in (1) represents unspecified terms that are second order in $\hat{\epsilon}^2$, while the ellipsis indicates higher-order terms that become progressively smaller.

If the (approximately Gaussian-distributed) squared generalized noise is small, second- and higher-order terms in the G-AIC are vanishingly small and can be discarded. The result is a first order (in terms of squared generalized noise), bias-corrected maximal likelihood measure that is to be minimized between the (raw/original) geometric image data and a set of judiciously chosen geometric models for that image data:

$$G\text{-}AIC(S) = \hat{J} + 2(dN + n)\hat{\epsilon}^2 \qquad (2).$$

The standard SSR $\hat{J}$ for model $S$ comprising the first term of equation (2) is balanced by a model-specific geometric bias correction term that accounts for the complexity of this geometric model. The presence of this balancing term allows for unbiased comparisons of multiple models with respect to their predictive power to explain the same image data.

In connection with objective crystallographic image classifications into plane symmetry groups, the noise term in (2) refers to pixel intensity noise per asymmetric unit, whereby that unit is, for primitive plane symmetry groups, up to 12 times smaller than the translation-periodic unit cell of the 2D periodic array itself. Non-structure bearing noise is automatically suppressed just by filtering in Fourier space for the structure bearing Fourier Coefficients (FCs), and using only these coefficients to represent the periodic signal-filtered part of a 2D periodic image.

Noise suppression by CIP based on the K-L-best model of the image data is even more effective than traditional Fourier filtering, as will become clear in sections 7 and 8. In experimental work, the easiest way to prepare for crystallographic symmetry classifications and good CIP results is to record many repeats of the unit cells at a sufficiently high spatial resolution. Due to the inbuilt noise suppression in Fourier space, the "small noise" requirement of equation (2) can often be considered to be fulfilled in practical work. Note that G-AIC based methods are bound to fail when the small and approximately Gaussian distributed generalized noise preconditions are no longer fulfilled. This is indeed observed for one of our 56 test cases in section 7.

For two geometric models $S_l$ and $S_m$, assuming $L_l < L_m$, the more constrained model $S_m$ is preferred over the less constrained model $S_l$ if

$$G\text{-}AIC(S_m) < G\text{-}AIC(S_l) \qquad (3).$$

A straightforward algebraic manipulation yields the following equivalent inequality:

$$\frac{\hat{J}_m}{\hat{J}_l} < 1 + \frac{[2(d_l - d_m)N + 2(n_l - n_m)]}{\hat{J}_l}\hat{\epsilon}^2 \qquad (4).$$

In [29-31], Kanatani shows that

$$\hat{\epsilon}^2 = \frac{\hat{J}_l}{r_l N - n_l} \qquad (5),$$

is an unbiased estimator for squared generalized noise, where $r_l$ is the co-dimension of $S_l$. Using equation (5) in inequality (4) yields

$$\frac{\hat{J}_m}{\hat{J}_l} < 1 + \frac{[2(d_l - d_m)N + 2(n_l - n_m)]}{r_l N - n_l} \qquad (6).$$

Inequality (6) allows for pairwise comparisons of non-disjoint geometric models. In the case of geometric models for different plane symmetry groups, pairs of models that are in minimal supergroup and maximal subgroup relationships and "translationengleich" [35], are entered into equation (6). As 2D Laue classes are centrosymmetric point groups, the concept of translationengleich does not apply to them. Note that a minimal translationengleiche supergroup of a given plane symmetry group arises from the addition of a single site/point symmetry operation that is not already present in the corresponding maximal subgroup. Analogously, a minimal superclass of a given Laue class arises from the addition of a point symmetry operation that is not already present in the corresponding maximal subclass.

On the other hand, adding an additional translation vector with components [½ ½] to a primitive plane symmetry group is not permitted as this violates the translationengleich restriction, i.e., that the infinite-order set of translations is preserved between maximal subgroups and minimal supergroups. Such an addition leads to a centered minimal plane symmetry supergroup and is referred to as a centering. Minimal plane symmetry supergroups that arise from a centering are known as "klassengleich" with respect to their maximal subgroups. In the context of more-or-less 2D periodic images and plane symmetry groups, klassengleich also refers to images that possess the same Laue class.

The absence of a generalized noise term in inequality (6) implies that there is no need to determine the generalized noise level independently when the pair of symmetry models being compared have minimal supergroup/class and maximal subgroup/class relationships. An analogous feature is absent from traditional statistical analysis, where noise must always be accounted for. On the other hand, if the model classes under consideration are disjoint, generalized noise in the K-L-best model needs to be estimated according to equation (5), so that equations (2) and (3) can be used instead.

G-AIC values, like traditional AIC values, are relative and on the scale of information [13, 37]. As such, they have no physical analog and do not allow for probabilistic geometric model selections when considered on their own. Inequalities (3) and (6), on the other hand, allow for the unambiguous identification of the geometric models which minimize the estimated K-L divergence when used to represent the image data.

For quantified geometric model selections from image data, more is needed than to identify the K-L-best model in the set. The necessary framework for these kinds of quantifications is provided by a generalization/extension of well-known quantifiers from traditional statistics to geometric statistics [13, 21]. What matters for quantifications of geometric model selections are the relative differences of the G-AIC values of the *i* geometric models in a (disjoint or non-disjoint) set. These differences are standardized on the basis of the K-L-best model (i.e. the model with smallest G-AIC) of the set:

$$\Delta_i = G\text{-}AIC_i - G\text{-}AIC_{best} \qquad (7),$$

for all *i* geometric models in the set [13].

The K-L divergence is here a measure of the amount of information lost when a geometric model is used to describe full reality as depicted in 2D images. The totality of the individual pixel values of the raw image represent a good approximation to the full reality of the imaged phenomenon. Although the full reality is not known exactly, one can always estimate model specific K-L divergences as they correspond to the information loss when a model, i.e. a particular idealization, is used to represent the full image reality. Equation (7) is, thus, a measure of distance between each geometric model of a set and the estimated K-L-best model in the set. This measure is obviously zero for the estimated K-L-best model [13, 37], i.e. the one with the smallest K-L divergence.

The relative likelihoods of each of the *i* geometric models are obtained by Akaike's transformation [37]:

$$\ell_i \propto \exp(-\frac{1}{2}\Delta_i) \qquad (8),$$

where $\propto$ means proportional to. Since Equation (7) is equal to zero for the K-L-best model in the set, the K-L-best model will have a relative likelihood of unity, whereas all other models will have relative likelihoods less than unity [37].

The evidence ratio of the $i^{th}$ geometric model to the $j^{th}$ geometric model in a given model set is given by:

$$E_{ij} = \frac{\ell_i}{\ell_j} = \exp[-\frac{1}{2}(\Delta_i - \Delta_j)] \qquad (9).$$

Evidence ratios quantify the strength of evidence in favor of one geometric model with respect to another within the same disjoint or non-disjoint model set [13].

Likelihoods and evidence ratios, while useful inasmuch as they give some quantification of the strength of evidence in favor of a given geometric model in a set, are still relative and not probabilistic. What is needed is a measure of the probability that a geometric model for the data in a set of models for the same data is, in fact, the K-L-best model of that set.

Such a measure exists in the case of traditional (information theory-based) statistics, where individual model probabilities that sum to 100% for an entire set of models are known as Akaike weights or "Bayesian posterior model probabilities" [37]. Geometric Akaike weights (G-AWs), like traditional Akaike weights, are obtained by normalization of the *i* relative model likelihoods for all *R* models in the set:

$$w_i = \frac{\exp(-\frac{1}{2}\Delta_i)}{\sum\limits_{r=1}^{R} \exp(-\frac{1}{2}\Delta_r)}(100\%) \tag{10},$$

where $w_i$ is the geometric Akaike weight (G-AW), or probability that model *i* is the actual K-L-best model in a set of *R* geometric models. Geometric model selection quantifications were (to our knowledge) first extended to geometric model selection by the senior author of this paper [13, 14, 21].

### 3. Often-Ignored Complications to Crystallographic Symmetry Classifications

To illustrate the first of these complications, we need to explain the hierarchic nature of the primitive plane symmetry groups that are pertinent to this paper, as shown in Figure 1. For a complete overview of the hierarchy of plane symmetry groups, the reader is referred to [35]. A graph that represents the complete hierarchy tree of the 17 plane symmetry groups is given in [38].

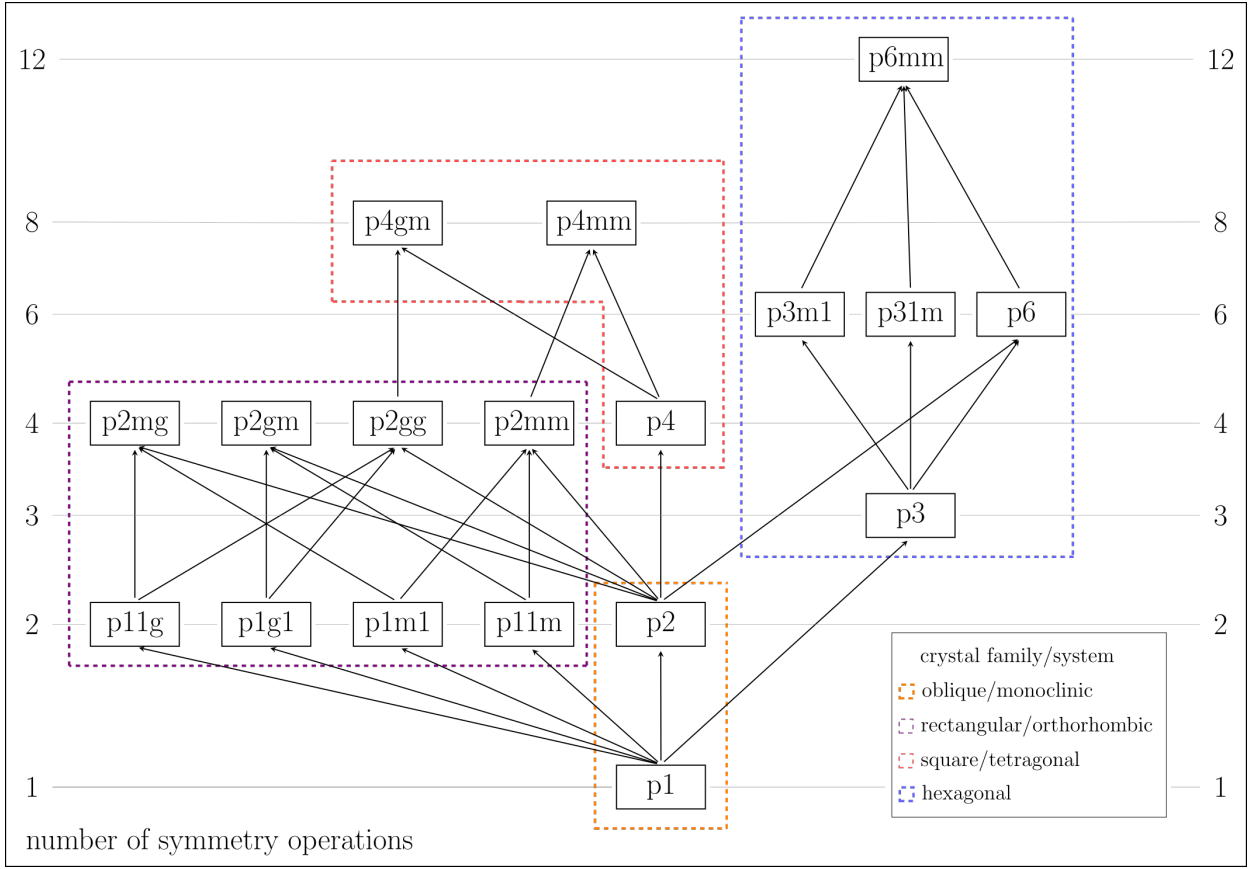

**Figure 1:** : Partial plane symmetry hierarchy tree for our pixel-intensity-value-based G-AIC that allows for the identification of the Kullback-Leibler-best model of a more or less 2D periodic image that features a certain primitive plane symmetry group. Note that this is a partial hierarchy tree that is valid for identifications of the Kullback-Leibler-best plane symmetry models that belong either to the oblique, the rectangular-primitive, or the square crystal family, but not to the hexagonal crystal family. The flocks of plane symmetry groups that belong to each of the four crystal families/systems in 2D are grouped within dotted lines.

Type-I (translationengleiche) maximal subgroup and minimal supergroup relations [35] among the 15 primitive plane symmetry groups are illustrated in Figure 1 as lines connecting certain groups further down in the graph with certain other groups higher up in the graph. Plane symmetry groups appearing lower in the diagram feature less 2D space symmetry than those appearing higher up. A minimal translationengleiche supergroup results from the addition of a new point/site symmetry operation to the already existing point/site symmetry operations of one of its maximal subgroups. Thus, supergroups possess more symmetry operations than subgroups. The values of $k$ represent the number of symmetry operations of each plane symmetry group in both direct space and reciprocal space. (The order of all space groups is infinite as they contain a finite number of point symmetry operations, but infinitely-many translation operations.)

The number of symmetry operations, $k$, of a plane symmetry group is also equal to the multiplicity of the general position in the direct space unit cell of a primitive lattice for that plane symmetry group. For example, any general point in the unit cell of a hexagonal lattice with plane symmetry *p6mm* appears 12 times, while any general point in the unit cell of a square lattice with plane symmetry *p4* appears 4 times. Points with a higher site symmetry than *1* (the identity rotation) possess in all higher symmetric plane symmetry groups a reduced multiplicity. When the image of an approximately 2D translation periodic unit cell is composed of a very large number of pixels (as in this study), there will only be a few pixels with higher site symmetries than *1* for all plane symmetry groups higher than *p1*, so that the differences in the multiplicity of general and special positions of the unit cells can be safely neglected.

The leading letter *p* in the plane symmetry group symbols indicates that the translation-periodic unit cell is primitive, i.e. contains a single lattice point. The numerical symbol that follows the leading letter refers to rotation points in the plane (parallel to the original +z axis of the unit cell in a projection from 3D). In the oblique and rectangular crystal systems, the following two symbols refer to symmetry operations perpendicular to the [1 0] and [0 1] directions, i.e. the +x and +y axes, of the 2D unit cell. In the square and hexagonal crystal family, the two symbols in the third and fourth position of the full Hermann-Mauguin symbol refer to the [1 0] and [1 $\bar{1}$] directions and their symmetry equivalent directions. As both of these directions have symmetry equivalent directions in the square and hexagonal crystal systems, the short Hermann-Mauguin symbol group *pm* is (in its two distinct settings) contained in *p31m* and *p3m1* three times. The same applies to the *p2mm* group with respect to the *p6mm* group.

Note that there are 17 plane symmetry groups and four crystal families/systems in 2D. The flocks of plane symmetry groups belonging to each crystal family/system are grouped within dotted lines in Figure 1. Three of the 15 primitive plane symmetry groups possess two settings, e.g. *p11g* and *p1g1*, so that there is a total of 18 entries in Figure 1. In order to demonstrate the generality of our crystallographic symmetry classification approach, we will apply it to sets of noisy images in section 7 assuming only that each image is higher-symmetric than *p1*, and that its periodicity is completely described by two translation vectors. In other words, the images are assumed to possess site symmetries higher than the 360° (identity) rotation and one of the 14 higher symmetric primitive plane symmetry groups.

As mentioned above, the hierarchic nature of crystallographic symmetries leads to ambiguities when traditional distance measures (e.g. linear residuals) are used for classifications of noisy images into plane symmetry groups. If a noisy input image is best described by a model with plane symmetry *p4mm*, for instance, this would not be readily apparent from examining traditional residuals in the individual symmetry hierarchy branches that end in this particular group; i.e., from the top *p4mm* in Figure 1, downwards over *p4* to the bottom at *p2,* or from *p4mm* over *p2mm*, to

*p1m1*, *p11m,* and *p2* at the bottom. Since the geometric models corresponding to plane symmetry groups *p2*, *p1m1* and *p11m* are the least-constrained symmetry models for image data in these particular plane symmetry hierarchy branches, their residuals will, for noisy (real world) images, always be smaller than those of the *p2mm*, *p4,* and *p4mm* models. Whereas a whole plane symmetry hierarchy branch may feature low traditional residuals, one has no objective measure for how high one could go up in that branch. Hence the above-mentioned advice in [36] can be rephrased as *"when you are in doubt, go up in the most promising symmetry hierarchy branch as high as you think is reasonable".*

The bias correction term in the G-AIC of equation (2), on the other hand, allows one to begin at the image model with the lowest SSR amongst the plane symmetry groups at the bottom rows of Figure 1 (corresponding to multiplicities of the general position 2 and 3) and "climb up" to the next more-symmetric groups of geometric models in each of the individual plane symmetry hierarchy branches, as long as inequality (6) is fulfilled. Because the G-AIC is a bias corrected squared residual, it also informs one when climbing up is no longer permitted, i.e. when inequality (6) is not fulfilled. Note that the image model with plane symmetry group *p1* features a zero-valued SSR in our Fourier space approach, so one cannot start from a model that features only translation symmetry.

In order for a more-symmetric plane symmetry model, i.e. a miminal supergroup to be preferred over a less symmetric model, inequality (6) must be satisfied for all of its maximal type-I translationengleiche subgroups. Upon finding the K-L-best model for the image data in this manner, the generalized noise level for the image data can be estimated using equation (5). This in turn allows one to calculate the G-AIC values, evidence ratios, and G-AWs for all plane symmetry models in any chosen set of plane symmetry models. This is important when an image features genuine pseudosymmetries, as they can correspond to models of plane symmetry groups that are either disjoint from the K-L-best model for the (raw) image data, or will have type-I minimal supergroup—maximal subrgoup relations with the K-L-best model.

Also as mentioned above, crystallographic symmetry classifications are further complicated by pseudosymmetries of the Fedorov type [34] as well as genuine translational and motif-based pseudosymmetries [22]. Fedorov-type pseudosymmetries occur when a 2D periodic image exhibits an approximate plane symmetry that is compatible with either its Bravais lattice or a crystallographically-defined superlattice thereof [35, 40]. Metric-based (translational) pseudo-symmetries occur when a 2D-periodic image has Bravais lattice parameters capable of supporting higher plane symmetry than is actually present in the image (or would be present in the absence of noise) [20]. Motif-based pseudo-symmetries exist when at least one genuine 2D site symmetry closely resembles another site symmetry in the same point symmetry hierarchy branch. A more thorough discussion of these types of pseudosymmetries, with examples, is given in [22].

A very strong translational pseudosymmetry or a metric specialization (within analysis errors) is required to "climb up" from a translationengleiche maximal subgroup to its minimal translationengleiche supergroup whenever the groups belong to different crystal families/systems as marked in Figure 1 by dotted lines enclosing different flocks of plane symmetry groups. Whereas a very strong translational pseudo-symmetry (or metric specialization within error bars) suffices to climb up from the oblique group *p2* to the hexagonal group *p6*, climbing up from the rectangular groups *p1m1* and *p11m* requires re-indexing (also called centering) of the Fourier coefficients, using relations such as $a_{primitive,1} = a_{centered,1} + 0\ a_{centered,2}$ and $b_{primitive,1} = -b_{centered,1} + 2\ b_{centered,2}$ as well as their three-fold rotation symmetry equivalent relations. This fact is the reason why there are no connecting lines between *p1m1* and *p3m1*, *p11m* and *p31m*, as well as *p2mm* and

*p6mm* in Figure 1. Note that re-indexings were not employed in this study, because they are extraneous to its main ideas.

Ambiguities introduced into crystallographic symmetry classifications by inclusion relations and genuine pseudosymmetries as mentioned above are exacerbated by noise and can lead to misclassifications if either the algorithm producing the results or the analyst interpreting them does not take these complications into consideration. As of September 2020, only the information theoretic approach to crystallographic symmetry classifications implemented by the senior author's research group has been able to overcome these obstacles in an objective, probabilistic manner.

## 4. G-AIC Applied to Classifications into Plane Symmetry Groups and Laue Classes

For purposes of plane symmetry group and Laue class classifications from digital images, the data space is taken to be composed of pixel gray level intensities. This data space is one-dimensional. For all Laue classes and plane symmetry groups, the dimension of the model space *d* is zero because we are dealing with points in the Euclidean plane, in the form of pixels. As such, each class/group has co-dimension $r = 1 - 0 = 1$, and degrees of freedom *n* given by:

$$n = \frac{N}{k} \tag{11},$$

where *N* is the number of data points and *k* is either the order of the Laue class in the case of Laue classes or the number of point/site symmetry operations of the plane symmetry group, as given in Figure 1 for the primitive plane symmetry groups. Inserting (11) into inequality (6) yields a much simpler form, suitable for pairwise comparisons of non-disjoint Laue classes/plane symmetry groups:

$$\frac{\hat{J}_m}{\hat{J}_l} < 1 + \frac{2(k_m - k_l)}{k_m(k_l - 1)} \tag{12},$$

where, as above, the subscripts *l* and *m* indicate the less- and more-symmetric models, respectively. Note that inequality (12) is undefined for $k_l = 1$, which corresponds to plane symmetry group *p1*. As mentioned above, no SSR can be calculated for *p1* on the basis of the DFT of the raw image data.

Note that Liu and coworkers derived the same inequality from the same premises [3] for pairwise comparisons of non-disjoint plane symmetry groups in direct space. This inequality is also valid in Fourier space [13]. Inequality (12) is valid when all symmetry models being compared represent the same number of data points, *N*. Appendix 1 gives the generalization of this inequality for the comparison of non-disjoint symmetry models with variable *N*. In the present study, *N* is to a reasonable approximation constant for each image selection (see data tables below and in the appendix), and the SSRs for models with plane symmetry group *p2* are sufficiently small in most cases that inequality (12) will be used to identify a K-L-best geometric model out of a set of 14 models for the test image data that feature the higher-symmetric plane symmetry groups in Figure 1.

Note in passing that a different definition of the G-AIC is necessary for classifications into Bravais lattice types because in that case the generalized noise is not pixel noise, but noise of a positional nature in Fourier space [15, 23].

Inequality (12) is used in the same manner as inequality (6) given above. In the previous example involving the two plane symmetry hierarchy branches that end in *p4mm*, one would begin with the geometric model that features the smallest SSR at the lowest nodes (above *p1*) of each branch in Figure 1, i.e. either the *p2* model, the *p1m1* model, or the *p11m* model. If this were the *p2* model, one would check if inequality (12) is fulfilled for that plane symmetry model and the *p2mm* model, as well as for the *p4* model. Upon doing so, a few distinct outcomes are possible.

First, if ascending to both *p2mm* and *p4* from *p2* is forbidden, the model for the image data corresponding to plane symmetry group *p2* would be the estimated K-L-best of the primitive plane symmetry models under consideration, see Figure 1. If the discussion were to be expanded to cover more than the two symmetry hierarchy branches that end in *p4mm*, in Fig. 1, one must also check that inequality (12) is also not fulfilled for the geometric model pairs *p2mg/p2, p2gm/p2, p2gg/p2*, and *p6/p2* before one can conclude that the model with plane symmetry group *p2* is the estimated K-L-best geometric model of the image data.

Second, if ascending to *p4* from *p2* is allowed, the *p4* model would be preferred over the *p2* model in the sense that using it as a geometric model for the image data instead of the *p2* model yields a smaller K-L information loss. This means that the plane symmetries *p4* and *p2* are both genuine, and that there are genuine four- and two-fold rotation points in the translation periodic image to be classified. However, at this point no assessment has been made as to whether there could also be a genuine set of perpendicular mirror lines present in the image. The presence of mirror lines must be ruled out before concluding the *p4* model is the K-L-best geometric model of the image data.

To check for genuine mirror lines in the image, one starts again with the geometric models with plane symmetry groups *p1m1* and *p11m* at the row with $k = 2$ in Figure 1. When inequality (12) is fulfilled for the models that feature both of these maximal subgroups, one can ascend to their minimal supergroup, plane symmetry group *p2mm*, at the row corresponding to $k = 4$, as genuine two-fold rotation points have already been confirmed. The previously confirmed four-fold rotation points will create an additional set of perpendicular mirror lines at 45° angles to the two previously-confirmed mirror lines, and glide lines will also be created. This will be reflected in the model with plane symmetry group *p4mm* fulfilling inequality (12) with respect to both the *p2mm* and *p4* models, and the corresponding *p4mm* model would in this case be the K-L-best geometric model for the image data. If mirror lines are not present, the *p4* model would be the K-L-best plane symmetry model for the image data.

If, on the other hand, ascending to the *p2mm* model from all three of the models corresponding to its maximal type-I subgroups is permitted, but ascending to *p4* from *p2* is forbidden, the *p2mm* model would be the K-L-best plane symmetry model for the image data among the primitive plane symmetry models being considered here.

If one were to broaden the discussion to include symmetry hierarchy branches other than the two that end in *p4mm* in Figure 1, it would become necessary to check if inequality (12) is fulfilled for all of the minimal type-I supergroups of plane symmetry group *p2mm*. Plane symmetry group *p6mm* is such a group, but its $\hat{J}$ residuals need to be calculated on the basis of a re-indexing, which is outside the scope of this study. Note that there is no direct connection between plane symmetry groups *p6mm* and *p2mm* in Figure 1. A maximal subgroup—minimal supergroup relationship does, however, exist between plane symmetry groups *p2* and *p6*, which is realized by a metric specialization, see Figure 1.

In [15], an ad-hoc definition for confidence levels for similar kinds of decisions as obtainable by applying inequality (12) was given for pairwise comparisons of G-AICs for non-disjoint Bravais lattice types. Analogous ad-hoc-defined confidence levels can be obtained for Laue classes and plane symmetry groups in a straightforward manner.

The information content *K* of an experimental image with respect to a decision in favor of the more symmetric model $S_m$ has been introduced by Kanatani in [28] as:

$$K = \sqrt{\frac{G\text{-}AIC(S_m)}{G\text{-}AIC(S_l)}} \leq 1 \tag{13}.$$

For a given pair of non-disjoint plane symmetry group or Laue class models that refer to pixel intensity values, this results in:

$$K = \sqrt{\frac{1 - \frac{1}{k_l}}{1 + \frac{1}{k_l}} \left( \frac{\hat{J}_m}{\hat{J}_l} + \frac{\frac{2}{k_m}}{1 - \frac{1}{k_l}} \right)} \leq 1 \tag{14}.$$

Assuming inequality (12) has been satisfied for a pair of non-disjoint geometric models, *K* will be less than unity, and $S_m$ is a better representation of the image data than the less symmetric model, $S_l$. In the special case that *K* = 1, no decision can be made regarding the more symmetric model being a better K-L representation of the image data, and the confidence level of a decision in favor of either model is zero.

The maximal confidence level of a decision in favor of the more symmetric model (i.e. 100%) is obtained when $\hat{J}_m = \hat{J}_l$, so equation (14) becomes:

$$K_{critical} = \sqrt{\frac{k_m - \frac{k_m}{k_l} + 2}{k_m + \frac{k_m}{k_l}}} < 1 \tag{15},$$

where $K_{critical}$ is the critical information content, at or beyond which such a decision is possible. $K_{critical}$ can be used to normalize the maximal confidence level to certainty (100%). This yields:

$$C_m = \frac{1 - K}{1 - K_{critical}} (100\%) \tag{16},$$

for the confidence level of a decision in favor of a more symmetric model on the basis of a pair of geometric models that are non-disjoint.

It is important to emphasize once more that equation (16) is only an ad-hoc definition and not part of Kanatani's original framework. A confidence level below 50% does not imply inequality (12) and any conclusions drawn from it are invalid [15]. Low confidence levels instead mean one is near, but still below, the maximal value of $\hat{J}_m/\hat{J}_l$ where the more symmetric model is preferred over its less symmetric counterpart (inequality 12), and careful consideration is needed to ensure the preconditions of applying the G-AIC methodology have in fact been met [15].

As mentioned above, these preconditions are that systematic errors are reasonably small in comparison to random errors, and generalized noise is both small in comparison to the plane symmetry signal and distributed in an approximately Gaussian manner. Note that it is the generalized noise per asymmetric unit of the translation-periodic unit cell that enters equations (1), (2), (4), and (5). The generalized noise level can always be suppressed further by processing an *experimental* image (rather than a *stitched-together* image as mentioned below) with a large number of unit cells over which one averages automatically just by calculating the discrete Fourier transform of the image intensity (which is equivalent to enforcing plane symmetry group *p1* on the Fourier coefficients). Because noise is so effectively suppressed by calculating the DFT, the precondition of small generalized noise can in general be assumed to be satisfied when a large number of unit cell repeats is present in the input image or selection of the image for which the classification is made.

The assumption of effective noise reduction per unit cell rests on the assumption that the noise from pixel to pixel is, to a sufficient approximation, not correlated, but Gaussian distributed. In statistical terms such a property of noise is often referred to as independent and identically distributed (and abbreviated as *i.i.d.* or *iid*). The generalized noise is then reduced by a factor of approximately one over the square root of the number of unit cell repeats, just by calculating the discrete Fourier transform. As implicitly mentioned above, averaging over the asymmetric unit is obtained by the appropriate symmetrization of the structure-bearing Fourier coefficients of the image intensity, as described below in Section 5.

The K-L-best geometric model, $S_{best}$, is chosen according to the above-stated procedure by the subsequent examination of non-disjoint symmetry model pairs in all pertinent symmetry hierarchy branches. This model minimizes the expected Kullback-Leibler divergence [13, 37] when it is used to represent the image data. In other words, information loss is minimized when a processed image that has had the restrictions of the underlying plane symmetry group of the K-L-best model imposed upon it is used to represent the raw image data (after back-transformation to direct space).

From this K-L-best geometric model, an estimate of squared generalized noise, in analogy with equation (5), is obtainable according to:

$$\hat{\epsilon}^2 \approx \frac{\hat{J}_{best}}{r_{best}N - n_{best}}$$ (17).

This estimate of squared generalized noise is then used to compute full G-AIC values for all models $S_i$ within a set of geometric models, disjoint or not, according to Equation (2).

Once the G-AIC values have been computed for all geometric models, relative likelihoods, evidence ratios, and G-AWs can be computed according to equations (7) through (10) for a set of models which one wants to compare with respect to their predictive power. What results is an information-theoretic framework for crystallographic symmetry classifications that allows for probabilistic (as opposed to purportedly definitive) classifications. These classifications are spread out over several geometric models and can be updated when better (i.e. less noisy) image data and more accurate processing algorithms become available. In the case of experimental data, better results will also be obtained with more accurate image recording equipment and microscopes. Recording experimental images with as many pixels as possible while ensuring that there is also a high number of unit cells of the more or less 2D periodic pattern is strongly recommended.

## 5. Fourier Space and CIP

As mentioned above, CIP is a Fourier/reciprocal space averaging technique that takes advantage of the more-or-less 2D periodic nature of atomic-resolution transmission electron microscopy images of inorganic [10, 25] and organic [25] crystalline arrays. CIP can work just as well for any image that is (or contains regions that are) more or less 2D periodic. However, as with traditional Fourier filtering, better results will be obtained with more periodic repeats of the unit cell motif present in the image/region. Figure 2 serves as illustration of key concepts in this section.

As the computational symmetry community frequently deals with images that contain relatively few 2D periodic repeats, they tend to avoid Fourier space methods at the cost of decreased computational efficiency and inferior noise reduction. Freely available software exists, such as Microsoft Image Composite Editor (ICE) [33], that is capable of seamlessly stitching together periodic images into much larger images containing many more repeats with relative ease. No additional noise is introduced by this process and such an approach is a simple solution that allows members of the computational symmetry community to reap the benefits of CIP. However, one needs to be aware that because several copies of the same image with the same generalized noise are being stitched together, the overall noise in the image will then with necessity be somewhat correlated, so that the i.i.d. assumption is no longer strictly fulfilled.

CIP can be described as an essentially three-step procedure. First, the 2D discrete Fourier transform (DFT), $R(h\ k)$, of a $P \times Q$ input image $D(x, y)$ is calculated, where $P$ and $Q$ are direct space image dimensions in pixels, as:

$$R(h\ k) = \sum_{x=0}^{Q-1} \sum_{y=0}^{P-1} D(x,y) \exp\left[2\pi i \left(\frac{xh}{Q} + \frac{yk}{P}\right)\right]$$
(18).

While equation (18) is correct in general, in practice many algorithms for calculating the 2D DFT, such as the fast Fourier transform algorithm, require the input image or selection to be square $Q \times Q$ pixels where $Q$ is a power of two [41], so equation (18) becomes:

$$R(h\ k) = \sum_{x=-\frac{Q}{2}}^{\frac{Q}{2}-1} \sum_{y=-\frac{Q}{2}}^{\frac{Q}{2}-1} D(x,y) \exp\left[\frac{2\pi i}{Q}(xh + yk)\right]$$
(19),

where the summation limits are from $-Q/2$ to $Q/2-1$. This form of the summation places the direct space origin (x = 0, y = 0) at the center of the raw image (or selection from that image).

The summation yields a $Q \times Q$ array of complex Fourier coefficients $R(h\ k)$, (abbreviated below as FCs) each with associated amplitude $|R(h\ k)|$ and phase angle $\phi = [(2\pi/Q)(xh + yk)]$. These FCs contain all of the crystallographic structural information in the image (or image selection) in addition to all of the information on the generalized noise. The inverse 2D discrete Fourier transform of all $Q$ complex FCs reproduces the original (noisy) image.

The amplitude map (a $Q \times Q$ array of pixels containing only FC amplitudes) of the transformed image reveals a regular array of bright spots that lie on the nodes of the image's reciprocal lattice, if the image is more-or-less-2D periodic. The amplitude map calculated from a noisy 2D periodic image is similar in appearance to a Buerger precession X-ray diffraction pattern [42] or a precession electron diffraction pattern from a crystal [43, 44]. The 2D point symmetry of these bright spots corresponds to the image's Laue class. The FCs corresponding to the centroids of these bright spots contain all idealized periodic structural information present in all pixels of the input image plus a small perturbation due to generalized noise values at these particular locations in reciprocal space. All other intensity variations between the peaks corresponding to structure-bearing FCs in the amplitude map are due to the effects of the generalized noise.

The reciprocal lattice of a more or less 2D periodic image is fully described by basis vectors $a^*$ and $b^*$, as well as the angle between them $\gamma^*$. The basis vectors are estimated by least squares at this stage. A very high degree of accuracy is necessary in the estimation of reciprocal lattice parameters in order to obtain accurate direct-space lattice parameters [20], to properly index the structure-bearing FCs, and most importantly to obtain reasonably accurate phases and amplitudes for the structure-bearing FCs [10, 12]. Indexing the structure-bearing FCs refers to assigning each such FC a pair of indices *hk*, where *h* and *k* are integers that refer to position relative to the center of the amplitude map with indices *00* in the [1 0]* and [0 1]* directions, respectively. Note that the reciprocal lattice basis vector $a^*$ points in the [1 0]* direction, whereas reciprocal lattice basis vector $b^*$ points in the [0 1]* direction.

Reciprocal space lattice parameters $a^*$, $b^*$, and $\gamma^*$ are related to direct space lattice parameters *a, b,* and *γ* according to:

$$a^* \cdot a = 1 \tag{20},$$

$$b^* \cdot b = 1 \tag{21},$$

$$a^* \cdot b = b^* \cdot a = 0 \tag{22},$$

and

$$\gamma = 180° - \gamma^* \tag{23}.$$

Equations (20) and (21) imply that if *a* is longer than *b* for a given 2D-periodic image, then $b^*$ is longer than $a^*$. Equation (22) implies that $a^*$ is perpendicular to *b* and $b^*$ is perpendicular to *a*. By crystallographic standard convention [35], *γ* is taken to be greater than 90°, which by equation (23) necessarily means $\gamma^*$ is always less than 90°.

An example image that possesses plane symmetry group *p1* by design and is free of added noise, but with very strong *p2* pseudosymmetry as well as strong *p1g1* pseudosymmetry is depicted in direct space in Figure 2a [20]. Its power spectrum (as calculated with MATLAB) is shown in Figure 2b. The power spectrum is similar in appearance to an amplitude map, but is a distinct mathematical entity as it contains the squared amplitudes of the complex FCs. By visual inspection of the reciprocal space image, it has twofold rotational symmetry about the central pixel, and all of the *0k* FCs with odd *k* indices are comparatively weak due to the glide line pseudosymmetry (perpendicular to the [1 0] direction). The twofold rotation point in reciprocal space is genuine, as the Fourier transform is centrosymmetric. The Laue class of this image is, therefore, *2*.

The clearly resolved two-fold rotation symmetry of Fig. 2b is the result of calculating the discrete Fourier transform so that the image necessarily possesses Laue class *2*, although there are no genuine two-fold rotation points in the direct space image, Figure 2a. Figure 2c contains an illustration of the image's direct space unit cell and lattice parameters *a, b,* and *γ*, while Figure 2d contains an illustration of its reciprocal unit cell and lattice parameters $a^*$, $b^*$, and $\gamma^*$.

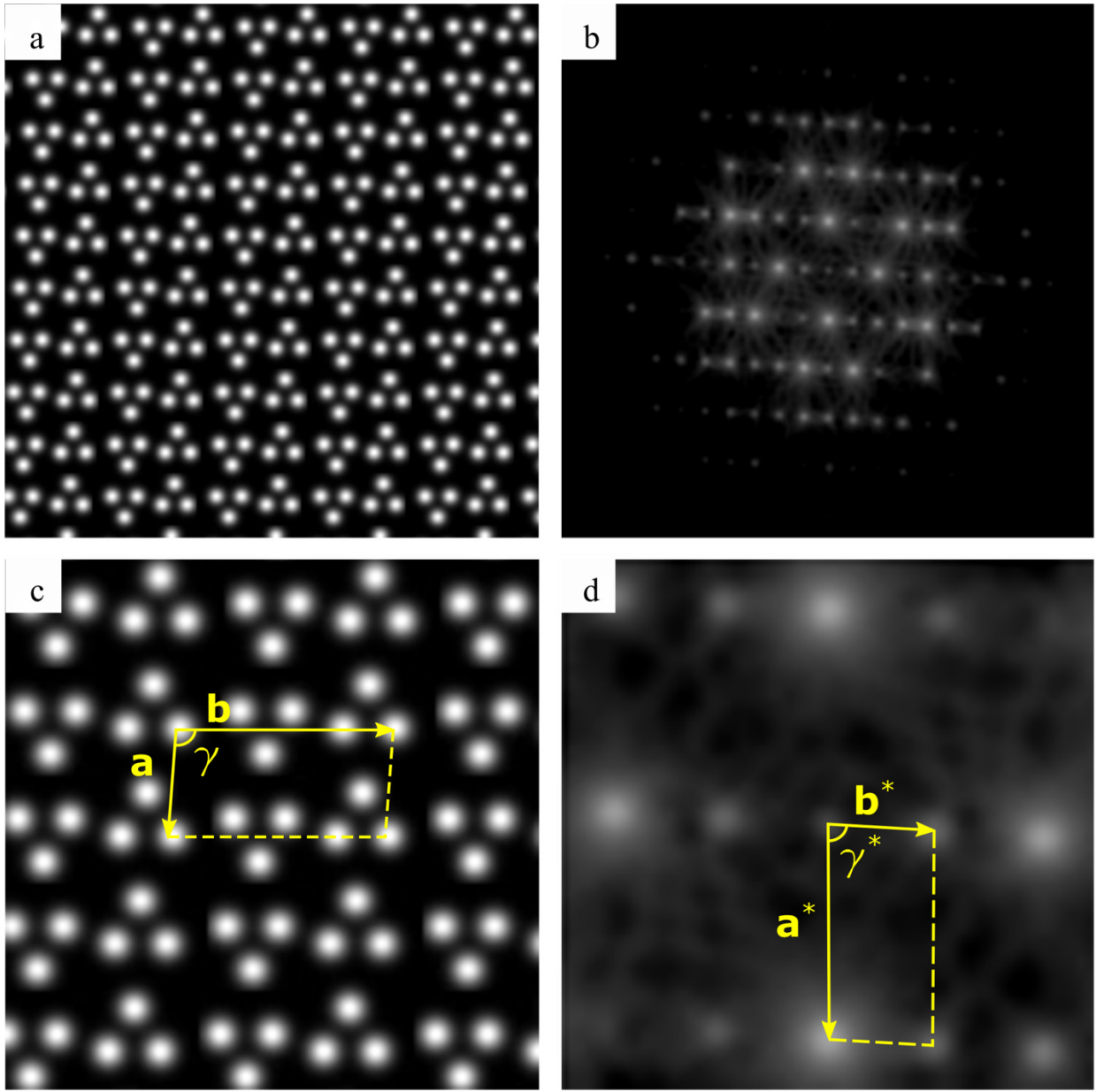

**Figure 2:** A synthetic, strictly 2D-periodic image (without added noise) that features plane symmetry *p1* by design, possessing prominent *p2* and *p1g1* pseudosymmetries as presented in [20] (a), and its power spectrum (b). The regularly arrayed bright spots, i.e. the real-valued squares of the structure-bearing FCs in the power spectrum, have two-fold rotational symmetry about the central pixel. This corresponds to Laue class 2. A direct space unit cell, with parameters a, b, and $\gamma$ is illustrated in (c), while a reciprocal space unit cell with parameters ***a*****, ***b*****, and $\gamma$* is illustrated in (d). The star after the lattice parameter symbols represents here a 2D periodic reciprocal space as that is the crystallographic standard representation [42]. Note that there is very little intensity between the regularly arrayed bright spots as clearly visible in (d). The fact that there is some intensity between the bright spots is due to both the finite size of the image and the particulars of the image processing algorithm that has been used. As the image is free of noise by design, the intensity between the bright spots is particularly low.

It is worth mentioning that, while the unit cell origin can be placed anywhere when the underlying plane symmetry group is *p1*, there are standard crystallographic origin conventions for all plane symmetries higher than *p1* [35]. In other words, *p1* is the only plane symmetry group where arbitrary placement of the unit cell origin is allowed. For all higher-symmetric plane symmetry groups, systematic placement of the unit cell origin at a position of greatest site symmetry (e.g. on rotation points and in their absence on mirror or glide lines) greatly simplifies subsequent calculations [35]. This is achieved by what is called "origin refinement" in CRISP [10, 36].

Once the *N* structure-bearing FCs, denoted by *F*(*h k*) to distinguish them from the non-structure-bearing FCs *R*(*h k*), of the discrete Fourier transform of an image are indexed, the symmetry of each of the 16 plane symmetry groups higher than *p1* can be enforced on them and the respective symmetries referred to the standard crystallographic origins. In practice, this consists of replacing *n* symmetry-related-structure-bearing FC amplitudes by their average:

$$|F_{sym}(h\ k)| = \frac{1}{n} \sum_{j=1}^{n} |F_{j,\ related}(h\ k)| \qquad (24),$$

whereas each set of *n* symmetry-related-structure-bearing FC phases are replaced according to:

$$\phi_{sym}(h\ k) = \tan^{-1}\left[\frac{\sum_j \sigma_j \eta_j \sin(\phi_{j,obs}(h\ k))}{\sum_j \sigma_j \eta_j \cos(\phi_{j,obs}(h\ k))}\right] + \begin{cases} 0° & \text{if } \sum_j \sigma_j \eta_j \cos(\phi_{j,obs}(h\ k)) > 0 \\ 180° & \text{if } \sum_j \sigma_j \eta_j \cos(\phi_{j,obs}(h\ k)) < 0 \end{cases} \qquad (25).$$

The summations in equation (25) are implicitly taken over *n* symmetry-related FC phases. The term $\sigma_j = 1$ if the symmetrized phases should be equal to observed phases according to plane group relations or $\sigma_j = -1$ if they should differ by 180°, and $\eta_j$ is a weighting factor that is usually set to the amplitude of the corresponding *F*(*h k*) [10]. Unrelated FC amplitudes and phases are left unchanged [10].

Equation (25) requires the calculation of $2N + 1$ trigonometric functions, and as such a more compact and computationally efficient form of the same function is desirable. If, instead of separating each structure-bearing FC into real-valued amplitude and phase angle components, each structure-bearing FC is handled as a full complex number in four-dimensional (*h*, *k*, *F*, *ϕ*) Fourier space, such a form can be derived, as shown originally in the appendix of [16] but reiterated here in Appendix 2, for completeness.

The origin refinement is obtained by calculating and minimizing the CRISP phase residuals (to be discussed in more detail below) for all points (pixels) within a unit cell with arbitrary origin that arose from the particulars of the summations in (18) or (19). The calculated amplitude residuals do not change much from point to point. As a matter of fact, they would be independent of the position within the unit cell for a perfectly 2D periodic (noise-free) image that is processed with a perfect algorithm. The phase residuals do, however, change significantly with the position of the point about which they are calculated. The points of lowest phase residual determine the proper crystallographic origin of the unit cell and all FC phases are subsequently shifted to that standard origin.

Without the phase origin refinement, i.e. finding the crystallographic origin of the unit cell, the CRISP residuals would be utterly meaningless. Similarly, if one were to calculate SSRs for an arbitrary unit cell origin that are to be used to calculate G-AIC and G-AW values without prior phase origin refinement, these values would not allow for a sensible crystallographic model selection for the input image data.

After enforcement of a particular plane symmetry on FCs, and the subsequent origin refinement, one can calculate the inverse 2D DFT of the *N* symmetrized structure-bearing FCs:

$$f(x,y) = \frac{1}{N^2} \sum_{h=-\frac{N}{2}}^{\frac{N}{2}-1} \sum_{k=-\frac{N}{2}}^{\frac{N}{2}-1} F(h\ k) \exp\left[-\frac{2\pi i}{N}(xh+yk)\right]$$

(26),

in order to obtain a symmetrized direct space image, *f(x, y)*. (While this enables CIP to obtain the coordinates of atoms or molecules in a noisy 2D periodic image from an inorganic or organic crystal, this is not necessary for the crystallographic symmetry classifications into plane symmetry groups as such.)

This symmetrized version of the input image strictly adheres to the crystallographic constraints imposed by the plane symmetry group to which it has been symmetrized. This applies even to the rather small amount of noise that could not be removed by the filtering for the structure-bearing FCs, but was reduced significantly by averaging over the properly defined asymmetric unit. In other words, symmetry-enforced models of the input data corresponding to different plane symmetry groups have been created by this procedure that can now be compared against each other, and the one model that best adheres to the 2D periodic motif signal can be selected by our G-AIC procedure. There is, however, no need to produce a symmetrized image as one would do within CIP, because our geometric model selection is, for computational efficiency, done in Fourier space as well.

If the structure bearing FCs were symmetrized to a plane symmetry group that is outside of the correct symmetry hierarchy branch, the Fourier back-transformed image would look significantly different from the raw image. In other words, enforcing an incorrect plane symmetry group yields an incorrectly symmetrized version that does not resemble the original image, and from which no meaningful information can be extracted by CIP. Symmetrized images corresponding to maximal type-I subgroups or minimal type-I supergroups of the correct plane symmetry model will, on the other hand, be similar in appearance to the raw image and have similar amplitude and phase residuals to the correct plane symmetry model. However, the benefits of the symmetrization process will then not have been optimally reaped. An image that has been symmetrized to a minimal type-I supergroup of the correct plane symmetry group will again look different from the raw image.

In other words, the maximum amount of structural information can be extracted from the set of structure-bearing FCs that have had the restrictions of correct plane symmetry group enforced upon them. Enforcing the restrictions imposed by correct plane symmetry group on the image data, i.e., the observed structure-bearing FCs calculated from an input image, amounts to averaging over up to 12 times more information than traditional Fourier filtering for primitive unit cells. This is because the unit cell motif is averaged on the basis of the correctly defined asymmetric unit as opposed to the entire unit cell. Averaging over the smaller asymmetric unit results in greater reduction of generalized noise, which allows for more reliable extraction of pertinent atomic structural information in direct space within CIP.

If, as is frequently the case, the correct plane symmetry group possessed by an input image is not known beforehand, either an educated guess has to be made by the user, or our G-AIC approach is to be used. To assist in making such a guess, commercially available CIP software CRISP provides (after a proper phase origin refinement) three figures of merit to assist users in deciding which plane symmetry group might be present in the input image. These are the amplitude residuals:

$$F_{res} = \frac{\sum_h \sum_k ||F_{obs}(h\ k))| - |F_{sym}(h\ k))||}{\sum_h \sum_k |F_{obs}(h\ k)|} \qquad (27),$$

the phase angle residuals:

$$\phi_{res} = \frac{\sum_h \sum_k |F_{obs}(h\ k)||\phi_{obs}(h\ k) - \phi_{sym}(h\ k)|)}{\sum_h \sum_k |F_{obs}(h\ k)|} \qquad (28),$$

and the so-called extinction ratios [10]:

$$\frac{A_o}{A_e} = \frac{\frac{1}{n}\sum_{hk} |F_{obs}(h\ k)|_{forbidden}}{\frac{1}{m}\sum_{hk} |F_{obs}(h\ k)|_{expected}} \qquad (29),$$

where *n* is the number of forbidden (by the restrictions imposed by the particular plane symmetry group) FCs that were nonetheless observed and *m* is the number of expected FCs that were observed. The summations in (27), (28), and (29) are taken over all structure-bearing FCs.

Of these three figures of merit, the phase residual is considered the most important, as FC phases contain much more structural information than FC amplitudes [10, 23]. The extinction ratios are not a residual in any sense of the word and do not provide a direct comparison between the raw and symmetrized FCs, whereas both the CRISP amplitude and phase residuals do. Instead they refer to the raw FCs only and give the ratio of the average amplitudes of the observed FCs that should be systematically absent (if the image possesses a plane symmetry group with glide lines or a lattice point centering), to the average of the FCs for which no such restrictions apply [10].

In the $^{Ao}/_{Ae}$ ratio, the subscript *o* stands for *odd*, subscript *e* stands for *even*. Both meanings are somewhat metaphorical, as crystallographic reflection conditions are listed in the recent editions of the International Tables for Crystallography [35]. "Even reflections" are listed there that are not extinct (absent) in a systematic way. For example, in plane symmetry group *p1g1*, the reflection condition is that *k* must be an even number for all *0k* reflections.

The extinction ratios calculated according to equation (29) are, therefore, only useful for identifying whether a centered lattice and/or glide lines might be present, as they would lead to systematic absences of certain peaks (those with certain odd indices or odd *h + k* sums) in the amplitude map of images with corresponding underlying symmetry operations. Note that lattice centering results in a doubling of the direct space area of the unit cell and it is only due to this doubling that odd *hk* FCs are systematically absent in a DFT amplitude map.

The interpretation of all three CRISP figures of merit is essentially subjective, and the reliability of crystallographic symmetry classifications based on them is dependent on the experience and intuition of the user. An experienced crystallographer will typically have knowledge of the general crystal chemistry of their sample, and may have 3D data and/or prior knowledge of the imaged object, so they will typically make an acceptable decision, although they may over- or underestimate the plane symmetry due to the hierarchic nature of symmetries. This means they are likely to identify at least the correct plane symmetry hierarchy branch on the basis of a set of low phase residuals, but may not always identify the best plane symmetry model of the image data.

It is crucial to note that equations (27) through (29) do not consist of SSRs, which means they are not compatible with the information theoretic approach outlined in Section 4. In the following section, we present new SSRs for complex FCs as well as (real-valued) FC amplitudes, which are suitable for calculating G-AIC values, as well as associated evidence ratios and G-AWs, for plane symmetry groups and Laue classes, respectively.

## 6. New SSRs for Information Theory-Based Crystallographic Symmetry Classifications

The CIP software package CRISP outputs data files corresponding to each enforced plane symmetry group on a noisy 2D periodic image that contain both observed and symmetrized structure-bearing FC amplitudes and phases. These can be used to re-express the observed and symmetrized FCs as complex numbers Fobs and Fsym, respectively. By summing the absolute squares of the difference between each pair of observed and symmetrized FCs, an SSR for full complex FCs is then obtained as:

$$\hat{J}_{FC} = \sum_{j=1}^{N} (F_{j,obs} - F_{j,sym})^{*}(F_{j,obs} - F_{j,sym}) = \sum_{j=1}^{N} |F_{j,obs} - F_{j,sym}|^2 \tag{30},$$

where the sum is over all $N$ complex structure-bearing Fourier coefficients. Note that the symbol * signifies in this case the complex conjugate (of the complex difference between observed and symmetrized FCs). It is worthwhile to mention also that crystallographic origin refinements, as discussed above, could just as easily be carried out on the basis of equation (30) as opposed to the traditional CRISP phase residual of equation (28). A similar SSR can be defined for the FC amplitudes, which is useful for geometric model selections for Laue classes:

$$\hat{J}_{L} = \sum_{j=1}^{N} ||F_{j,obs}| - |F_{j,sym}||^2 \tag{31},$$

where the sum is again over all $N$ structure-bearing complex Fourier coefficients.

CRISP arbitrarily maps observed amplitudes to values between 1 and 10000 = $|F_{max,obs}|$ in equations (32) and (33) below. As such, using the observed and symmetrized amplitudes without modification can yield values for $\hat{J}_{FC}$ and $\hat{J}_{L}$ that, when used in equation (2), can subsequently yield nonsensical results for equations (8) through (10), due to the presence of the exponential function in conjunction with large numbers in those equations.

To provide a pragmatic solution to this problem, we normalized the observed and symmetrized FC amplitudes by the maximum observed value (i.e. $10^4$), yielding:

$$\left|\tilde{F}_{j,obs}\right| = \frac{|F_{j,obs}|}{|F_{max,obs}|} \tag{32},$$

and

$$\left|\tilde{F}_{j,sym}\right| = \frac{|F_{j,sym}|}{|F_{max,obs}|} \tag{33}.$$

Doing so does not alter the basic form of (30) and (31), i.e. both are still sums of squared residuals, and their use in our information-theoretic approach discussed above is still justified. All we have done is map numbers from one arbitrary scale to another with a range between $5 \times 10^{-3}$ and one (in the default setting of CRISP).

Using these normalized FC amplitudes in place of their non-normalized versions, equations (30) and (31) were implemented by the first author in a MATLAB script that imports a set of *.hka files output by CRISP, calculates equations (30) and (31), as well as (27) and (28) for comparison. The script allows the user to save output as Microsoft Excel spreadsheets containing calculated values in tabulated form.

Results contained in the output of our script can then be utilized to classify the most-likely plane symmetry group and Laue class of the input image according to the information theoretic approach outlined in Section 4. The methodical details of and results from such classifications are described below in the following section.

One needs to be aware that the *.hka files of CRISP were designed to provide good results for electron crystallography analyses based on typical images that were recorded in high-resolution phase-contrast transmission electron microscopes. Similarly, the whole program in its default settings has been optimized for this purpose. There is a range of optional settings which may lead to better or worse results, which are more or less close to the unknown truth, with respect to the extracted atomic coordinates for certain crystals and imaging conditions. Of particular importance for good results is the integration of the structure-bearing FC amplitudes, for which several optional settings exist in CRISP.

For the purpose that we used *.hka files of CRISP here, they were definitively not optimized. Small systematic errors are therefore made by us by relying on these files in their current form. As a first step to the development of our own computer program that eventually allows for objective crystallographic symmetry classifications, in future work we will augment the *.hka files that CRISP outputs in order to reduce these systematic errors. In line with the aims of this paper, we want to demonstrate the principal outline of our method here (without being overly concerned with the optimization of our results at the present time). Moderately large sources of systematic errors in our analysis stem from the *.hka files not being optimized for our approach, as these files often do not contain a uniform number of structure-bearing FCs, as amplitudes of zero are not allowed.

Equations (1) through (6), as well as inequality (12), assume the number of data points $N$ is non-model-specific, i.e. the same for all models. To account for this, a more general form of (12) was derived to be used for pairwise comparisons of non-disjoint plane symmetry groups with model-specific numbers of data points $N$:

$$\frac{\hat{J}_m}{\hat{J}_l} < 1 + \frac{2(k_m - \frac{N_m}{N_l}k_l)}{k_m(k_l - 1)} \qquad (34).$$

A derivation of (34) is provided in the second appendix. Due to the nature of the images that we processed in this study, we have a rather large and nearly constant $N$ in all cases. As such, using inequality (34) in place of inequality (12) to estimate the K-L-best plane symmetry model in a set of plane symmetry group models does not greatly impact the application of our information-theoretic approach in this study. G-AIC values, associated evidence ratios and G-AWs, are calculated in the same manner as before, but with model-specific $N$ values. The only difference when inequality (34) is used instead of (12) is that the right-hand side of the inequality will be slightly larger or smaller, depending on whether $N_m/N_l$ is smaller or larger than unity.

## 7. Test Image Sets and Results of their Plane Symmetry Classification

*7.1 Description of Test Images and Procedural Details*

*7.1.1 Test images*

Our test images were created from a synthetic reconstruction of an Islamic mosaic found at the Aq-Saray Palace of Shar-i-Sabz in modern-day Uzbekistan that originally appeared in reference [9]. This synthetic image possesses *p2* symmetry by design, but also contains strong metric- and motif-based *p3* and *p6* pseudosymmetries. It contains approximately 30 periodic repeats, and no added noise. Since 30 periodic repeats is sometimes too few for crystallographic symmetry classifications in Fourier space to be effective, six copies of the original image were stitched together using Microsoft ICE to create a 2970 × 2048-pixel image with approximately 150 full repeats of the unit cell motif. This straightforward step ensures proper functioning of the CRISP program in its default settings and of our subsequent classifications. A selection from the original image containing no added noise is shown in Figure 3.

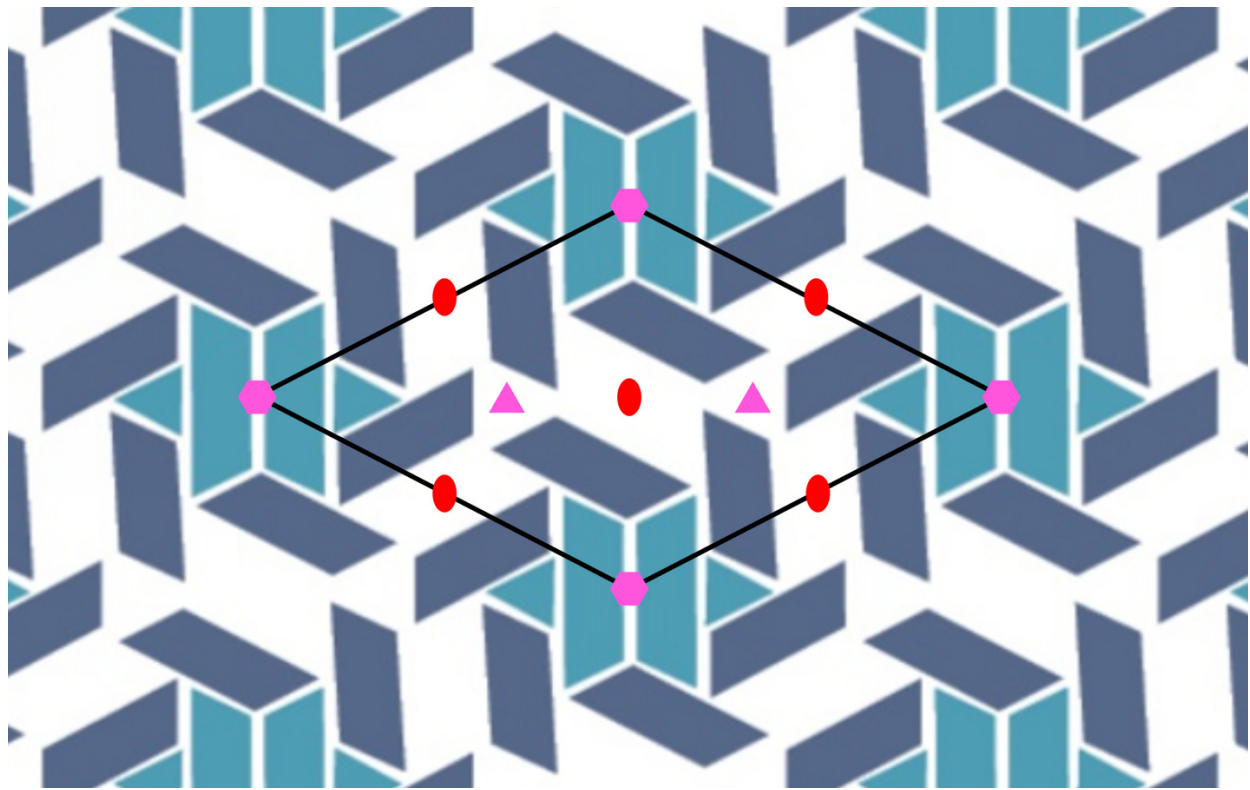

**Figure 3:** A section of a synthetic, noise-free, 2D periodic image with approximately 150 periodic repeats. A single unit cell is outlined in black. Genuine, two-fold rotational symmetry points (that are genuine site symmetries of plane symmetry group *p2*) are indicated by red ellipses; pseudo-3- and 6-fold rotational symmetry points are indicated by magenta triangles and hexagons, respectively. Note that the pseudo-6-fold rotational points are in fact genuine two-fold rotational symmetry points (as 2D point group/site symmetry 6 contains 2D point group/site symmetry 2), while the positions where the pseudo-3-fold rotation points are located are not so-called special positions in plane symmetry group *p2*, meaning they have multiplicity of 1 rather than 3.

The highest site/point symmetry present in this image is two-fold rotational symmetry about the centers of the blue "pinwheels", midway between each pinwheel and its nearest neighbors, as well as in the center of the unit cell. Several two-fold rotation points are denoted in Figure 3 by red ellipses. However, the motif also contains elements that strongly resemble, but are distinctly not, 3- and 6-fold rotational symmetry points, which are denoted by magenta triangles and hexagons, respectively.

Additionally, while the Bravais lattice of this image is oblique by design, meaning $|\mathbf{a}| \neq |\mathbf{b}|$ and $\gamma \neq 120°$ or $90°$, there is very strong hexagonal translational pseudosymmetry that falls short of a metric specialization. This means that within reasonable error bars of a crystallographic analysis program [20], the direct space lattice angle $\gamma$ is definitively not equal to $120°$. In case of a metric specialization, the properties of the lattice, i.e. its metric tensor, would within limits that are set by the analysis algorithm allow for a higher-symmetric translation-periodic motif than is actually present in the image.

The combination of translation- and motif-based pseudosymmetries is compounded by the fact that *p2* and *p3* are both maximal type-I subgroups of *p6*. This is apparent in Figure 1, as both *p2* and *p3* are connected to (and therefore contained within) *p6,* but not to each other. In other words, *p6* is non-disjoint with both *p2* and *p3*, but *p2* and *p3* are disjoint from one another.

From the "noise-free image with approximately 150 repeats, nine noisy versions (comprising two sets) were created using the freely available image editing software GIMP [45]. Two basic types of noise were added: so-called "red-green-blue" (RGB) noise and "spread" noise. Whereas RGB noise does not change a pixel's position and only affects the pixels gray level intensity, spread noise does change a pixels position in a random manner. The effects of each type of added noise are illustrated in Figure 4, which contains representative samples consisting of slightly more than one unit cell from each noisy (and the noise free starting) image.

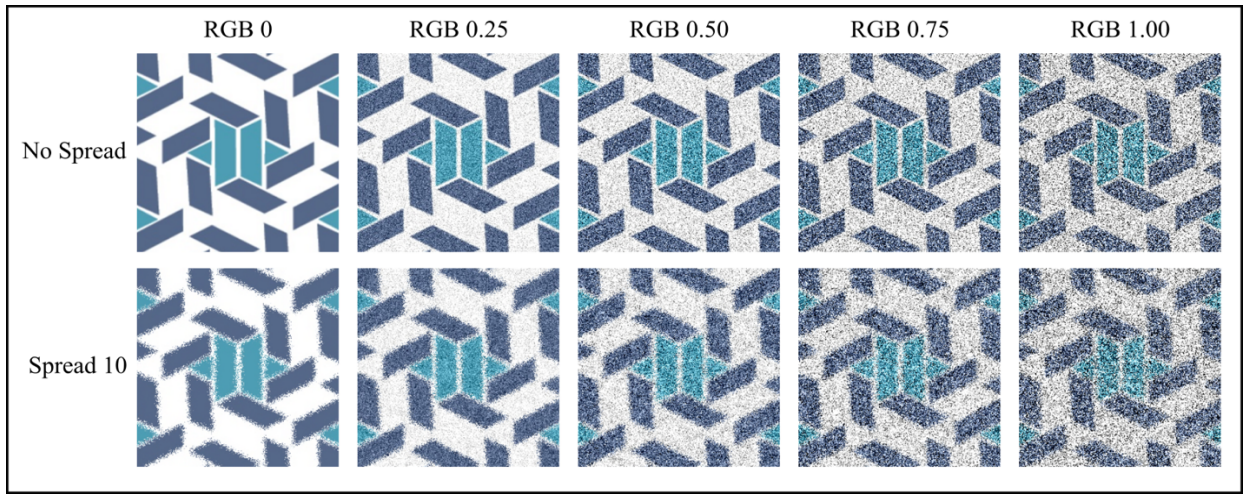

**Figure 4:** Illustration of the effects of both types of added noise on the noise-free motif (top left). The first row corresponds to test images without spread noise, whereas images in the second row have spread noise within a 10-pixel horizontal and vertical distance added. Each column corresponds to a different RGB noise level that increases by 0.25 increments from each column left-to-right. As more noise is added, it becomes more difficult to determine the point symmetry of the pinwheel. What is clearly two-fold rotational symmetry in the noise-free image could reasonably pass for 3- or 6-fold rotational symmetry in the noisier versions.

The spread noise filter of GIMP *"swaps each pixel in the active layer or selection with another randomly chosen pixel"* within a user-specified horizontal and vertical distance [41]. Whereas the word "randomly" in the previous sentence suggests some form of a Gaussian distribution, equations (11) to (17) are not set up for the purpose of considering this type of noise as Gaussian when the intensities of pixels are concerned.

Adding spread noise will, therefore, change the distribution of the overall generalized noise. The more such noise is added to an image with RGB noise, the less will the resulting generalized noise be Gaussian distributed with respect to the pixel intensity. (For experimental images, spread noise is somewhat akin to the effect of very fast random drifts of a sample that is imaged in a scanning probe microscope including a scanning transmission electron microscope).

To each of the ten images containing various levels of RGB noise, spread noise was added with horizontal and vertical distance of 10 pixels.

It is apparent in Figure 4 that, even in the image with the most added noise, the motif still contains elements that seem to preclude plane symmetry groups *p3* or *p6* from consideration as the underlying plane symmetry, e.g. vertical and inclined white stripes in the pinwheels that clearly violate restrictions imposed by 3- or 6-fold rotational symmetry. In other words, a preliminary classification as *p2* could conceivably be made by visual inspection of any of the images represented in Figure 4 (although any such classification would be purely subjective and without any quantification).

In order to examine the performance of our approach with input images that are degraded by noise to the point where no classification by visual inspection can reasonably be made, four additional images were created by adding RGB noise of level 1.00 and spread noise with horizontal and vertical distances of 20, 30, 40, and 50 pixels, examples of which are shown in Figure 5. Note the above-mentioned white stripes (which are approximately six pixels wide in the noise-free image) can still be easily resolved in the image with RGB noise level 1.00 and 20-pixel spread noise. With RGB noise level 1.00 and 30-pixel spread noise, the white stripes are still somewhat visible, but they can no longer be reliably resolved in the two noisiest images.

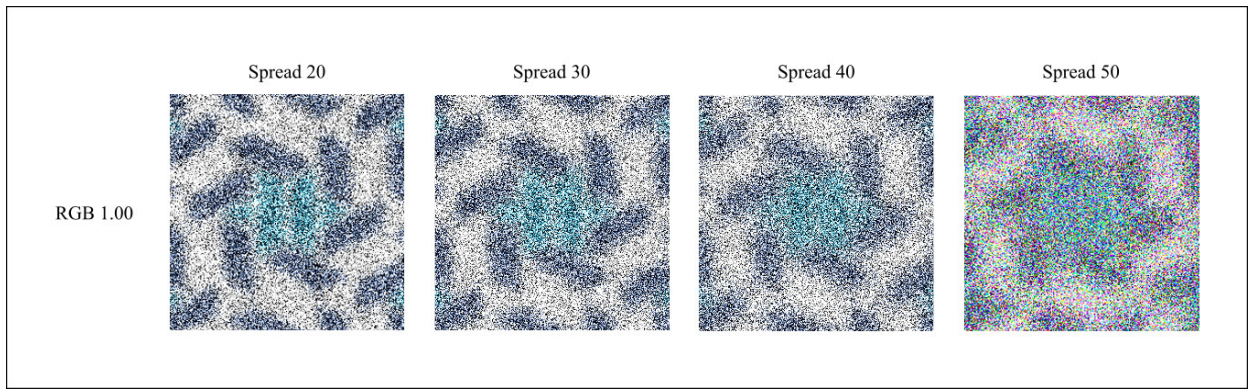

**Figure 5:** Demonstration of the effects of large amounts of spread noise combined with RGB noise of level 1.00. Already with the addition of 20-pixel horizontal and vertical spread noise, the motif has been sufficiently degraded as to render distinctions between 2-fold, 3-fold, and 6-fold rotational symmetry points ambiguous. This ambiguity is exacerbated with each subsequent spread noise level; the addition of 40-pixel or larger spread noise effectively obliterates any trace of the (originally six-pixels-wide) vertical white stripes.

The 14 test images in this study can be divided into three distinct sets. There are essentially three "baseline" images: the noise-free image, the image with 10-pixel spread noise, and the image with RGB noise level 1.00 and 10-pixel spread noise. From the noise-free image, the first set is generated by systematically adding increasing amounts of RGB noise. From the image containing 10-pixel spread noise, the second set was also generated by adding increasing amounts of RGB noise. The third set is generated by adding spread noise of systematically increasing horizontal and vertical distances to the image with RGB noise 1.00 and 10-pixel spread noise.

*7.1.2 Applying CRISP routines to the test images*

Upon importing a direct-space image to the CRISP workspace, the user can select square regions (from 128 × 128 pixels to 4096 × 4096 pixels) or arbitrarily shaped regions of the image from which the 2D DFT is to be calculated. The user can also change a square selection to the largest possible circular selection that fits within the preselected square. Four selections from each test image were processed: 1024 × 1024-pixel square, containing approximately 31 repeats; 1024-pixel diameter circle, containing approximately 23 repeats; 2048 × 2048-pixel square, containing approximately 125 repeats; and 2048-pixel diameter circle, containing approximately 98 repeats. Figure 6 contains the amplitude maps calculated by CRISP for each selection size and shape of the noise-free test image.

The size and shape of the DFT selection window affect the quality of the amplitude map, as demonstrated in Figure 6. Smaller direct space selections necessarily contain fewer periodic repeats, and this leads to broader peaks in the amplitude map, visible in Figures 6a and 6b. In the theoretical limit of an infinitely large input image, its amplitude map would resemble a perfectly periodic array of 2D Dirac delta functions (with specific amplitudes) and both detection of structure-bearing FCs and reciprocal lattice parameter estimation would be trivial. Streaking artifacts parallel to the image edges that are visible in Figures 6a and 6c are caused by discontinuity in the 2D periodicity at the boundaries of the square selection. These artifacts are mostly eliminated by using a circular selection, as shown in Figures 6b and 6d. According to the CRISP 2.2 manual, the extracted FCs from both types of selections of the same area are "*almost the same*" [36], regardless of the larger number of unit cell repeats in the square selections.

At the next stage in the CRISP application, the user can also reduce the radius (in pixels) of a circle enclosing FCs to be included in further calculations. Due to the straight-sharp edges of the individual high contrast blocks in the motif under consideration, at maximum radius (the default setting) this yielded approximately ten thousand FCs for each image, which on occasion produced unresponsiveness and crashes of the CRISP program (under the Windows 10 operating system). To avoid these problems, we discarded all FCs outside a 128-pixel radius circle in the amplitude map for the 1024-pixel selections, and those outside a 256-pixel radius for the 2048-pixel selections. These selections include all of the FCs with reasonably low hk indices, which feature typically the highest amplitudes, see Figure 6.

In the general praxis, discarding FCs outside a certain radius from the center of the DFT reduces the resolution of a symmetrized image produced by Fourier synthesis. FCs closer to the center of the DFT contain lower resolution structural information but are more reliably located in the presence of large amounts of noise because they stand out better in the amplitude map against background intensity variations. The converse is true for peaks located farther from the center of the DFT; they contain higher resolution structural information (as they are the harmonics of the lower resolution peaks) but are more difficult to detect in the presence of increasing amounts of noise. (For crystallographic symmetry classifications of noisy atomic resolution images from organic or inorganic crystals, restricting the analysis to low-index FCs also makes sense, as the resolution of the recording instrument is always approximately known.)

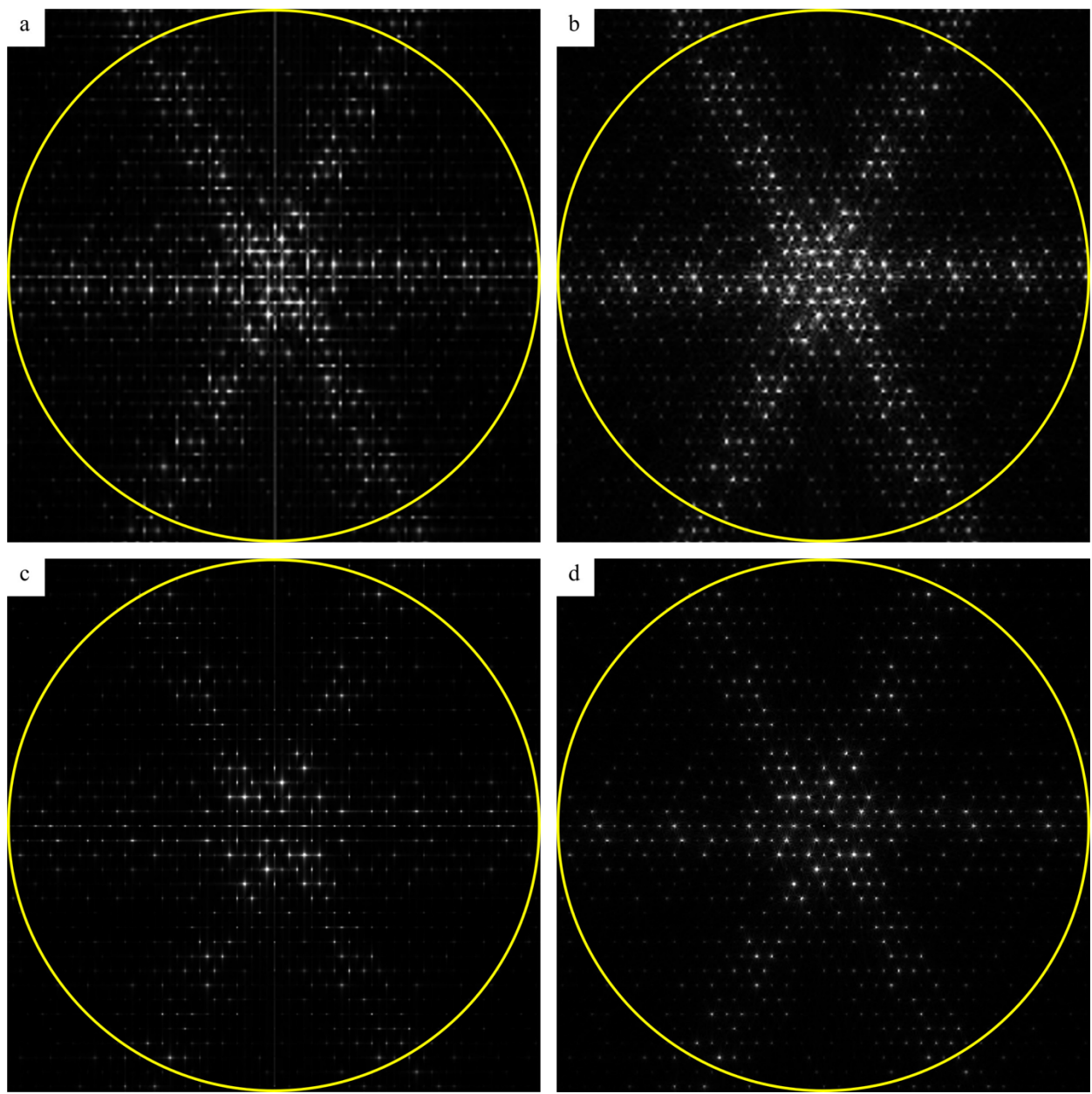

**Figure 6:** Amplitude maps of (a) 1024 × 1024 pixel square selection; (b) 1024-pixel diameter circular selection; (c) 2048 × 2048 pixel square selection; and (d) 2048-pixel diameter circular selection of the noise free image. Images are not to scale; the yellow circles in (a) and (b) correspond to 128-pixel radius circles in the original amplitude maps, whereas in (c) and (d) the yellow circles correspond to 256-pixel radius circles. FCs laying outside of these circles were discarded. The pronounced streaking of the peaks in (a) is an artifact of edge discontinuities in the square DFT window. Less-pronounced streaking is visible in (c), while (b) and (d) have virtually no such artifacts because they are greatly reduced by circular DFT windows. The marked increase in the sharpness of peaks in (c) and (d), in comparison to the more smeared-out peaks in (a) and (b), is due to the larger number of periodic repeats in those selections.

Once the DFT has been calculated, CRISP displays the amplitude map, and the user must run the subroutine that detects peaks in the amplitude map corresponding to the structure-bearing FCs and fits a (reciprocal) lattice to them. By default, observed FCs with amplitudes of less than 50 (which implies normalized amplitudes of less than $5 \times 10^{-3}$) are rejected as non-structure-bearing, as mentioned above. In this study, the default setting was used, but the user is able to adjust the minimum acceptable amplitude to as low as 1.

The accuracy of estimated reciprocal lattice parameters, and all subsequent calculations, depends largely on the quality of the amplitude map of the image selection. Just as selection size and type affects the quality of the resulting amplitude map, so too does the generalized noise level in the input image, see Figure 7. A benefit of working in Fourier/reciprocal space is the suppression of non-structure-bearing noise that occurs as a byproduct of calculating the DFT.

Figure 7a depicts the amplitude map of the 1024-pixel diameter circular selection of the image with RGB noise level 1.00. In spite of the presence of the largest amount of added RGB noise, this amplitude map is of comparable quality to that of the same selection of the noise-free image (Figure 6b), which demonstrates how effectively RGB noise is suppressed by taking the DFT. Figure 7b, on the other hand, depicts the 1024-pixel diameter circular selection of the image with RGB noise level 1.00 and 50-pixel spread noise. The marked degradation of features compared to Figures 6b and 7a, especially higher-resolution peaks, is a result of the large amount of added generalized noise that is no longer strictly Gaussian distributed. Clearly, spread noise is suppressed by the DFT less effectively than RGB noise.

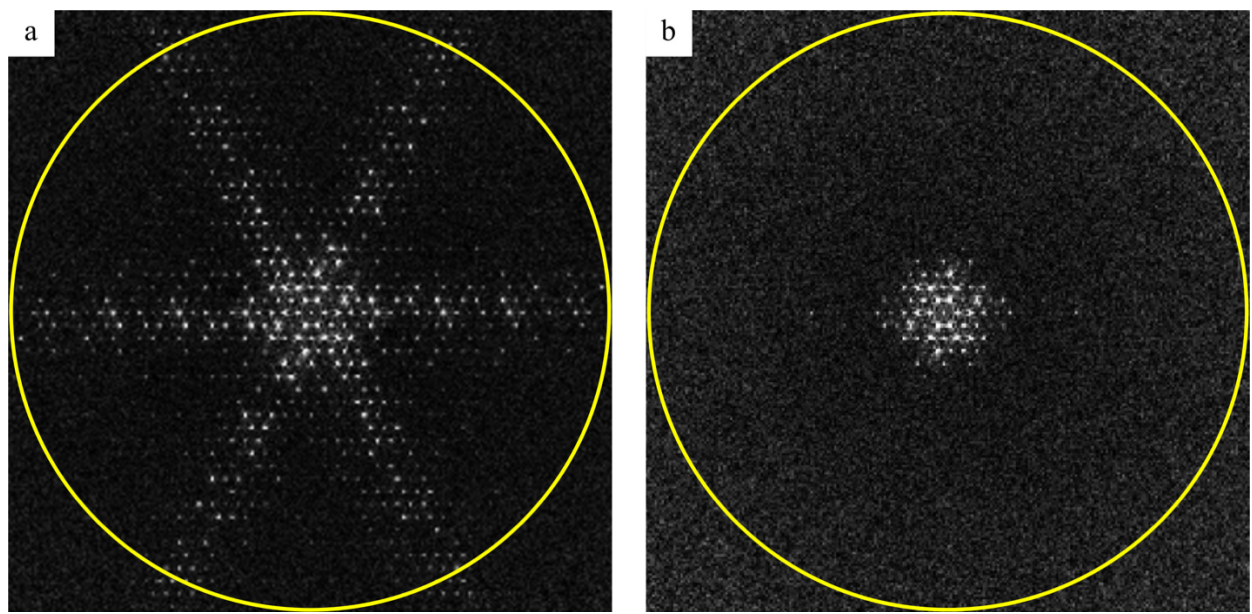

**Figure 7:** Amplitude maps of 1024-pixel diameter circular selections of a) image with RGB noise level 1.00 and b) RGB noise level 1.00 and 50-pixel spread noise. The obvious presence of many high-resolution peaks in a) (even some falling outside the 128-pixel radius cutoff shown in yellow) demonstrates that RGB noise does not present significant challenges to objective crystallographic symmetry classifications and CIP. The drastic degradation in quality of the amplitude map shown in b) is due to the addition of a large amount of non-Gaussian generalized noise, demonstrating the limitations of our method. Note that in b) only lower resolution peaks are clearly visible, meaning only low-resolution 2D periodic structural information is present in the direct space image. This corresponds to the visual appearance of the pinwheel in the right-most column of Figure 5.

The final step in CRISP for our purposes is to perform the phase origin refinement. FC phases, unlike amplitudes, are initially calculated on the basis of an arbitrarily chosen phase origin within the unit cell (that corresponds to the center of the image). The purpose of origin refinement is to systematically locate the phase origin that minimizes the phase residuals for each plane symmetry group that is to be enforced on the input image data, and to re-calculate FC amplitudes and phases, and subsequent FC amplitude and phase residuals, on the basis of this.

After the phase origin refinement has been completed, CRISP displays a data table containing the minimal values of the calculated $F_{res}$, $\phi_{res}$, and $^{Ao}/_{Ae}$ (where applicable) for each of the selected plane symmetry groups, and highlights its overall suggestion for the underlying plane symmetry group of the original image. The user is free to override this suggestion, as the algorithm in CRISP can (and on occasion does) suggest the wrong plane symmetry group [17].

CRISP handles primitive plane symmetry groups differently than centered groups since, as mentioned above, centered groups require a re-indexing of the FCs, and the default setting of the program is to perform origin refinement for primitive groups. For this study, we considered only primitive plane groups because the centered groups *cm* and *c2mm* do not have type-I subgroup-supergroup relations with *p3* or *p6*. As such, numerical results from the centered plane symmetry groups are inconsequential in the context of this study because the underlying plane symmetry and pseudosymmetries of the image are known beforehand (and the focus of the study is to track how the genuine *p2* symmetry and the *p3* and *p6* pseudosymmetries are quantified as a function of increasing general noise levels). Further justification for the exclusion of centered groups from consideration on the basis of our results is provided in the discussion (section 8) below.

At this point in CRISP, the user may save *.hka files for each enforced plane symmetry group image model of the input data. These contain a list of all indexed FCs, observed and symmetrized, with columns corresponding to observed and symmetrized FC amplitudes and phases. CRISP accounts for a total of 21 settings for the 17 plane symmetry groups: two each for *pm* (*p1m1* and *p11m* in full Hermann-Mauguin notation), *pg* (*p1g1* and *p11g*), *cm* (*c1m1* and *c11m*), and *p2mg* (*p2mg* and *p2gm*), because there are two distinct ways of inserting mirror/glide lines into the unit cell for these plane symmetry groups. Since centered groups were ignored and the *.hka file for *p1* cannot be used to calculate meaningful SSRs, each selection from each of the test images yielded 17 *.hka files to be imported into our MATLAB script.

Numerical values from the *.hka files, are handled by our script as double precision numbers. These are used to calculate equations (30) and (31) using normalized observed and symmetrized amplitudes given by (32) and (33), as well as traditional CRISP-like amplitude and phase residuals (27) and (28) for comparison. Once finished, the script prompts the user to export a table of results to a *.xls (Microsoft Excel) file.

The results contained in the *.xls files output from our script (each of which corresponds to a different image selection type and noise level) were used to estimate the K-L-best model of the specified image selection in Microsoft Excel. G-AICs and associated evidence ratios and G-AWs were then also calculated using Microsoft Excel for the 17 plane symmetry group settings considered in each *.xls file.

G-AIC values, evidence ratios, and G-AWs for Laue classes were not considered in this work because the noise-free image and all noisy images feature plane symmetry group *p2* by design, so that their amplitude maps always feature a two-fold rotational symmetry about the central pixel and belong to Laue class *2*. By a version of Friedel's law that states that the amplitude maps of Fourier transforms are always centrosymmetric, this is the minimum point symmetry any 2D amplitude map can have. As a result, the observed FC amplitudes for *p2* are effectively already symmetrized, meaning for that plane symmetry group equations (27) and (31) reduce to zero, and our information theory-based classification cannot be applied in a meaningful way.

*7.2 Numerical classification results*

Due to the high volume of the numerical analysis results to be presented, we restrict our focus here and in the discussion section that follows primarily to full results for all four selections of both the noise free image and the image with RGB noise level 1.00 combined with 50-pixel horizontal and vertical spread noise, shown in Tables 1-8 below. The remaining numerical results for other selection sizes, selection types, and generalized noise levels are tabulated in Appendix 3.

Table 1: Results for **1024-pixel square** selection of **noise-free** image

| Geometric Model | $J_{FC}$ | $F_{res}$ | $\phi_{res}$ (°) | CRISP's Suggestion | Estimated K-L-Best | G-AIC | G-AW full set (%) | G-AW p2, p3, p6 (%) | $E_{best, j}$ | $E_{p6, p3}$ | $N$ | $\hat{\epsilon}^2$ |
|---|---|---|---|---|---|---|---|---|---|---|---|---|
| **p2** | **0.0406** | - | **4.17** | | | **0.122** | **20.3** | **46.1** | - | | **280** | |
| p1m1 | 4.32 | 39.9 | 17.3 | | | 4.40 | 2.40 | - | 8.48 | | 280 | |
| p11m | 2.67 | 39.9 | 8.40 | | | 2.75 | 5.47 | - | 3.72 | | 280 | |
| p1g1 | 2.68 | 39.9 | 10.4 | | | 2.76 | 5.43 | - | 3.74 | | 280 | |
| p11g | 3.68 | 39.9 | 16.2 | | | 3.76 | 3.30 | - | 6.16 | | 280 | |
| p2mm | 2.92 | 39.9 | 14.5 | | | 2.96 | 4.93 | - | 4.13 | | 280 | |
| p2mg | 2.71 | 39.9 | 14.4 | | | 2.75 | 5.47 | - | 3.72 | | 280 | |
| p2gm | 2.71 | 39.9 | 14.4 | | | 2.75 | 5.47 | - | 3.72 | | 280 | |
| p2gg | 2.92 | 39.9 | 14.5 | p6 | p2 | 2.96 | 4.93 | - | 4.13 | 2.25 | 280 | 2.90E-04 |
| p4 | 3.99 | 44.2 | 21.2 | | | 4.03 | 2.89 | - | 7.04 | | 280 | |
| p4mm | 4.85 | 52.2 | 32.3 | | | 4.87 | 1.89 | - | 10.7 | | 280 | |
| p4gm | 4.85 | 52.2 | 32.3 | | | 4.87 | 1.89 | - | 10.7 | | 280 | |
| **p3** | **1.14** | **30.4** | **15.6** | | | **1.20** | **11.9** | **26.9** | **1.71** | | **279** | |
| p3m1 | 2.78 | 38.3 | 15.5 | | | 2.81 | 5.32 | - | 3.83 | | 280 | |
| p31m | 5.90 | 38.3 | 38.4 | | | 5.93 | 1.12 | - | 18.2 | | 280 | |
| **p6** | **1.16** | **30.4** | **17.1** | | | **1.19** | **11.9** | **27.0** | **1.70** | | **279** | |
| p6mm | 2.78 | 38.3 | 25.6 | | | 2.80 | 5.34 | - | 3.81 | | 280 | |

Table 2: Results for **1024-pixel** diameter **circular** selection of **noise-free** image

| Geometric Model | $J_{FC}$ | $F_{res}$ | $\phi_{res}$ (°) | CRISP's Suggestion | Estimated K-L-Best | G-AIC | G-AW full set (%) | G-AW p2, p3, p6 (%) | $E_{best, j}$ | $E_{p6, p3}$ | $N$ | $\hat{\epsilon}^2$ |
|---|---|---|---|---|---|---|---|---|---|---|---|---|
| **p2** | 7.61E-03 | - | 2.26 | | | 0.0228 | 16.5 | 46.7 | - | | 258 | |
| p1m1 | 2.10 | 33.3 | 6.63 | | | 2.12 | 5.80 | - | 2.85 | | 258 | |
| p11m | 2.10 | 33.3 | 7.26 | | | 2.11 | 5.81 | - | 2.84 | | 258 | |
| p1g1 | 2.10 | 33.3 | 7.68 | | | 2.12 | 5.80 | - | 2.85 | | 258 | |
| p11g | 2.25 | 33.3 | 8.19 | | | 2.26 | 5.40 | - | 3.06 | | 258 | |
| p2mm | 2.11 | 33.3 | 13.9 | | | 2.11 | 5.81 | - | 2.85 | | 258 | |
| p2mg | 2.25 | 33.3 | 15.2 | | | 2.26 | 5.40 | - | 3.06 | | 258 | |
| p2gm | 2.25 | 33.3 | 15.2 | | | 2.26 | 5.40 | - | 3.06 | | 258 | |
| p2gg | 2.11 | 33.3 | 13.9 | p6 | p2 | 2.11 | 5.81 | - | 2.85 | 1.87 | 258 | 5.90E-05 |
| p4 | 3.54 | 35.7 | 22.0 | | | 3.55 | 2.83 | - | 5.83 | | 258 | |
| p4mm | 3.81 | 45.9 | 29.6 | | | 3.82 | 2.48 | - | 6.66 | | 257 | |
| p4gm | 3.81 | 45.9 | 29.6 | | | 3.82 | 2.48 | - | 6.66 | | 257 | |
| **p3** | **1.12** | **30.8** | **14.6** | | | **1.13** | **9.49** | **26.8** | **1.74** | | **257** | |
| p3m1 | 2.40 | 39.7 | 16.5 | | | 2.41 | 5.01 | - | 3.30 | | 256 | |
| p31m | 4.65 | 39.7 | 39.3 | | | 4.66 | 1.63 | - | 10.1 | | 256 | |
| **p6** | **1.16** | **30.8** | **22.1** | | | **1.16** | **9.36** | **26.5** | **1.77** | | **257** | |
| p6mm | 2.42 | 39.7 | 29.7 | | | 2.42 | 4.98 | - | 3.32 | | 256 | |

Table 3: Results for **2048-pixel square** selection of **noise-free** image

| Geometric Model | $J_{FC}$ | $F_{res}$ | $\phi_{res}$ (°) | CRISP's Suggestion | Estimated K-L-Best | G-AIC | G-AW full set (%) | G-AW p2, p3, p6 (%) | $E_{best, j}$ | $E_{p6, p3}$ | $N$ | $\hat{\epsilon}^2$ |
|---|---|---|---|---|---|---|---|---|---|---|---|---|
| **p2** | 0.0214 | - | 2.98 | | | 0.0641 | 23.5 | 49.8 | - | | 277 | |
| p1m1 | 3.92 | 39.6 | 16.3 | | | 3.96 | 3.35 | - | 7.02 | | 277 | |
| p11m | 3.21 | 39.6 | 9.49 | | | 3.25 | 4.78 | - | 4.91 | | 277 | |
| p1g1 | 3.20 | 39.6 | 7.00 | | | 3.24 | 4.79 | - | 4.90 | | 277 | |
| p11g | 4.13 | 39.6 | 15.9 | | | 4.17 | 3.02 | - | 7.79 | | 277 | |
| p2mm | 3.33 | 39.6 | 13.6 | | | 3.35 | 4.53 | - | 5.18 | | 277 | |
| p2mg | 3.22 | 39.6 | 13.6 | | | 3.24 | 4.80 | - | 4.90 | | 277 | |
| p2gm | 3.22 | 39.6 | 13.6 | | | 3.24 | 4.80 | - | 4.90 | | 277 | |
| p2gg | 3.33 | 39.6 | 13.6 | p6 | p2 | 3.35 | 4.53 | - | 5.18 | 2.66 | 277 | 1.54E-04 |
| p4 | 4.72 | 43.0 | 20.7 | | | 4.74 | 2.26 | - | 10.4 | | 277 | |
| p4mm | 5.82 | 50.7 | 33.1 | | | 5.83 | 1.31 | - | 17.9 | | 277 | |
| p4gm | 5.82 | 50.7 | 33.1 | | | 5.83 | 1.31 | - | 17.9 | | 277 | |
| **p3** | **1.41** | **29.9** | **15.8** | | | **1.44** | **11.8** | **25.1** | **1.99** | | **276** | |
| p3m1 | 3.38 | 37.2 | 14.5 | | | 3.39 | 4.45 | - | 5.28 | | 276 | |
| p31m | 3.38 | 37.2 | 25.9 | | | 3.40 | 4.44 | - | 5.29 | | 276 | |
| **p6** | **1.42** | **29.9** | **16.9** | | | **1.43** | **11.9** | **25.1** | **1.98** | | **276** | |
| p6mm | 3.38 | 37.2 | 25.8 | | | 3.39 | 4.46 | - | 5.27 | | 276 | |

Table 4: Results for **2048-pixel** diameter **circular** selection of **noise-free** image

| Geometric Model | $J_{FC}$ | $F_{res}$ | $\phi_{res}$ (°) | CRISP's Suggestion | Estimated K-L-Best | G-AIC | G-AW full set (%) | G-AW p2, p3, p6 (%) | $E_{best, j}$ | $E_{p6, p3}$ | N | $\hat{\epsilon}^2$ |
|---|---|---|---|---|---|---|---|---|---|---|---|---|
| **p2** | 2.06E-03 | - | 1.15 | | | 6.17E-03 | 16.0 | 46.7 | - | | 263 | |
| p1m1 | 2.13 | 34.6 | 7.34 | | | 2.14 | 5.51 | - | 2.90 | | 263 | |
| p11m | 2.13 | 34.6 | 7.59 | | | 2.14 | 5.52 | - | 2.90 | | 263 | |
| p1g1 | 2.24 | 34.6 | 8.41 | | | 2.25 | 5.22 | - | 3.06 | | 263 | |
| p11g | 2.24 | 34.6 | 7.61 | | | 2.25 | 5.22 | - | 3.07 | | 263 | |
| p2mm | 2.14 | 34.6 | 13.0 | | | 2.14 | 5.51 | - | 2.90 | | 263 | |
| p2mg | 2.25 | 34.6 | 14.3 | | | 2.25 | 5.22 | - | 3.07 | | 263 | |
| p2gm | 2.25 | 34.6 | 14.3 | | | 2.25 | 5.22 | - | 3.07 | | 263 | |
| p2gg | 2.14 | 34.6 | 13.0 | p6 | p2 | 2.14 | 5.51 | - | 2.90 | 1.81 | 263 | 1.56E-05 |
| p4 | 3.49 | 36.4 | 21.4 | | | 3.49 | 2.81 | - | 5.70 | | 263 | |
| p4mm | 3.75 | 46.8 | 29.4 | | | 3.75 | 2.46 | - | 6.50 | | 262 | |
| p4gm | 3.75 | 46.8 | 29.4 | | | 3.75 | 2.46 | - | 6.50 | | 262 | |
| **p3** | **1.12** | **31.1** | **11.8** | | | **1.13** | **9.14** | **26.7** | **1.75** | | **261** | |
| p3m1 | 2.32 | 40.3 | 13.9 | | | 2.32 | 5.04 | - | 3.18 | | 262 | |
| p31m | 2.33 | 40.3 | 27.8 | | | 2.33 | 5.01 | - | 3.19 | | 262 | |
| **p6** | **1.13** | **31.1** | **21.7** | | | **1.13** | **9.12** | **26.6** | **1.75** | | **261** | |
| p6mm | 2.32 | 40.3 | 28.0 | | | 2.32 | 5.03 | - | 3.18 | | 262 | |

Table 5: Results for **1024-pixel square** selection of image with **RGB noise level 1.00 and 50-pixel spread noise**

| Geometric Model | $J_{FC}$ | $F_{res}$ | $\phi_{res}$ (°) | CRISP's Suggestion | Estimated K-L-Best | G-AIC | G-AW full set (%) | G-AW p2, p3, p6 (%) | $E_{best, j}$ | $E_{p6, p3}$ | N | $\hat{\epsilon}^2$ |
|---|---|---|---|---|---|---|---|---|---|---|---|---|
| **p2** | 0.0582 | - | 15.9 | | | 0.175 | 9.41 | 33.8 | - | | 158 | |
| p1m1 | 0.886 | 38.2 | 8.21 | | | 1.00 | 6.22 | - | 1.51 | | 158 | |
| p11m | 1.64 | 38.2 | 6.12 | | | 1.75 | 4.28 | - | 2.20 | | 158 | |
| p1g1 | 0.886 | 38.2 | 7.60 | | | 1.00 | 6.22 | - | 1.51 | | 158 | |
| p11g | 0.885 | 38.2 | 7.34 | | | 1.00 | 6.23 | - | 1.51 | | 158 | |
| p2mm | 1.68 | 38.2 | 23.3 | | | 1.74 | 4.30 | - | 2.19 | | 158 | |
| p2mg | 0.930 | 38.2 | 21.2 | | | 0.988 | 6.27 | - | 1.50 | | 158 | |
| p2gm | 0.929 | 38.2 | 20.9 | | | 0.987 | 6.27 | - | 1.50 | | 158 | |
| p2gg | 1.68 | 38.2 | 23.5 | p3 | p2 | 1.74 | 4.30 | - | 2.19 | 1.27 | 158 | 7.36E-04 |
| p4 | 1.85 | 40.0 | 26.2 | | | 1.91 | 3.96 | - | 2.38 | | 157 | |
| p4mm | 2.06 | 52.9 | 29.1 | | | 2.09 | 3.61 | - | 2.61 | | 153 | |
| p4gm | 2.06 | 52.9 | 29.1 | | | 2.09 | 3.61 | - | 2.61 | | 153 | |
| **p3** | **0.151** | **28.2** | **9.90** | | | **0.226** | **9.18** | **33.0** | **1.03** | | **152** | |
| p3m1 | 0.657 | 38.1 | 17.4 | | | 0.694 | 7.26 | - | 1.30 | | 151 | |
| p31m | 2.93 | 38.1 | 39.1 | | | 2.96 | 2.33 | - | 4.03 | | 151 | |
| **p6** | **0.172** | **28.2** | **20.4** | | | **0.209** | **9.25** | **33.2** | **1.02** | | **152** | |
| p6mm | 0.669 | 38.1 | 25.9 | | | 0.688 | 7.28 | - | 1.29 | | 151 | |

Table 6: Results for **1024-pixel** diameter **circular** selection of image with **RGB noise level 1.00 and 50-pixel spread noise**

| Geometric Model | $J_{FC}$ | $F_{res}$ | $\phi_{res}$ (°) | CRISP's Suggestion | Estimated K-L-Best | G-AIC | G-AW full set (%) | G-AW p2, p3, p6 (%) | $E_{best,j}$ | $E_{p6,p3}$ | N | $\hat{\epsilon}^2$ |
|---|---|---|---|---|---|---|---|---|---|---|---|---|
| **p2** | 9.36E-02 | - | 17.34 | | | 0.281 | 9.48 | 32.9 | 1.02 | | 164 | |
| p1m1 | 1.42 | 41.1 | 5.30 | | | 1.61 | 4.87 | - | 1.99 | | 164 | |
| p11m | 1.10 | 41.1 | 4.76 | | | 1.28 | 5.74 | - | 1.69 | | 164 | |
| p1g1 | 1.09 | 41.1 | 5.00 | | | 1.28 | 5.75 | - | 1.69 | | 164 | |
| p11g | 1.09 | 41.1 | 6.02 | | | 1.28 | 5.75 | - | 1.69 | | 164 | |
| p2mm | 1.49 | 41.1 | 21.6 | | | 1.58 | 4.94 | - | 1.97 | | 164 | |
| p2mg | 1.16 | 41.1 | 20.8 | | | 1.25 | 5.83 | - | 1.67 | | 164 | |
| p2gm | 1.16 | 41.1 | 20.7 | | | 1.25 | 5.84 | - | 1.66 | | 164 | |
| p2gg | 1.49 | 41.1 | 21.5 | p3 | p6 | 1.58 | 4.95 | - | 1.96 | 1.42 | 164 | 1.14E-03 |
| p4 | 1.65 | 43.5 | 23.7 | | | 1.74 | 4.57 | - | 2.13 | | 164 | |
| p4mm | 2.02 | 54.2 | 29.3 | | | 2.06 | 3.89 | - | 2.50 | | 160 | |
| p4gm | 2.02 | 54.2 | 29.3 | | | 2.06 | 3.89 | - | 2.50 | | 160 | |
| **p3** | 0.129 | 24.1 | 8.91 | | | 0.250 | 9.6 | 33.4 | 1.01 | | 159 | |
| p3m1 | 0.88 | 32.4 | 17.5 | | | 0.940 | 6.82 | - | 1.42 | | 158 | |
| p31m | 3.91 | 32.4 | 38.3 | | | 3.97 | 1.50 | - | 6.49 | | 158 | |
| **p6** | 0.171 | 24.1 | 20.5 | | | 0.232 | 9.7 | 33.7 | - | | 159 | |
| p6mm | 0.902 | 32.4 | 25.9 | | | 0.932 | 6.84 | - | 1.42 | | 158 | |

Table 7: Results for **2048-pixel square** selection of image with **RGB noise level 1.00 and 50-pixel spread noise**

| Geometric Model | $J_{FC}$ | $F_{res}$ | $\phi_{res}$ (°) | CRISP's Suggestion | Estimated K-L-Best | G-AIC | G-AW full set (%) | G-AW p2, p3, p6 (%) | $E_{best,j}$ | $E_{p6,p3}$ | N | $\hat{\epsilon}^2$ |
|---|---|---|---|---|---|---|---|---|---|---|---|---|
| **p2** | 0.0203 | - | 7.48 | | | 0.0608 | 9.78 | 34.0 | - | | 105 | |
| p1m1 | 1.33 | 42.8 | 5.13 | | | 1.37 | 5.08 | - | 1.92 | | 104 | |
| p11m | 1.08 | 42.8 | 6.91 | | | 1.13 | 5.74 | - | 1.70 | | 104 | |
| p1g1 | 1.08 | 42.8 | 6.02 | | | 1.12 | 5.76 | - | 1.70 | | 104 | |
| p11g | 1.07 | 42.8 | 4.04 | | | 1.11 | 5.78 | - | 1.69 | | 104 | |
| p2mm | 1.34 | 42.8 | 11.7 | | | 1.36 | 5.12 | - | 1.91 | | 104 | |
| p2mg | 1.09 | 42.8 | 11.9 | | | 1.11 | 5.79 | - | 1.69 | | 104 | |
| p2gm | 1.09 | 42.8 | 11.9 | | | 1.11 | 5.79 | - | 1.69 | | 104 | |
| p2gg | 1.34 | 42.8 | 11.8 | p3 | p2 | 1.36 | 5.11 | - | 1.91 | 1.46 | 104 | 3.86E-04 |
| p4 | 1.53 | 46.6 | 15.7 | | | 1.55 | 4.65 | - | 2.10 | | 105 | |
| p4mm | 1.91 | 51.4 | 21.3 | | | 1.92 | 3.87 | - | 2.53 | | 100 | |
| p4gm | 1.91 | 51.4 | 21.4 | | | 1.92 | 3.87 | - | 2.53 | | 100 | |
| **p3** | 0.0982 | 18.9 | 5.32 | | | 0.123 | 9.48 | 33.0 | 1.03 | | 98 | |
| p3m1 | 0.861 | 26.4 | 7.24 | | | 0.872 | 6.52 | - | 1.50 | | 88 | |
| p31m | 3.61 | 26.4 | 38.3 | | | 3.62 | 1.65 | - | 5.92 | | 88 | |
| **p6** | 0.108 | 18.9 | 10.2 | | | 0.121 | 9.49 | 33.0 | 1.03 | | 98 | |
| p6mm | 0.867 | 26.4 | 15.6 | | | 0.873 | 6.51 | - | 1.50 | | 88 | |

Table 8: Results for **2048-pixel** diameter **circular** selection of image with **RGB noise level 1.00 and 50-pixel spread noise**

| Geometric Model | $J_{FC}$ | $F_{res}$ | $\phi_{res}$ (°) | CRISP's Suggestion | Estimated K-L-Best | G-AIC | G-AW full set (%) | G-AW p2, p3, p6 (%) | $E_{best, j}$ | $E_{p6, p3}$ | N | $\hat{\epsilon}^2$ |
|---|---|---|---|---|---|---|---|---|---|---|---|---|
| **p2** | **0.0147** | - | **6.95** | | | **0.0441** | **9.45** | **34.0** | - | | **113** | |
| p1m1 | 1.17 | 41.3 | 3.39 | | | 1.19 | 5.32 | - | 1.78 | | 113 | |
| p11m | 0.932 | 41.3 | 4.63 | | | 0.962 | 5.97 | - | 1.58 | | 113 | |
| p1g1 | 0.931 | 41.3 | 4.45 | | | 0.960 | 5.98 | - | 1.58 | | 113 | |
| p11g | 0.930 | 41.3 | 3.59 | | | 0.959 | 5.98 | - | 1.58 | | 113 | |
| p2mm | 1.18 | 41.3 | 10.5 | | | 1.19 | 5.33 | - | 1.77 | | 113 | |
| p2mg | 0.940 | 41.3 | 11.2 | | | 0.955 | 5.99 | - | 1.58 | | 113 | |
| p2gm | 0.940 | 41.3 | 11.3 | | | 0.955 | 5.99 | - | 1.58 | | 113 | |
| p2gg | 1.18 | 41.3 | 10.6 | *p3* | *p2* | 1.19 | 5.33 | - | 1.77 | 1.45 | 113 | 2.60E-04 |
| p4 | 1.54 | 46.9 | 14.5 | | | 1.56 | 4.43 | - | 2.13 | | 113 | |
| p4mm | 1.81 | 50.3 | 20.0 | | | 1.82 | 3.89 | - | 2.43 | | 104 | |
| p4gm | 1.81 | 50.3 | 20.0 | | | 1.82 | 3.89 | - | 2.43 | | 104 | |
| **p3** | **0.0893** | **18.6** | **5.15** | | | **0.107** | **9.16** | **33.0** | **1.03** | | **99** | |
| p3m1 | 0.840 | 25.7 | 8.21 | | | 0.848 | 6.32 | - | 1.49 | | 91 | |
| p31m | 3.75 | 25.7 | 39.4 | | | 3.76 | 1.48 | - | 6.40 | | 91 | |
| **p6** | **0.0945** | **18.6** | **7.87** | | | **0.103** | **9.18** | **33.0** | **1.03** | | **99** | |
| p6mm | 0.842 | 25.7 | 14.7 | | | 0.846 | 6.33 | - | 1.49 | | 91 | |

In order to succinctly represent the most important results from all 56 selections from a total of 14 images, Figures 8 through 11 below are clustered column plots of the calculated G-AWs of *p2*, *p3*, and *p6* for each image, corresponding to the 4 selection sizes and shapes used. These plots illustrate the noise level-dependence of the calculated individual model probabilities. Figure 12 below is a clustered column plot of the estimated squared generalized noise level for each test image and selection type.

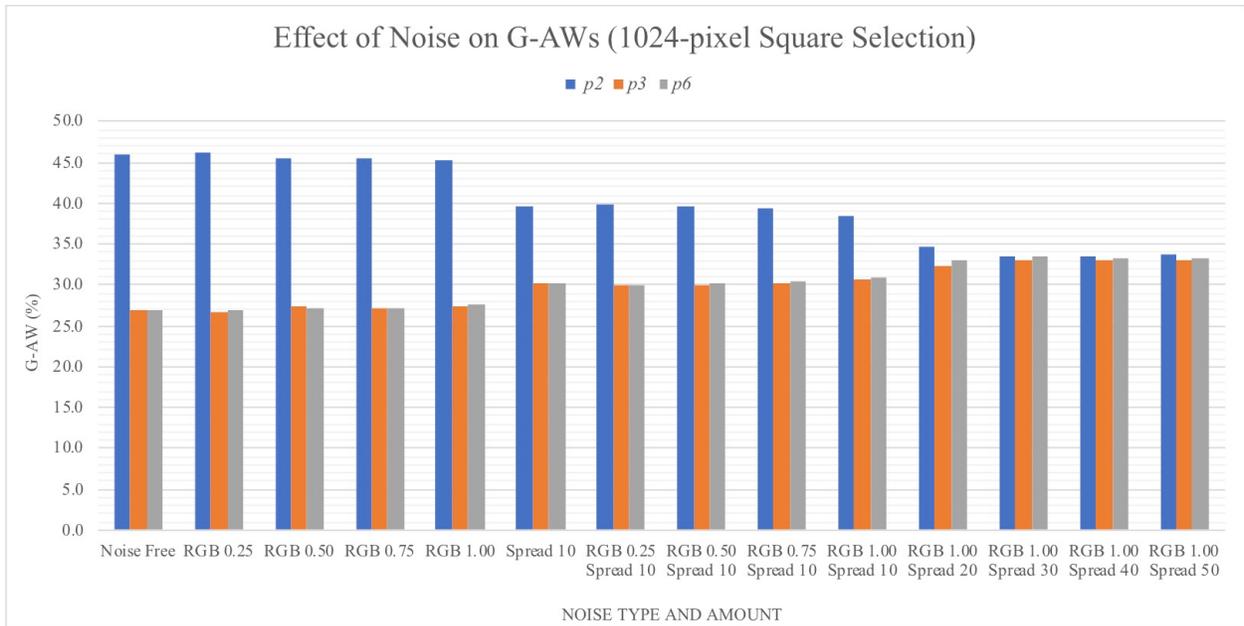

**Figure 8:** clustered column plot of G-AWs for *p2*, *p3*, and *p6* vs. noise type and amount for 1024-pixel square selections.

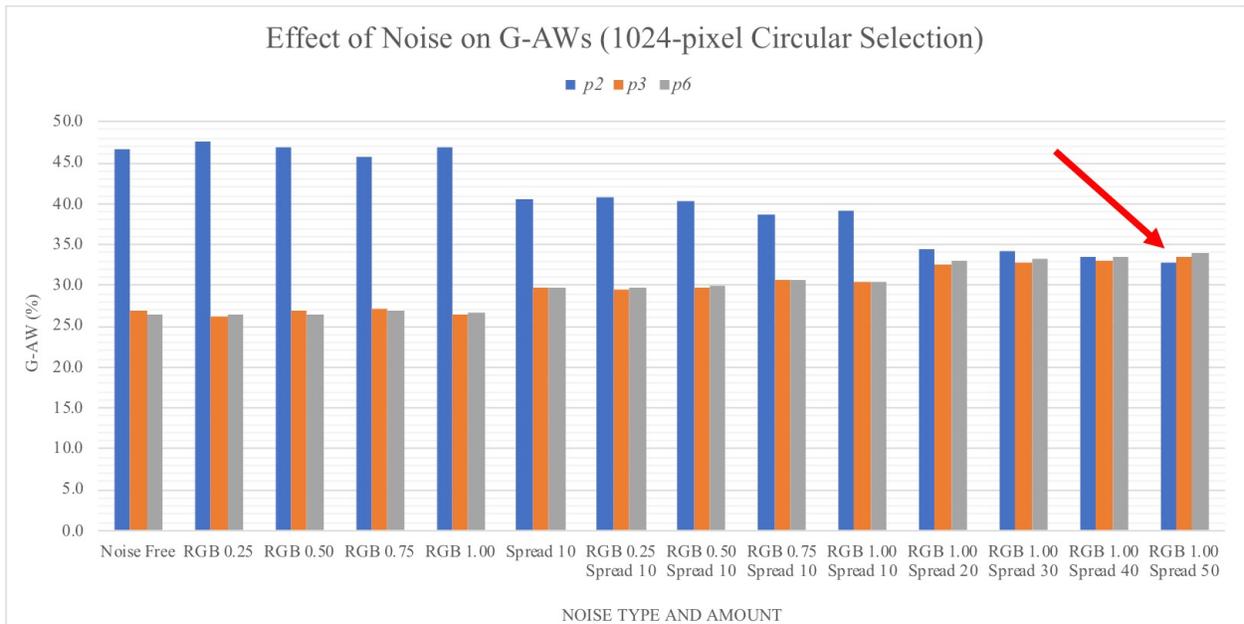

**Figure 9:** clustered column plot of G-AWs for *p2*, *p3*, and *p6* vs. noise type and amount for the 1024-pixel diameter circular selection. The red arrow indicates the sole breakdown of our method, for this type of selection of the image with the most non-Gaussian added noise.

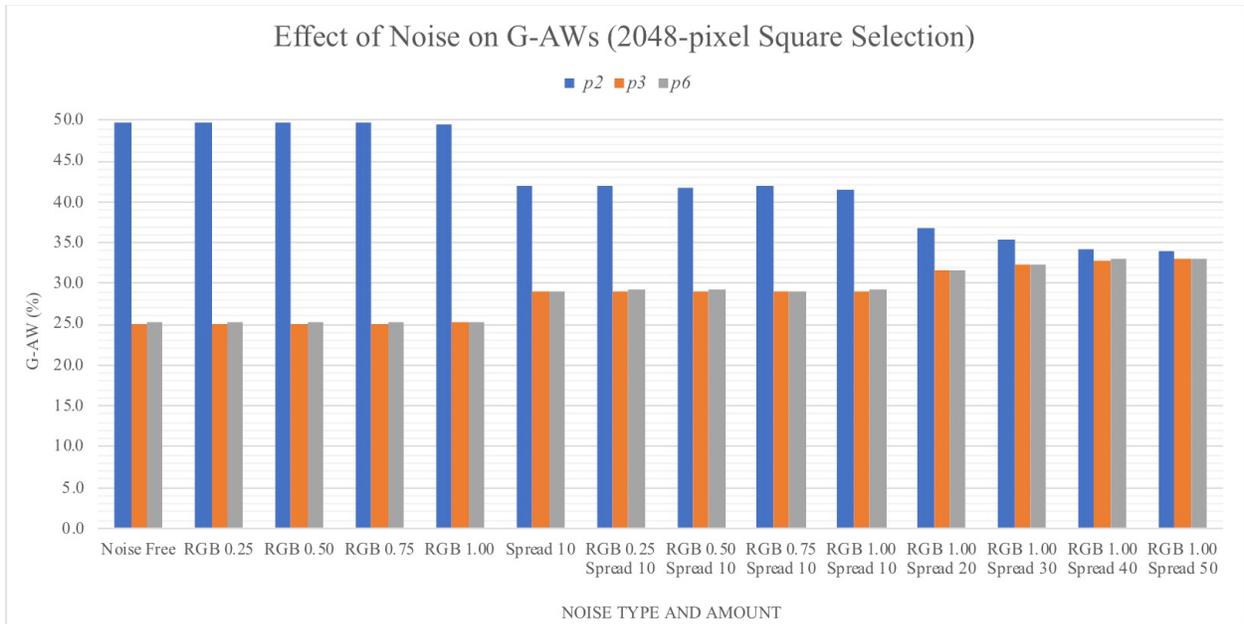

**Figure 10:** clustered column plot of G-AWs for *p2*, *p3*, and *p6* vs. noise type and amount for the 2048-pixel square selection.

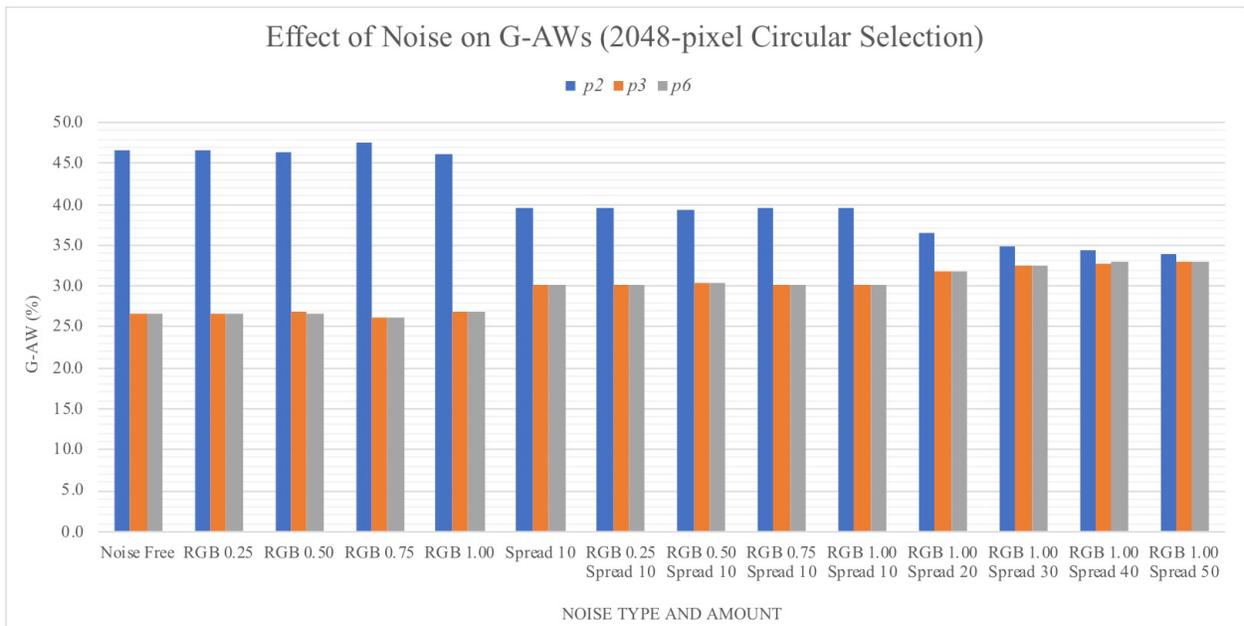

**Figure 11:** clustered column plot of G-AWs for *p2*, *p3*, and *p6* vs. noise type and amount for the 2048-pixel diameter circular selection.

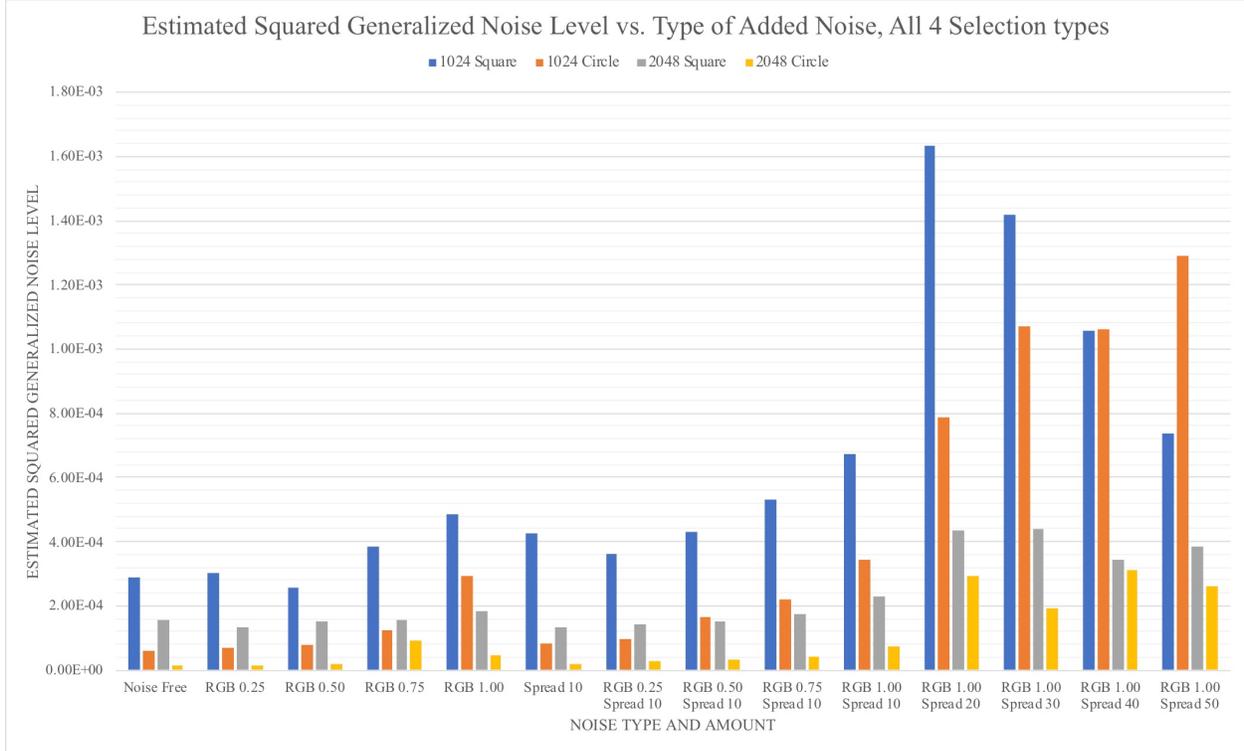

**Figure 12:** Clustered column plot of estimated squared generalized noise level according to equation (17) vs. noise type and amount for each selection size and type.

## 8. Discussion

For both the noise-free image and the noisiest image (with RGB noise level 1.00 and 50-pixel spread noise), regardless of selection size and shape, the *p2* model had the smallest $\hat{J}_{FC}$ for any of the 17 settings of the 14 primitive plane symmetry groups considered, by roughly one to three orders of magnitude. According to our information-theoretic approach to plane symmetry group classifications, this means the next step for each image is to compute inequality (12) for each of the five symmetry hierarchy branches containing *p2* and see if it is satisfied in each case.

In order for inequality (12) to be satisfied, the ratios $\hat{J}_{p2mm}/\hat{J}_{p2}$, $\hat{J}_{p2mg}/\hat{J}_{p2}$ (both settings), $\hat{J}_{p2gg}/\hat{J}_{p2}$, and $\hat{J}_{p4}/\hat{J}_{p2}$ must be less than 2, while $\hat{J}_{p6}/\hat{J}_{p2}$ must be less than 2.33. Because $\hat{J}_{FC}$ is so much smaller for the *p2* model than for the other geometric models in all cases, one can see by inspection of Tables 1-8 (and from further inspection of the remaining 48 data tables in Appendix 3) that inequality (12) is fulfilled only in the case of the 1024-pixel circular selection of the test image with RGB noise level 1.00 in addition to 50-pixel horizontal and vertical spread noise, where it is fulfilled only for $\hat{J}_{p6}/\hat{J}_{p2}$ (see Table 6 above). As already mentioned above, this is the smallest selection (i.e. only approximately 23 unit cell repeats) of the test image with the most added noise.

According to our approach, this means that the *p2* model is the estimated K-L-best geometric model for the test images in 55 of the 56 image selections we processed. This is in contrast to the suggested classification by CRISP, which suggested the *p6* model was the correct geometric model of the underlying plane symmetry for the selections of the noise free image, while for the noisiest image CRISP suggested the *p3* model for all selections types, as shown in column 5 of the Tables 1-8.

Based on this, we estimated squared noise for the estimated K-L-best geometric model (either the *p2* model or the *p6* model) according to equation (17) for all selection shapes and sizes, and used this to calculate G-AICs according to equation (2) for all considered plane symmetry models. The estimated squared generalized noise level for each selection is given in column 13 of Tables 1-8 (and the remaining data tables in Appendix 3). Estimated generalized noise according to equation (17) is understandably larger for selections from the noisiest test image than it is for selections from the noise-free test image by roughly one to two orders of magnitude. Additionally, circular selections have smaller estimated generalized noise than square selections of the same width, likely due to the absence of streaking artifacts in the amplitude maps of circular selections.

An ideal *p2* image processed with an ideal algorithm would not produce a $\hat{J}$ residual for *p2*, and as such would feature zero generalized noise. Note that, for our non-ideal test images and algorithms, even the noise-free image has a non-zero generalized noise term that was estimated on the basis of the K-L-best image model with *p2* symmetry. This is probably to a large extent a result of the particulars of the algorithms we used, including the manner in which CRISP extracts structure-bearing FCs from the image in its default setting, and the manner in which our script handles double-precision numbers. It remains unknown how far the reconstructed Islamic building ornament image in [9] that served as basis of all our test images deviates from a strict *p2* plane symmetry as one would need some algorithm to estimate this and no algorithm is perfect in accordance with Kanatani's dictum. There can also be contributions to the generalized noise by the algorithmic stitching and noise additions procedures that we employed.

Because every path upward from the *p2* model in the hierarchy tree of Figure 1 is blocked in all but one case, no decision in favor of more-symmetric models can be made in the remaining cases, and as such ad-hoc confidence levels could not be calculated for 55 of 56 selections, as they are meaningless in this situation. The decision in favor of *p6* over *p2* for the 1024-pixel circular selection of the image with the most added spread noise has an ad-hoc confidence level of 34.5%. For the identification of plane symmetry group *p6* as the K-L-best group, it must also be preferred over its other maximal translationengleiche subgroup, *p3*. This is indeed so, and the corresponding confidence level is 32.9%.

G-AWs were calculated for both the full set of 17 geometric models, as well as for the subset of models containing those for plane symmetry groups *p2*, *p3*, and *p6*, as shown in columns 8 and 9, respectively, of the data tables. Considering a smaller subset of geometric models does not alter the proportionality of the G-AWs of models in the subset, it simply re-distributes the total probability among them. In all cases, with the exception of the 1024-pixel circular selection of the noisiest image, the model corresponding to plane symmetry group *p2* had the largest G-AW values, and therefore the highest probability of being the K-L-best model of the underlying plane symmetry of the image in question.

For each selection, G-AIC values were also used to calculate evidence ratios of the K-L-best model to each of the 16 other geometric models in the full model set, as well as the evidence ratio of the *p6* model to the *p3* model, shown in columns 10 and 11 of the data tables. The evidence ratios of the *p2* model with respect to both the *p3* and *p6* models range from 1.25 for the 1024-pixel square selection of the image with RGB noise level 1.00 and 10-pixel spread noise, to 1.98 for the 2048-pixel square selection of the noise-free image. The 3- and 6-fold pseudosymmetries are, therefore, rather pronounced as one would expect from the visual inspection of Figs. 4 and 5.

The largest evidence ratio was 18.2 for the *p2* model with respect to the *p31m* model in the 1024-pixel square selection of the noise-free image. The evidence ratio of *p6* to *p3* ranges from 1.42 for the 1024-pixel circular selection of the noisiest image (Table 6) to 2.66 for the 2048-pixel square selection of the noise-free image (Table 3). Whereas these values indicate the evidence in favor of *p2* over *p3* or *p6* is not particularly strong because both pseudosymmetries are very pronounced, it must be borne in mind that a comparatively large G-AW value for a given plane symmetry group is the most significant indicator of that group being the correct model of the plane symmetry present in an image.

The model-specific number of structure-bearing complex FCs $N$ for each model is given in column 12 of the data tables. It is of course preferable for all geometric models to be for the same number of data points, and the derivation of a more general form of inequality (12) should not be taken as an opportunity to compare two geometric models that are for wildly different numbers of data points. If $N$ is not constant for all geometric models, it should at least not vary over a large range. In most cases, the range over which $N$ varied for a given image selection was less than 10; in the worst case the range was 26 and in two separate cases, $N$ was constant for all models.

The clustered column plots of the G-AWs for *p2*, *p3*, and *p6* versus noise type and amount for each image selection shown in Figures 8 to 11 exhibit a number of striking features. First, G-AWs for the models corresponding to plane symmetry groups *p2*, *p3*, and *p6* do not appear to be affected in an appreciable way by the addition of increasing amounts of RGB noise. This is as one would expect because RGB noise is exactly Gaussian distributed per the definition of the G-AIC that we used in this study. Notably, for the 1024-pixel diameter circular selection, the model corresponding to plane symmetry group *p2* has a smaller G-AW value for the noise-free image than it does for the image with RGB noise level 1.00.

Secondly, Figures 8 to 11 show that the addition of 10-pixel horizontal and vertical spread noise decreased the G-AW for the *p2* model by roughly 10% (just as it increased those of the *p3* and *p6* models by roughly 5% each), but again there is no appreciable effect due to the addition of RGB noise. As spread noise with increasing horizontal and vertical distances is added, the individual model probability of the *p2* model decreases further, while those for the *p3* and *p6* models increase until they are approximately equal for all three geometric models of the image data. This is again as expected, because the distribution of spread noise is non-Gaussian in nature with respect to our G-AIC definition. It is, therefore, quite impressive how well our method performs if the generalized noise is to a relatively large extent non-Gaussian.

Thirdly, Figures 8 to 11 show that the *p3* and *p6* models have nearly equal G-AWs for every image and selection type. This and the low evidence ratios for the *p2* model to both the *p3* model and the *p6* model, and for the *p6* model to the *p3* model, are consequences of the strong translational and motif-based pseudosymmetries in the test images. The downward trend in G-AWs for the *p2* model as spread noise is added in greater amounts demonstrates how difficult discriminating between pseudosymmetry and genuine symmetry becomes in the presence of large amounts of generalized noise that is only approximately Gaussian distributed.

Finally, when considered together Figures 8 to 11 demonstrate the effects of the different selection sizes and types we considered on the outcome of our approach. In terms of the individual model probability (G-AW) of the *p2* model, the 2048-pixel square selection yielded the best results overall, probably because it contains the largest number of unit cell repeats. The larger two selections yielded better results in the presence of spread noise compared to the smaller two selections. Of the smaller two selections, the 1024-pixel diameter circle exhibited the worst performance with large amounts of spread noise but yielded slightly better results than the 1024-pixel square selection for images with RGB noise only. These results demonstrate that while there is no single answer to the question of what DFT selection shape is optimal for our approach (or any Fourier space approach), a larger number of processed unit cells is certainly better.

Figure 12 is a clustered column plot of the estimated squared generalized noise level vs. type and amount of added noise, where each cluster of four columns (one for each selection size and shape) represents a different type and/or amount of added noise. A striking feature of this graph is that circular selections typically result in smaller estimates of squared generalized noise than square selections of the same width. This trend is violated only for the 1024-pixel width selections of the two noisiest images, and at present we can only speculate why this is the case. These two images feature the largest amounts of spread noise, which is not Gaussian distributed according to the definition of our pixel value G-AIC. The effect of added spread noise in reciprocal space is shown in Fig. 7 and may be described as a "washing out" of the structure bearing FCs so that their number drops significantly in the corresponding data tables. Whereas there were $N = 280$ structure-bearing FCs for most of the models for the noise-free test image for a 1024 × 1024-pixel square selection, there were only 88 to 105 such coefficients for the same selection of the noisiest image. According to equation (17), smaller $N$ results in larger squared generalized noise levels.

The observed trend in Figure 12 of circular selections having smaller estimated squared generalized noise levels contradicts the above-mentioned assertion in the CRISP documentation that there is little difference between the data from both types of selections. At the very least, our results show that circular selections typically facilitate more accurate extraction of structure-bearing FCs by the CRISP program, which manifests as smaller estimated generalized noise levels when these FCs are used in our approach.

The traditional residuals calculated by our program, shown in columns 2 and 3 of the data tables do not exactly represent what is output by CRISP itself (probably due to the fact that we presumably used a different algorithm). The traditional residuals as calculated by CRISP cannot be exported as a file by that program. In most cases, the values calculated by our MATLAB script are approximately equal to what is displayed in the corresponding CRISP window, and the overall trends in the CRISP-like residuals are preserved.

Examining $F_{res}$ and $\phi_{res}$ in columns 2 and 3 of the data tables, it is not readily apparent to the untrained observer how CRISP arrived at its conclusion of either the *p6* model or the *p3* model as the most likely geometric model of the underlying symmetry of each image. While the *p2* model does have the lowest phase residual, the ambiguity caused by the hierarchic nature of plane symmetry groups, and exacerbated by the presence of metric and motif-based pseudosymmetries in addition to various types and amounts of noise, means the algorithm seems to make a judgment based on some subjective rules of thumb, and errs on the side of higher symmetry. Further, CRISP does not provide any manner of quantifying the probability of its suggestion of the underlying plane symmetry in a more or less 2D-periodic image.

The classification results from our approach are interpreted in a much more straightforward manner. Our classification of the underlying plane symmetry of all but one image and image selection as *p2* is inherently probabilistic. To determine how we reached the conclusion that plane symmetry group *p2* is most probably the correct classification, one simply has to examine the G-AWs for each model we considered. Our results are noise-level dependent and by no means definitive; they are subject to change with improved image data and data processing algorithms. For experimental images, improved classifications will always be obtained with more accurate imaging and recording equipment. Also, the analog image data should be represented by a sufficient number of digital image pixels. Moreover, the above classifications made by our approach were obtained in a computationally efficient manner because we deal with a few hundred points in reciprocal space instead of millions of pixels in direct space.

## 9. Summary and Conclusion

While it is impossible to ever *definitively* classify the approximate plane symmetry present in a noisy real-world digital image, we demonstrated here an information theoretic approach that makes such classifications in a *probabilistic* and objective manner, that is at any given time generalized noise level dependent. This approach implements a geometric Akaike information criterion (G-AIC) for the pixel values of digital images, i.e. employs a statistical framework developed for geometric inference from images and requires the usage of a new type of structure-bearing Fourier coefficient residual presented above in order to be used in conjunction with crystallographic image processing (CIP). It incorporates ad hoc-defined confidence levels for decisions between pairs of geometric models corresponding to non-disjoint plane symmetry groups in type-I minimal supergroup—maximal subgroup relationships, as well as evidence ratios and individual model probabilities that sum to 100% for sets of geometric models, disjoint or non-disjoint. The geometric Akaike weights (G-AWs) are conditional on the selected model set but essentially quantify (in percent) both genuine plane symmetries and pseudosymmetries. The underlying concepts of our approach extend readily to other types of crystallographic (and non-crystallographic) symmetry classifications.

The noise-free image used in this study was chosen because of the relatively large number of pixels per unit cell, strong contrast, and sharp edges, which we expected would yield good results. Similarly, RGB and pixel spread noise were chosen because they were the simplest types of noise to add to images in a systematic way with a popular, free image processing program. These two types of noise seem to be particularly advantageous to demonstrate the effectiveness of our approach. As such, we implore researchers to consider confidence levels and G-AWs for several models to identify pseudosymmetries that are often present in an input image.

Our approach produces accurate results even in the presence of relatively large amounts of generalized, approximately Gaussian distributed noise. In 55 of 56 test image selections, our method correctly selected the image model corresponding to plane symmetry group *p2* to be the most-likely geometric model for the 14 test images from which those selections were taken. Our method failed only in the test case containing the most noise, where no human could reasonably be expected to choose the correct plane symmetry group by visual inspection, and with the fewest periodic repeats in the selection, which hinders reliable application of CIP, especially in the presence of large amounts of noise. For all three larger image selections, our method gave the correct result for that particular image as well due to the better noise suppression that comes with processing larger image selections.

Furthermore, we described the underlying principles of Fourier space plane symmetry classifications as part of crystallographic image processing in some detail. We also demonstrated that apprehensions among the computational symmetry community toward wider adoption of Fourier techniques (such as CIP) caused by the precondition of sufficiently many periodic repeats are unfounded. Image stitching with Microsoft's freely available Image Composite Editor software is a simple pre-processing step that would allow members of the computer vision and computational symmetry communities to utilize more powerful and efficient Fourier space methods when analyzing and classifying noisy 2D periodic images. Unfortunately, the stitching will with necessity produce correlated noise that will have different noise reduction properties in comparison to the generalized noise that this paper is mainly concerned with.

The seemingly intractable, decades-old problem of symmetry detection and classification from real-world images stated in the fourth paragraph of the introduction is indeed readily and effectively solvable when approached with the Fourier-space and information-theoretic techniques presented in this paper. Our approach is not compatible with bestowing upon any particular symmetry detection/classification algorithm the honor of being the "best for all images," in any objective sense. On the contrary, our approach recognizes that no such "honor" exists or can ever exist, and demands all symmetry classifications must be quantitative in nature.

Just as no scientific measurement is acceptable without an associated statement of uncertainty, no symmetry classification should be considered acceptable without an associated statement of either the probability of that symmetry classification being correct or the confidence level(s) of the classification over all maximal type-I subgroup(s) that are more symmetric than plane symmetry group *p1*. Similarly, just as scientific hypotheses can be modified or discarded in light of better data or improved data recording and processing techniques, symmetry classifications should not be considered to be "set in stone." These considerations are built-in to our approach, which provides results that are inherently quantitative, probabilistic, and subject to change upon availability of better image data and/or data recording equipment and processing algorithms.

The MATLAB code and the test images of this paper are freely available at https://github.com/nanocrystallography/hkaAIC_Public. In follow up projects, we will improve on the efficiency and accuracy of our code. Such improvements will include calculations of CRISP-like phase residuals with the more computationally-efficient form of (25) derived in Appendix 2, systematically augmented *.hka files from CRISP that allow for the calculation of more accurate sums of squared residuals, and automated estimation of the K-L-best geometric model in a given set.

**Acknowledgements**

This work is the culmination of the first author's Master of Science in Physics project as supported by a teaching assistantship from Portland State University's (PSU's) Department of Physics. The senior author served as supervisor. The current members of PSU's Nano-Crystallography Group, Gabriel Eng, Regan Garner, Sara Holeman, Grayson Kolar, Choomno Moos, and Connor Shu are thanked for careful proof readings of the manuscript. Additional thanks to Regan Garner for creating Figure 1.

**Appendix 1: Derivation of Generalized Inequalities for Pairwise Comparisons of non-Disjoint Geometric Models with Model-Specific Numbers of Data Points**

The G-AIC of equation (1) assumes a constant number of data points *N*. If this constraint is relaxed, the *N* term in this equation becomes model-specific, i.e. picks up an index *j*:

$$G\text{-}AIC(S_j) = \hat{J}_j + 2(d_j N_j + n_j)\hat{\epsilon}^2 \tag{A1}.$$

A more constrained geometric model $S_m$ is, as before, preferred over a less-constrained model $S_l$ if inequality (2) is satisfied. Stated explicitly, this condition is:

$$\hat{J}_m + 2(d_m N_m + n_m)\hat{\epsilon}^2 < \hat{J}_l + 2(d_l N_l + n_l)\hat{\epsilon}^2 \tag{A2},$$

which reduces algebraically to a more general analog of inequality (4):

$$\frac{\hat{J}_m}{\hat{J}_l} < 1 + \frac{[2(d_l N_l - d_m N_m) + 2(n_l - n_m)]}{\hat{J}_l} \hat{\epsilon}^2 \tag{A3}.$$

Just as before, an unbiased estimator for squared generalized noise can be obtained as:

$$\hat{\epsilon}^2 = \frac{\hat{J}_l}{r_l N_l - n_l} \tag{A4},$$

which is just equation (5), but with model specific $N_l$ in place of a constant $N$. Substituting the right-hand side of (A4) into (A3) yields a more general analog to (6):

$$\frac{\hat{J}_m}{\hat{J}_l} < 1 + \frac{[2(d_l N_l - d_m N_m) + 2(n_l - n_m)]}{r_l N_l - n_l} \tag{A5}.$$

Because our model space (points in the Euclidean plane) has dimension zero, $d_m = d_l = 0$, and the first term in the numerator of (A5) vanishes. Furthermore, the $j^{th}$ model has $n_j$ degrees of freedom given by:

$$n_j = \frac{N_j}{k_j} \tag{A6},$$

and the data space for all geometric models being considered is pixel gray level intensity space, which has dimension 1 and co-dimension $r_j = 1$. Taking this into consideration, (A5) can be re-expressed as:

$$\frac{\hat{J}_m}{\hat{J}_l} < 1 + \frac{2(k_m - \frac{N_m}{N_l} k_l)}{k_m (k_l - 1)} \tag{A7},$$

which is just inequality (34), a more general analog to inequality (12) for use when the number of data points $N$ is not constant for all geometric models in the set being considered.

**Appendix 2: Derivation of a More Computationally Efficient Form of Equation (25)**

Consider a set of $n$ ideal symmetry-related-structure-bearing FCs, e.g., the DFT was calculated without errors from an infinitely large, perfectly 2D-periodic image, and the resulting structure-bearing FCs are related by an $n$-fold rotation, where $n = 2, 4,$ or $6$. These FCs will necessarily have equal amplitudes, $|F|$, and phases that begin at some angle $\phi$ and increase by increments of $\pi/n$ for each subsequent symmetry-related FC. Thus, the $j^{th}$ symmetry-related-structure-bearing FC can be expressed as:

$$\Phi_j = |F| \exp\left[i\left(\phi + \frac{(j-1)}{n}\pi\right)\right] \tag{A8}.$$

By (A8), $\Phi_1 = |F|exp[i\phi]$. Keeping this in mind, along with the fact that multiplication of complex numbers is equivalent to a rotation and/or scaling in the complex plane, the $j^{th}$ symmetry-related-structure-bearing FC can be expressed as:

$$\Phi_j = \zeta_n^{j-1}\Phi_1 \quad (A9),$$

where the complex pre-factors $\zeta_n$ are given by:

$$\zeta_n = \begin{cases} -1 & \text{if } n = 2 \\ i & \text{if } n = 4 \\ \frac{1}{2} + \frac{\sqrt{3}}{2}i & \text{if } n = 6 \end{cases} \quad (A10).$$

Note that these pre-factors correspond to different numbers on the complex unit circle, and have amplitude of 1. As such, their presence in (A9) represents a pure rotation of $\Phi_j$ in the complex plane without any concomitant scaling.

Because the observed structure-bearing FCs are always calculated from a real-world (i.e. finite and noisy) image, their amplitudes will not be equal, and their phases will differ by more than an integer multiple of $\pi/n$. The purpose of symmetrization is to impose the restrictions of a particular symmetry on the observed FCs by assuming that symmetry and replacing FCs that should be related by the given symmetry operation with their complex average:

$$\Phi_{sym} = \frac{1}{n}\sum_{j=1}^{n}\Phi_j = \frac{1}{n}\sum_{j=1}^{n}|F|_j \exp[i\phi_j] \quad (A11).$$

The amplitude and phase of the resulting symmetrized FC are obtained from equation (24) and a modified form of equation (25):

$$\phi_{sym} = \tan^{-1}\left[\frac{\sum_{j=1}^{n}|F_j(h\ k)|\sin(\phi_j(h\ k))}{\sum_{j=1}^{n}|F_j(h\ k)|\cos(\phi_j(h\ k))}\right] \quad (A12),$$

respectively. Note the absence of $\sigma_j$ factors from (A12), and that the weighting factors $\eta_j$ from (25) are in (A12) explicitly expressed as the corresponding FC amplitude. The $\sigma_j$ factors arise from the fact that glide lines, when present, alter the complex pre-factors of (A10) by causing an additional rotation of 90° or 180°, and affected FCs must be rotated back in order to obtain correct averaging.

The $j^{th}$ symmetrized symmetry-related-structure-bearing FC can be re-expressed as:

$$\Phi_j = \zeta_n^{j-1}\Phi_{sym} \quad (A13).$$

The purpose of equations (25) and (A12) is to compute the argument function of a sum of several complex numbers. By considering the full complex-valued FCs instead of their real-valued amplitudes and phases, equation (25) can be re-cast as:

$$\phi_{sym} = \arg\left[\sum_{j=1}^{n} \sigma_j \Phi_j\right]$$
(A14).

Using equation (A14) instead of equation (25) to calculate symmetrized phases results in an improvement in computational efficiency of two to four times, depending on the number of observed relations [16].

**Appendix 3: Additional Data Tables**

Table A1: Results for **1024-pixel square** selection of image with **RGB noise level 0.25**

| Geometric Model | $J_{FC}$ | $F_{res}$ | $\phi_{res}$ (°) | CRISP's Suggestion | Estimated K-L-Best | G-AIC | G-AW full set (%) | G-AW p2, p3, p6 (%) | $E_{best, j}$ | $E_{p6, p3}$ | $N$ | $\hat{\epsilon}^2$ |
|---|---|---|---|---|---|---|---|---|---|---|---|---|
| *p2* | 0.0435 | - | 4.41 | | | 0.130 | 20.4 | 46.3 | - | | 286 | |
| *p1m1* | 4.28 | 39.7 | 17.0 | | | 4.36 | 2.46 | - | 8.30 | | 286 | |
| *p11m* | 2.66 | 39.7 | 9.33 | | | 2.75 | 5.51 | - | 3.70 | | 286 | |
| *p1g1* | 2.68 | 39.7 | 10.2 | | | 2.76 | 5.47 | - | 3.73 | | 286 | |
| *p11g* | 3.70 | 39.7 | 16.2 | | | 3.79 | 3.28 | - | 6.22 | | 286 | |
| *p2mm* | 2.94 | 39.7 | 15.0 | | | 2.99 | 4.89 | - | 4.17 | | 286 | |
| *p2mg* | 2.70 | 39.7 | 14.4 | | | 2.74 | 5.52 | - | 3.69 | | 286 | |
| *p2gm* | 2.70 | 39.7 | 14.4 | | | 2.74 | 5.52 | - | 3.69 | | 286 | |
| *p2gg* | 2.94 | 39.7 | 15.0 | **p6** | **p2** | 2.99 | 4.89 | - | 4.17 | 2.22 | 286 | 3.04E-04 |
| *p4* | 3.99 | 44.4 | 21.4 | | | 4.04 | 2.89 | - | 7.06 | | 286 | |
| *p4mm* | 4.86 | 52.2 | 32.4 | | | 4.88 | 1.89 | - | 10.8 | | 286 | |
| *p4gm* | 4.86 | 52.2 | 32.4 | | | 4.88 | 1.89 | - | 10.8 | | 286 | |
| **p3** | **1.17** | **30.6** | **16.7** | | | **1.23** | **11.8** | **26.7** | **1.73** | | **285** | |
| *p3m1* | 2.78 | 38.4 | 15.8 | | | 2.81 | 5.34 | - | 3.82 | | 286 | |
| *p31m* | 5.92 | 38.4 | 38.7 | | | 5.95 | 1.11 | - | 18.3 | | 286 | |
| **p6** | **1.18** | **30.6** | **17.2** | | | **1.21** | **11.9** | **27.0** | **1.72** | | **285** | |
| *p6mm* | 2.80 | 38.4 | 26.1 | | | 2.81 | 5.34 | - | 3.82 | | 286 | |

Table A2: Results for **1024-pixel** diameter **circular** selection of image with **RGB noise level 0.25**

| Geometric Model | $J_{FC}$ | $F_{res}$ | $\phi_{res}$ (°) | CRISP's Suggestion | Estimated K-L-Best | G-AIC | G-AW full set (%) | G-AW p2, p3, p6 (%) | $E_{best,j}$ | $E_{p6, p3}$ | $N$ | $\hat{\epsilon}^2$ |
|---|---|---|---|---|---|---|---|---|---|---|---|---|
| **p2** | 9.24E-03 | - | 2.40 | | | 0.0277 | 21.2 | 47.5 | - | | 273 | |
| p1m1 | 4.31 | 40.7 | 16.7 | | | 4.32 | 2.47 | - | 8.57 | | 273 | |
| p11m | 2.76 | 40.7 | 8.09 | | | 2.78 | 5.35 | - | 3.96 | | 273 | |
| p1g1 | 2.75 | 40.7 | 8.39 | | | 2.77 | 5.39 | - | 3.93 | | 273 | |
| p11g | 3.65 | 40.7 | 15.6 | | | 3.66 | 3.44 | - | 6.16 | | 273 | |
| p2mm | 2.91 | 40.7 | 12.7 | | | 2.92 | 5.00 | - | 4.24 | | 273 | |
| p2mg | 2.77 | 40.7 | 12.5 | | | 2.78 | 5.36 | - | 3.96 | | 273 | |
| p2gm | 2.77 | 40.7 | 12.5 | | | 2.78 | 5.36 | - | 3.96 | | 273 | |
| p2gg | 2.91 | 40.7 | 12.7 | p6 | p2 | 2.92 | 5.00 | - | 4.24 | 2.28 | 273 | 6.77E-05 |
| p4 | 4.01 | 44.6 | 18.8 | | | 4.02 | 2.89 | - | 7.34 | | 273 | |
| p4mm | 4.92 | 52.5 | 30.7 | | | 4.92 | 1.83 | - | 11.6 | | 272 | |
| p4gm | 4.92 | 52.5 | 30.7 | | | 4.92 | 1.83 | - | 11.6 | | 272 | |
| **p3** | **1.22** | **30.6** | **16.6** | | | **1.23** | **11.6** | **26.1** | **1.82** | | **273** | |
| p3m1 | 2.85 | 38.6 | 14.6 | | | 2.86 | 5.16 | - | 4.11 | | 273 | |
| p31m | 5.82 | 38.6 | 37.8 | | | 5.83 | 1.17 | - | 18.2 | | 273 | |
| **p6** | **1.20** | **30.6** | **15.7** | | | **1.21** | **11.8** | **26.4** | **1.80** | | **273** | |
| p6mm | 2.85 | 38.6 | 24.8 | | | 2.86 | 5.15 | - | 4.11 | | 273 | |

Table A3: Results for **2048-pixel square** selection of image with **RGB noise level 0.25**

| Geometric Model | $J_{FC}$ | $F_{res}$ | $\phi_{res}$ (°) | CRISP's Suggestion | Estimated K-L-Best | G-AIC | G-AW full set (%) | G-AW p2, p3, p6 (%) | $E_{best,j}$ | $E_{p6, p3}$ | $N$ | $\hat{\epsilon}^2$ |
|---|---|---|---|---|---|---|---|---|---|---|---|---|
| **p2** | 0.0188 | - | 2.92 | | | 0.0565 | 23.3 | 49.8 | - | | 278 | |
| p1m1 | 3.88 | 39.5 | 16.3 | | | 3.91 | 3.39 | - | 6.88 | | 278 | |
| p11m | 3.16 | 39.5 | 6.83 | | | 3.19 | 4.85 | - | 4.80 | | 278 | |
| p1g1 | 3.15 | 39.5 | 6.78 | | | 3.19 | 4.86 | - | 4.79 | | 278 | |
| p11g | 4.11 | 39.5 | 16.0 | | | 4.15 | 3.01 | - | 7.72 | | 278 | |
| p2mm | 3.32 | 39.5 | 13.7 | | | 3.34 | 4.51 | - | 5.17 | | 278 | |
| p2mg | 3.17 | 39.5 | 13.6 | | | 3.19 | 4.86 | - | 4.79 | | 278 | |
| p2gm | 3.17 | 39.5 | 13.6 | | | 3.19 | 4.86 | - | 4.79 | | 278 | |
| p2gg | 3.32 | 39.5 | 13.7 | p6 | p2 | 3.34 | 4.51 | - | 5.17 | 2.64 | 278 | 1.35E-04 |
| p4 | 4.67 | 43.0 | 20.6 | | | 4.69 | 2.30 | - | 10.1 | | 278 | |
| p4mm | 5.77 | 50.7 | 33.0 | | | 5.78 | 1.33 | - | 17.5 | | 277 | |
| p4gm | 5.77 | 50.7 | 33.0 | | | 5.78 | 1.33 | - | 17.5 | | 277 | |
| **p3** | **1.40** | **30.0** | **15.7** | | | **1.43** | **11.7** | **25.0** | **1.99** | | **277** | |
| p3m1 | 3.35 | 37.3 | 14.5 | | | 3.36 | 4.46 | - | 5.22 | | 277 | |
| p31m | 3.35 | 37.3 | 25.7 | | | 3.37 | 4.45 | - | 5.23 | | 277 | |
| **p6** | **1.41** | **30.0** | **16.7** | | | **1.42** | **11.8** | **25.2** | **1.97** | | **277** | |
| p6mm | 3.35 | 37.3 | 25.6 | | | 3.35 | 4.48 | - | 5.20 | | 277 | |

Table A4: Results for **2048-pixel** diameter **circular** selection of image with **RGB noise level 0.25**

| Geometric Model | $J_{FC}$ | $F_{res}$ | $\phi_{res}$ (°) | CRISP's Suggestion | Estimated K-L-Best | G-AIC | G-AW full set (%) | G-AW p2, p3, p6 (%) | $E_{best, j}$ | $E_{p6, p3}$ | N | $\hat{\epsilon}^2$ |
|---|---|---|---|---|---|---|---|---|---|---|---|---|
| **p2** | **2.07E-03** | - | **1.24** | | | **6.20E-03** | **15.9** | **46.6** | - | | **265** | |
| p1m1 | 2.11 | 34.2 | 7.59 | | | 2.12 | 5.53 | - | 2.87 | | 265 | |
| p11m | 2.11 | 34.2 | 7.02 | | | 2.12 | 5.53 | - | 2.87 | | 265 | |
| p1g1 | 2.22 | 34.2 | 8.49 | | | 2.23 | 5.23 | - | 3.04 | | 265 | |
| p11g | 2.22 | 34.2 | 7.98 | | | 2.23 | 5.23 | - | 3.04 | | 265 | |
| p2mm | 2.11 | 34.2 | 13.2 | | | 2.12 | 5.53 | - | 2.87 | | 265 | |
| p2mg | 2.23 | 34.2 | 14.6 | | | 2.23 | 5.23 | - | 3.04 | | 265 | |
| p2gm | 2.23 | 34.2 | 14.6 | | | 2.23 | 5.23 | - | 3.04 | | 265 | |
| p2gg | 2.11 | 34.2 | 13.2 | p6 | p2 | 2.12 | 5.53 | - | 2.87 | 1.81 | 265 | 1.56E-05 |
| p4 | 3.45 | 36.5 | 21.5 | | | 3.45 | 2.83 | - | 5.60 | | 265 | |
| p4mm | 3.72 | 46.6 | 29.6 | | | 3.72 | 2.48 | - | 6.39 | | 264 | |
| p4gm | 3.72 | 46.6 | 29.6 | | | 3.72 | 2.48 | - | 6.39 | | 264 | |
| **p3** | **1.11** | **31.1** | **11.7** | | | **1.12** | **9.12** | **26.7** | **1.74** | | **263** | |
| p3m1 | 2.31 | 40.1 | 14.2 | | | 2.31 | 5.02 | - | 3.16 | | 265 | |
| p31m | 2.31 | 40.1 | 28.2 | | | 2.31 | 5.01 | - | 3.17 | | 265 | |
| **p6** | **1.12** | **31.1** | **21.6** | | | **1.12** | **9.11** | **26.7** | **1.74** | | **263** | |
| p6mm | 2.31 | 40.1 | 28.4 | | | 2.31 | 5.02 | - | 3.16 | | 265 | |

Table A5: Results for **1024-pixel square** selection of image with **RGB noise level 0.50**

| Geometric Model | $J_{FC}$ | $F_{res}$ | $\phi_{res}$ (°) | CRISP's Suggestion | Estimated K-L-Best | G-AIC | G-AW full set (%) | G-AW p2, p3, p6 (%) | $E_{best, j}$ | $E_{p6, p3}$ | N | $\hat{\epsilon}^2$ |
|---|---|---|---|---|---|---|---|---|---|---|---|---|
| **p2** | **0.0350** | - | **5.00** | | | **0.105** | **15.8** | **45.5** | - | | **272** | |
| p1m1 | 2.04 | 32.9 | 8.69 | | | 2.11 | 5.78 | - | 2.72 | | 272 | |
| p11m | 2.03 | 32.9 | 8.40 | | | 2.10 | 5.82 | - | 2.71 | | 272 | |
| p1g1 | 2.05 | 32.9 | 10.0 | | | 2.12 | 5.76 | - | 2.73 | | 272 | |
| p11g | 2.14 | 32.9 | 9.28 | | | 2.21 | 5.49 | - | 2.87 | | 272 | |
| p2mm | 2.06 | 32.9 | 16.5 | | | 2.09 | 5.83 | - | 2.70 | | 272 | |
| p2mg | 2.16 | 32.9 | 17.0 | | | 2.20 | 5.53 | - | 2.85 | | 272 | |
| p2gm | 2.16 | 32.9 | 17.0 | | | 2.20 | 5.53 | - | 2.85 | | 272 | |
| p2gg | 2.06 | 32.9 | 16.5 | p6 | p2 | 2.09 | 5.83 | - | 2.70 | 1.81 | 272 | 2.57E-04 |
| p4 | 3.51 | 35.1 | 24.9 | | | 3.55 | 2.81 | - | 5.60 | | 272 | |
| p4mm | 3.76 | 45.1 | 32.8 | | | 3.77 | 2.52 | - | 6.26 | | 270 | |
| p4gm | 3.76 | 45.1 | 33.0 | | | 3.77 | 2.52 | - | 6.26 | | 270 | |
| **p3** | **1.08** | **30.1** | **14.7** | | | **1.12** | **9.48** | **27.4** | **1.66** | | **266** | |
| p3m1 | 2.30 | 39.3 | 17.3 | | | 2.32 | 5.20 | - | 3.03 | | 269 | |
| p31m | 4.67 | 39.3 | 38.7 | | | 4.69 | 1.59 | - | 9.91 | | 269 | |
| **p6** | **1.11** | **30.1** | **23.1** | | | **1.14** | **9.41** | **27.2** | **1.67** | | **266** | |
| p6mm | 2.33 | 39.3 | 30.7 | | | 2.34 | 5.16 | - | 3.05 | | 269 | |

Table A6: Results for **1024-pixel** diameter **circular** selection of image with **RGB noise level 0.50**

| Geometric Model | $J_{FC}$ | $F_{res}$ | $\phi_{res}$ (°) | CRISP's Suggestion | Estimated K-L-Best | G-AIC | G-AW full set (%) | G-AW p2, p3, p6 (%) | $E_{best, j}$ | $E_{p6, p3}$ | N | $\hat{\epsilon}^2$ |
|---|---|---|---|---|---|---|---|---|---|---|---|---|
| **p2** | **0.0102** | - | **3.12** | | | **0.0307** | **16.3** | **46.8** | - | | **269** | |
| p1m1 | 2.07 | 32.9 | 7.24 | | | 2.09 | 5.82 | - | 2.80 | | 269 | |
| p11m | 2.06 | 32.9 | 7.41 | | | 2.08 | 5.85 | - | 2.78 | | 269 | |
| p1g1 | 2.07 | 32.9 | 8.72 | | | 2.09 | 5.81 | - | 2.80 | | 269 | |
| p11g | 2.19 | 32.9 | 8.49 | | | 2.21 | 5.48 | - | 2.97 | | 269 | |
| p2mm | 2.07 | 32.9 | 14.7 | | | 2.08 | 5.84 | - | 2.79 | | 269 | |
| p2mg | 2.20 | 32.9 | 15.6 | | | 2.21 | 5.48 | - | 2.97 | | 269 | |
| p2gm | 2.20 | 32.9 | 15.6 | | | 2.21 | 5.48 | - | 2.97 | | 269 | |
| p2gg | 2.07 | 32.9 | 14.7 | p6 | p2 | 2.08 | 5.84 | - | 2.79 | 1.85 | 269 | 7.61E-05 |
| p4 | 3.47 | 35.8 | 22.4 | | | 3.48 | 2.91 | - | 5.60 | | 268 | |
| p4mm | 3.74 | 45.9 | 30.4 | | | 3.74 | 2.54 | - | 6.40 | | 269 | |
| p4gm | 3.74 | 45.9 | 30.4 | | | 3.74 | 2.54 | - | 6.40 | | 269 | |
| **p3** | **1.13** | **31.5** | **14.4** | | | **1.15** | **9.32** | **26.8** | **1.75** | | **265** | |
| p3m1 | 2.39 | 40.2 | 16.0 | | | 2.40 | 4.98 | - | 3.27 | | 265 | |
| p31m | 4.56 | 40.2 | 38.3 | | | 4.57 | 1.68 | - | 9.66 | | 265 | |
| **p6** | **1.17** | **31.5** | **22.1** | | | **1.18** | **9.19** | **26.4** | **1.77** | | **265** | |
| p6mm | 2.41 | 40.2 | 29.3 | | | 2.41 | 4.96 | - | 3.28 | | 265 | |

Table A7: Results for **2048-pixel square** selection of image with **RGB noise level 0.50**

| Geometric Model | $J_{FC}$ | $F_{res}$ | $\phi_{res}$ (°) | CRISP's Suggestion | Estimated K-L-Best | G-AIC | G-AW full set (%) | G-AW p2, p3, p6 (%) | $E_{best, j}$ | $E_{p6, p3}$ | N | $\hat{\epsilon}^2$ |
|---|---|---|---|---|---|---|---|---|---|---|---|---|
| **p2** | **0.0210** | - | **3.12** | | | **0.0630** | **23.1** | **49.7** | - | | **277** | |
| p1m1 | 3.86 | 39.7 | 16.2 | | | 3.91 | 3.38 | - | 6.83 | | 277 | |
| p11m | 3.12 | 39.7 | 6.74 | | | 3.16 | 4.90 | - | 4.72 | | 277 | |
| p1g1 | 3.12 | 39.7 | 6.64 | | | 3.16 | 4.90 | - | 4.71 | | 277 | |
| p11g | 4.08 | 39.7 | 16.0 | | | 4.12 | 3.03 | - | 7.62 | | 277 | |
| p2mm | 3.30 | 39.7 | 13.7 | | | 3.33 | 4.52 | - | 5.11 | | 277 | |
| p2mg | 3.14 | 39.7 | 13.5 | | | 3.16 | 4.91 | - | 4.70 | | 277 | |
| p2gm | 3.14 | 39.7 | 13.5 | | | 3.16 | 4.91 | - | 4.70 | | 277 | |
| p2gg | 3.30 | 39.7 | 13.7 | p6 | p2 | 3.33 | 4.52 | - | 5.11 | 2.60 | 277 | 1.52E-04 |
| p4 | 4.64 | 43.3 | 20.7 | | | 4.66 | 2.31 | - | 9.98 | | 277 | |
| p4mm | 5.72 | 51.1 | 32.6 | | | 5.73 | 1.36 | - | 17.0 | | 277 | |
| p4gm | 5.72 | 51.1 | 32.6 | | | 5.73 | 1.36 | - | 17.0 | | 277 | |
| **p3** | **1.40** | **30.2** | **15.9** | | | **1.43** | **11.7** | **25.1** | **1.98** | | **277** | |
| p3m1 | 3.32 | 37.6 | 14.6 | | | 3.34 | 4.49 | - | 5.14 | | 277 | |
| p31m | 3.33 | 37.6 | 26.1 | | | 3.34 | 4.48 | - | 5.15 | | 277 | |
| **p6** | **1.41** | **30.2** | **16.8** | | | **1.42** | **11.7** | **25.2** | **1.97** | | **277** | |
| p6mm | 3.33 | 37.6 | 26.0 | | | 3.33 | 4.50 | - | 5.13 | | 277 | |

Table A8: Results for **2048-pixel** diameter **circular** selection of image with **RGB noise level 0.50**

| Geometric Model | $J_{FC}$ | $F_{res}$ | $\phi_{res}$ (°) | CRISP's Suggestion | Estimated K-L-Best | G-AIC | G-AW full set (%) | G-AW p2, p3, p6 (%) | $E_{best, j}$ | $E_{p6, p3}$ | $N$ | $\hat{\epsilon}^2$ |
|---|---|---|---|---|---|---|---|---|---|---|---|---|
| **p2** | **2.61E-03** | - | **1.54** | | | **7.83E-03** | **15.8** | **46.5** | - | | **266** | |
| p1m1 | 2.11 | 34.3 | 7.75 | | | 2.12 | 5.51 | - | 2.87 | | 266 | |
| p11m | 2.11 | 34.3 | 8.15 | | | 2.11 | 5.52 | - | 2.87 | | 266 | |
| p1g1 | 2.21 | 34.3 | 8.55 | | | 2.21 | 5.25 | - | 3.01 | | 266 | |
| p11g | 2.21 | 34.3 | 8.06 | | | 2.21 | 5.25 | - | 3.01 | | 266 | |
| p2mm | 2.11 | 34.3 | 13.7 | | | 2.12 | 5.51 | - | 2.87 | | 266 | |
| p2mg | 2.21 | 34.3 | 14.7 | | | 2.22 | 5.25 | - | 3.02 | | 266 | |
| p2gm | 2.21 | 34.3 | 14.7 | | | 2.22 | 5.25 | - | 3.02 | | 266 | |
| p2gg | 2.11 | 34.3 | 13.7 | p6 | p2 | 2.12 | 5.51 | - | 2.87 | 1.80 | 266 | 1.96E-05 |
| p4 | 3.42 | 36.3 | 21.7 | | | 3.43 | 2.86 | - | 5.53 | | 266 | |
| p4mm | 3.69 | 46.7 | 29.8 | | | 3.69 | 2.51 | - | 6.31 | | 265 | |
| p4gm | 3.69 | 46.7 | 29.8 | | | 3.69 | 2.51 | - | 6.31 | | 265 | |
| **p3** | **1.10** | **30.9** | **13.4** | | | **1.11** | **9.13** | **26.8** | **1.73** | | **266** | |
| p3m1 | 2.30 | 40.1 | 14.8 | | | 2.30 | 5.03 | - | 3.15 | | 264 | |
| p31m | 2.30 | 40.1 | 28.4 | | | 2.31 | 5.01 | - | 3.15 | | 264 | |
| **p6** | **1.12** | **30.9** | **22.1** | | | **1.12** | **9.07** | **26.7** | **1.74** | | **266** | |
| p6mm | 2.30 | 40.1 | 28.8 | | | 2.30 | 5.03 | - | 3.15 | | 264 | |

Table A9: Results for **1024-pixel square** selection of image with **RGB noise level 0.75**

| Geometric Model | $J_{FC}$ | $F_{res}$ | $\phi_{res}$ (°) | CRISP's Suggestion | Estimated K-L-Best | G-AIC | G-AW full set (%) | G-AW p2, p3, p6 (%) | $E_{best, j}$ | $E_{p6, p3}$ | $N$ | $\hat{\epsilon}^2$ |
|---|---|---|---|---|---|---|---|---|---|---|---|---|
| **p2** | **0.0556** | - | **6.27** | | | **0.167** | **20.2** | **45.5** | - | | **290** | |
| p1m1 | 4.34 | 39.6 | 17.8 | | | 4.45 | 2.37 | - | 8.51 | | 290 | |
| p11m | 2.65 | 39.6 | 8.69 | | | 2.76 | 5.52 | - | 3.66 | | 290 | |
| p1g1 | 2.68 | 39.6 | 10.0 | | | 2.79 | 5.44 | - | 3.71 | | 290 | |
| p11g | 3.67 | 39.6 | 15.7 | | | 3.78 | 3.32 | - | 6.09 | | 290 | |
| p2mm | 2.99 | 39.6 | 16.5 | | | 3.04 | 4.80 | - | 4.21 | | 290 | |
| p2mg | 2.72 | 39.6 | 15.4 | | | 2.78 | 5.48 | - | 3.69 | | 290 | |
| p2gm | 2.72 | 39.6 | 15.4 | | | 2.78 | 5.48 | - | 3.69 | | 290 | |
| p2gg | 2.99 | 39.6 | 16.5 | p6 | p2 | 3.04 | 4.80 | - | 4.21 | 2.25 | 290 | 3.83E-04 |
| p4 | 4.07 | 43.9 | 23.2 | | | 4.13 | 2.79 | - | 7.24 | | 290 | |
| p4mm | 4.91 | 52.5 | 33.6 | | | 4.93 | 1.86 | - | 10.8 | | 289 | |
| p4gm | 4.91 | 52.5 | 33.6 | | | 4.93 | 1.86 | - | 10.8 | | 289 | |
| **p3** | **1.12** | **31.4** | **13.2** | | | **1.20** | **12.1** | **27.2** | **1.67** | | **284** | |
| p3m1 | 2.78 | 38.8 | 15.9 | | | 2.81 | 5.38 | - | 3.76 | | 288 | |
| p31m | 5.89 | 38.8 | 37.9 | | | 5.93 | 1.13 | - | 17.8 | | 288 | |
| **p6** | **1.16** | **31.4** | **17.3** | | | **1.19** | **12.1** | **27.2** | **1.67** | | **284** | |
| p6mm | 2.79 | 38.8 | 26.3 | | | 2.81 | 5.40 | - | 3.74 | | 288 | |

Table A10: Results for **1024-pixel** diameter **circular** selection of image with **RGB noise level 0.75**

| Geometric Model | $J_{FC}$ | $F_{res}$ | $\phi_{res}$ (°) | CRISP's Suggestion | Estimated K-L-Best | G-AIC | G-AW full set (%) | G-AW p2, p3, p6 (%) | $E_{best, j}$ | $E_{p6, p3}$ | $N$ | $\hat{\epsilon}^2$ |
|---|---|---|---|---|---|---|---|---|---|---|---|---|
| **p2**   | 0.0168 | -    | 4.35 |    |    | 0.0505 | 16.1 | 45.8 | -    |      | 275 |          |
| p1m1     | 2.06   | 32.6 | 7.73 |    |    | 2.09   | 5.81 | -    | 2.77 |      | 275 |          |
| p11m     | 2.05   | 32.6 | 8.53 |    |    | 2.09   | 5.83 | -    | 2.77 |      | 275 |          |
| p1g1     | 2.19   | 32.6 | 8.98 |    |    | 2.22   | 5.45 | -    | 2.96 |      | 275 |          |
| p11g     | 2.19   | 32.6 | 9.36 |    |    | 2.22   | 5.44 | -    | 2.96 |      | 275 |          |
| p2mm     | 2.07   | 32.6 | 15.7 |    |    | 2.09   | 5.82 | -    | 2.77 |      | 275 |          |
| p2mg     | 2.21   | 32.6 | 16.8 |    |    | 2.22   | 5.44 | -    | 2.96 |      | 275 |          |
| p2gm     | 2.21   | 32.6 | 16.8 |    |    | 2.22   | 5.44 | -    | 2.96 |      | 275 |          |
| p2gg     | 2.07   | 32.6 | 15.7 | p6 | p2 | 2.09   | 5.82 | -    | 2.77 | 1.87 | 275 | 1.22E-04 |
| p4       | 3.48   | 35.9 | 23.5 |    |    | 3.49   | 2.88 | -    | 5.60 |      | 275 |          |
| p4mm     | 3.75   | 46.0 | 30.9 |    |    | 3.76   | 2.52 | -    | 6.40 |      | 273 |          |
| p4gm     | 3.75   | 46.0 | 30.9 |    |    | 3.76   | 2.52 | -    | 6.40 |      | 273 |          |
| **p3**   | **1.07** | **30.7** | **14.1** |    |    | **1.09** | **9.57** | **27.2** | **1.68** |      | **268** |          |
| p3m1     | 2.35   | 39.7 | 16.0 |    |    | 2.36   | 5.08 | -    | 3.17 |      | 271 |          |
| p31m     | 4.56   | 39.7 | 38.1 |    |    | 4.57   | 1.68 | -    | 9.58 |      | 271 |          |
| **p6**   | **1.10** | **30.7** | **22.1** |    |    | **1.11** | **9.49** | **27.0** | **1.70** |      | **268** |          |
| p6mm     | 2.36   | 39.7 | 29.8 |    |    | 2.37   | 5.06 | -    | 3.19 |      | 271 |          |

Table A11: Results for **2048-pixel square** selection of image with **RGB noise level 0.75**

| Geometric Model | $J_{FC}$ | $F_{res}$ | $\phi_{res}$ (°) | CRISP's Suggestion | Estimated K-L-Best | G-AIC | G-AW full set (%) | G-AW p2, p3, p6 (%) | $E_{best, j}$ | $E_{p6, p3}$ | $N$ | $\hat{\epsilon}^2$ |
|---|---|---|---|---|---|---|---|---|---|---|---|---|
| **p2**   | 0.0216 | -    | 3.30 |    |    | 0.0649 | 23.2 | 49.8 | -    |      | 275 |          |
| p1m1     | 3.88   | 39.1 | 16.2 |    |    | 3.92   | 3.38 | -    | 6.87 |      | 274 |          |
| p11m     | 3.14   | 39.1 | 6.60 |    |    | 3.18   | 4.90 | -    | 4.74 |      | 274 |          |
| p1g1     | 3.13   | 39.1 | 7.48 |    |    | 3.17   | 4.92 | -    | 4.72 |      | 274 |          |
| p11g     | 4.12   | 39.1 | 15.5 |    |    | 4.16   | 2.99 | -    | 7.77 |      | 274 |          |
| p2mm     | 3.31   | 39.1 | 13.8 |    |    | 3.33   | 4.53 | -    | 5.13 |      | 274 |          |
| p2mg     | 3.15   | 39.1 | 13.5 |    |    | 3.17   | 4.91 | -    | 4.73 |      | 274 |          |
| p2gm     | 3.15   | 39.1 | 13.5 |    |    | 3.17   | 4.91 | -    | 4.73 |      | 274 |          |
| p2gg     | 3.31   | 39.1 | 13.8 | p6 | p2 | 3.33   | 4.53 | -    | 5.13 | 2.64 | 274 | 1.57E-04 |
| p4       | 4.69   | 43.2 | 20.7 |    |    | 4.71   | 2.28 | -    | 10.2 |      | 275 |          |
| p4mm     | 5.78   | 50.7 | 32.9 |    |    | 5.79   | 1.33 | -    | 17.5 |      | 275 |          |
| p4gm     | 5.78   | 50.7 | 32.9 |    |    | 5.79   | 1.33 | -    | 17.5 |      | 275 |          |
| **p3**   | **1.41** | **30.4** | **15.4** |    |    | **1.44** | **11.7** | **25.1** | **1.99** |      | **275** |          |
| p3m1     | 3.36   | 37.6 | 15.2 |    |    | 3.38   | 4.44 | -    | 5.23 |      | 275 |          |
| p31m     | 3.37   | 37.6 | 26.0 |    |    | 3.39   | 4.42 | -    | 5.26 |      | 275 |          |
| **p6**   | **1.42** | **30.4** | **16.9** |    |    | **1.43** | **11.7** | **25.1** | **1.98** |      | **275** |          |
| p6mm     | 3.36   | 37.6 | 25.6 |    |    | 3.37   | 4.46 | -    | 5.21 |      | 275 |          |

Table A12: Results for **2048-pixel** diameter **circular** selection of image with **RGB noise level 0.75**

| Geometric Model | $J_{FC}$ | $F_{res}$ | $\phi_{res}$ (°) | CRISP's Suggestion | Estimated K-L-Best | G-AIC | G-AW full set (%) | G-AW p2, p3, p6 (%) | $E_{best, j}$ | $E_{p6, p3}$ | N | $\hat{\epsilon}^2$ |
|---|---|---|---|---|---|---|---|---|---|---|---|---|
| **p2** | **0.0125** | - | **2.85** | | | **0.0374** | **20.8** | **47.5** | - | | **267** | |
| p1m1 | 4.08 | 39.9 | 17.2 | | | 4.10 | 2.72 | - | 7.64 | | 267 | |
| p11m | 2.82 | 39.9 | 6.67 | | | 2.84 | 5.11 | - | 4.06 | | 267 | |
| p1g1 | 2.80 | 39.9 | 6.31 | | | 2.83 | 5.15 | - | 4.04 | | 267 | |
| p11g | 3.76 | 39.9 | 15.2 | | | 3.79 | 3.19 | - | 6.52 | | 267 | |
| p2mm | 2.97 | 39.9 | 13.3 | | | 2.99 | 4.76 | - | 4.37 | | 267 | |
| p2mg | 2.83 | 39.9 | 12.9 | | | 2.84 | 5.12 | - | 4.06 | | 267 | |
| p2gm | 2.83 | 39.9 | 12.9 | | | 2.84 | 5.12 | - | 4.06 | | 267 | |
| p2gg | 2.97 | 39.9 | 13.3 | p6 | p2 | 2.99 | 4.76 | - | 4.37 | 2.39 | 267 | 9.33E-05 |
| p4 | 4.15 | 43.9 | 19.7 | | | 4.17 | 2.64 | - | 7.88 | | 267 | |
| p4mm | 5.12 | 51.7 | 31.6 | | | 5.12 | 1.63 | - | 12.7 | | 267 | |
| p4gm | 5.12 | 51.7 | 31.6 | | | 5.12 | 1.63 | - | 12.7 | | 267 | |
| **p3** | **1.21** | **30.5** | **13.6** | | | **1.23** | **11.5** | **26.2** | **1.81** | | **267** | |
| p3m1 | 2.96 | 38.2 | 13.6 | | | 2.97 | 4.81 | - | 4.33 | | 267 | |
| p31m | 2.97 | 38.2 | 24.9 | | | 2.97 | 4.79 | - | 4.34 | | 267 | |
| **p6** | **1.21** | **30.5** | **15.4** | | | **1.22** | **11.5** | **26.3** | **1.81** | | **267** | |
| p6mm | 2.96 | 38.2 | 24.7 | | | 2.96 | 4.82 | - | 4.32 | | 267 | |

Table A13: Results for **1024-pixel square** selection of image with **RGB noise level 1.00**

| Geometric Model | $J_{FC}$ | $F_{res}$ | $\phi_{res}$ (°) | CRISP's Suggestion | Estimated K-L-Best | G-AIC | G-AW full set (%) | G-AW p2, p3, p6 (%) | $E_{best, j}$ | $E_{p6, p3}$ | N | $\hat{\epsilon}^2$ |
|---|---|---|---|---|---|---|---|---|---|---|---|---|
| **p2** | **0.0709** | - | **6.87** | | | **0.213** | **19.3** | **45.2** | - | | **292** | |
| p1m1 | 4.22 | 38.4 | 17.9 | | | 4.36 | 2.43 | - | 7.97 | | 292 | |
| p11m | 2.58 | 38.4 | 12.0 | | | 2.73 | 5.50 | - | 3.51 | | 292 | |
| p1g1 | 2.58 | 38.4 | 10.4 | | | 2.72 | 5.52 | - | 3.51 | | 292 | |
| p11g | 3.50 | 38.4 | 16.1 | | | 3.65 | 3.47 | - | 5.57 | | 292 | |
| p2mm | 2.85 | 38.4 | 17.5 | | | 2.92 | 5.00 | - | 3.87 | | 292 | |
| p2mg | 2.62 | 38.4 | 16.3 | | | 2.69 | 5.61 | - | 3.45 | | 292 | |
| p2gm | 2.62 | 38.4 | 16.3 | | | 2.69 | 5.61 | - | 3.45 | | 292 | |
| p2gg | 2.84 | 38.4 | 17.0 | p6 | p2 | 2.91 | 5.01 | - | 3.86 | 2.19 | 292 | 4.86E-04 |
| p4 | 3.87 | 43.0 | 22.0 | | | 3.94 | 3.00 | - | 6.44 | | 292 | |
| p4mm | 4.67 | 51.4 | 32.6 | | | 4.71 | 2.05 | - | 9.45 | | 289 | |
| p4gm | 4.67 | 51.4 | 32.6 | | | 4.71 | 2.05 | - | 9.45 | | 289 | |
| **p3** | **1.13** | **31.0** | **16.2** | | | **1.22** | **11.7** | **27.3** | **1.66** | | **290** | |
| p3m1 | 2.72 | 38.6 | 15.8 | | | 2.77 | 5.38 | - | 3.59 | | 287 | |
| p31m | 5.72 | 38.6 | 37.6 | | | 5.77 | 1.20 | - | 16.1 | | 287 | |
| **p6** | **1.16** | **31.0** | **18.8** | | | **1.20** | **11.8** | **27.5** | **1.64** | | **290** | |
| p6mm | 2.75 | 38.6 | 27.3 | | | 2.77 | 5.38 | - | 3.59 | | 287 | |

Table A14: Results for **1024-pixel** diameter **circular** selection of image with **RGB noise level 1.00**

| Geometric Model | $J_{FC}$ | $F_{res}$ | $\phi_{res}$ (°) | CRISP's Suggestion | Estimated K-L-Best | G-AIC | G-AW full set (%) | G-AW p2, p3, p6 (%) | $E_{best,\,j}$ | $E_{p6,\,p3}$ | N | $\hat{\epsilon}^2$ |
|---|---|---|---|---|---|---|---|---|---|---|---|---|
| **p2** | **0.0419** | - | **5.77** | | | **0.126** | **20.4** | **46.8** | - | | **286** | |
| p1m1 | 4.02 | 40.5 | 15.9 | | | 4.10 | 2.80 | - | 7.29 | | 286 | |
| p11m | 2.73 | 40.5 | 8.56 | | | 2.82 | 5.32 | - | 3.84 | | 286 | |
| p1g1 | 2.71 | 40.5 | 9.19 | | | 2.79 | 5.38 | - | 3.80 | | 286 | |
| p11g | 3.56 | 40.5 | 15.0 | | | 3.65 | 3.51 | - | 5.82 | | 286 | |
| p2mm | 2.89 | 40.5 | 15.3 | | | 2.93 | 5.02 | - | 4.07 | | 286 | |
| p2mg | 2.76 | 40.5 | 14.4 | | | 2.81 | 5.35 | - | 3.82 | | 286 | |
| p2gm | 2.76 | 40.5 | 14.4 | | | 2.81 | 5.35 | - | 3.82 | | 286 | |
| p2gg | 2.89 | 40.5 | 15.3 | p6 | p2 | 2.93 | 5.02 | - | 4.07 | 2.22 | 286 | 2.93E-04 |
| p4 | 3.95 | 43.7 | 21.0 | | | 3.99 | 2.97 | - | 6.89 | | 286 | |
| p4mm | 4.82 | 52.6 | 31.2 | | | 4.84 | 1.94 | - | 10.5 | | 285 | |
| p4gm | 4.82 | 52.6 | 31.2 | | | 4.84 | 1.94 | - | 10.5 | | 285 | |
| **p3** | **1.21** | **31.6** | **16.3** | | | **1.27** | **11.6** | **26.5** | **1.77** | | **283** | |
| p3m1 | 2.82 | 39.1 | 14.3 | | | 2.84 | 5.25 | - | 3.89 | | 281 | |
| p31m | 5.72 | 39.1 | 36.5 | | | 5.75 | 1.23 | - | 16.6 | | 281 | |
| **p6** | **1.22** | **31.6** | **17.5** | | | **1.25** | **11.7** | **26.7** | **1.75** | | **283** | |
| p6mm | 2.83 | 39.1 | 25.6 | | | 2.84 | 5.25 | - | 3.89 | | 281 | |

Table A15: Results for **2048-pixel square** selection of image with **RGB noise level 1.00**

| Geometric Model | $J_{FC}$ | $F_{res}$ | $\phi_{res}$ (°) | CRISP's Suggestion | Estimated K-L-Best | G-AIC | G-AW full set (%) | G-AW p2, p3, p6 (%) | $E_{best,\,j}$ | $E_{p6,\,p3}$ | N | $\hat{\epsilon}^2$ |
|---|---|---|---|---|---|---|---|---|---|---|---|---|
| **p2** | **0.0257** | - | **3.91** | | | **0.0771** | **23.0** | **49.4** | - | | **278** | |
| p1m1 | 3.88 | 39.4 | 16.0 | | | 3.94 | 3.35 | - | 6.88 | | 278 | |
| p11m | 3.14 | 39.4 | 9.10 | | | 3.19 | 4.86 | - | 4.73 | | 278 | |
| p1g1 | 3.12 | 39.4 | 7.85 | | | 3.17 | 4.90 | - | 4.70 | | 278 | |
| p11g | 4.08 | 39.4 | 15.8 | | | 4.13 | 3.03 | - | 7.60 | | 278 | |
| p2mm | 3.29 | 39.4 | 14.2 | | | 3.31 | 4.57 | - | 5.04 | | 278 | |
| p2mg | 3.16 | 39.4 | 14.3 | | | 3.18 | 4.88 | - | 4.72 | | 278 | |
| p2gm | 3.16 | 39.4 | 14.8 | | | 3.19 | 4.86 | - | 4.74 | | 278 | |
| p2gg | 3.29 | 39.4 | 14.2 | p6 | p2 | 3.31 | 4.57 | - | 5.04 | 2.67 | 278 | 1.85E-04 |
| p4 | 4.62 | 43.3 | 20.8 | | | 4.64 | 2.35 | - | 9.80 | | 278 | |
| p4mm | 5.72 | 51.3 | 33.5 | | | 5.73 | 1.36 | - | 16.9 | | 278 | |
| p4gm | 5.72 | 51.3 | 33.5 | | | 5.73 | 1.36 | - | 16.9 | | 278 | |
| **p3** | **1.38** | **30.2** | **15.1** | | | **1.42** | **11.8** | **25.3** | **1.95** | | **276** | |
| p3m1 | 3.36 | 37.8 | 15.4 | | | 3.38 | 4.43 | - | 5.20 | | 275 | |
| p31m | 3.36 | 37.8 | 25.7 | | | 3.38 | 4.43 | - | 5.20 | | 275 | |
| **p6** | **1.39** | **30.2** | **16.8** | | | **1.41** | **11.8** | **25.3** | **1.95** | | **276** | |
| p6mm | 3.37 | 37.8 | 26.8 | | | 3.38 | 4.42 | - | 5.20 | | 275 | |

Table A16: Results for **2048-pixel** diameter **circular** selection of image with **RGB noise level 1.00**

| Geometric Model | $J_{FC}$ | $F_{res}$ | $\phi_{res}$ (°) | CRISP's Suggestion | Estimated K-L-Best | G-AIC | G-AW full set (%) | G-AW p2, p3, p6 (%) | $E_{best,j}$ | $E_{p6,p3}$ | N | $\hat{\epsilon}^2$ |
|---|---|---|---|---|---|---|---|---|---|---|---|---|
| **p2** | **6.48E-03** | - | **2.60** | | | **0.0195** | **15.6** | **46.3** | - | | **268** | |
| p1m1 | 2.06 | 33.6 | 8.41 | | | 2.07 | 5.59 | - | 2.79 | | 268 | |
| p11m | 2.06 | 33.6 | 7.83 | | | 2.07 | 5.60 | - | 2.79 | | 268 | |
| p1g1 | 2.17 | 33.6 | 8.93 | | | 2.18 | 5.30 | - | 2.95 | | 268 | |
| p11g | 2.17 | 33.6 | 8.40 | | | 2.18 | 5.29 | - | 2.95 | | 268 | |
| p2mm | 2.07 | 33.6 | 14.6 | | | 2.07 | 5.59 | - | 2.79 | | 268 | |
| p2mg | 2.18 | 33.6 | 15.8 | | | 2.18 | 5.29 | - | 2.95 | | 268 | |
| p2gm | 2.18 | 33.6 | 16.0 | | | 2.18 | 5.29 | - | 2.95 | | 268 | |
| p2gg | 2.07 | 33.6 | 14.6 | p6 | p2 | 2.07 | 5.59 | - | 2.79 | 1.82 | 268 | 4.84E-05 |
| p4 | 3.41 | 35.2 | 22.7 | | | 3.42 | 2.86 | - | 5.46 | | 268 | |
| p4mm | 3.67 | 46.0 | 30.5 | | | 3.68 | 2.51 | - | 6.23 | | 266 | |
| p4gm | 3.68 | 46.0 | 30.9 | | | 3.68 | 2.51 | - | 6.23 | | 266 | |
| **p3** | **1.10** | **30.9** | **12.9** | | | **1.10** | **9.07** | **26.9** | **1.72** | | **267** | |
| p3m1 | 2.31 | 39.9 | 14.9 | | | 2.31 | 4.96 | - | 3.15 | | 263 | |
| p31m | 2.31 | 39.9 | 28.1 | | | 2.31 | 4.96 | - | 3.15 | | 263 | |
| **p6** | **1.11** | **30.9** | **22.0** | | | **1.11** | **9.05** | **26.8** | **1.72** | | **267** | |
| p6mm | 2.32 | 39.9 | 29.3 | | | 2.32 | 4.94 | - | 3.16 | | 263 | |

Table A17: Results for **1024-pixel square** selection of image with **10-pixel spread noise**

| Geometric Model | $J_{FC}$ | $F_{res}$ | $\phi_{res}$ (°) | CRISP's Suggestion | Estimated K-L-Best | G-AIC | G-AW full set (%) | G-AW p2, p3, p6 (%) | $E_{best,j}$ | $E_{p6,p3}$ | N | $\hat{\epsilon}^2$ |
|---|---|---|---|---|---|---|---|---|---|---|---|---|
| **p2** | **0.0424** | - | **4.98** | | | **0.127** | **16.0** | **39.5** | - | | **200** | |
| p1m1 | 3.43 | 44.1 | 16.7 | | | 3.51 | 2.94 | - | 5.43 | | 200 | |
| p11m | 2.27 | 44.1 | 8.59 | | | 2.36 | 5.24 | - | 3.05 | | 200 | |
| p1g1 | 2.29 | 44.1 | 9.97 | | | 2.38 | 5.19 | - | 3.08 | | 200 | |
| p11g | 2.27 | 44.1 | 8.15 | | | 2.36 | 5.24 | - | 3.05 | | 200 | |
| p2mm | 2.51 | 44.1 | 14.9 | | | 2.55 | 4.76 | - | 3.36 | | 200 | |
| p2mg | 2.31 | 44.1 | 13.9 | | | 2.35 | 5.25 | - | 3.05 | | 200 | |
| p2gm | 2.31 | 44.1 | 13.8 | | | 2.36 | 5.24 | - | 3.05 | | 200 | |
| p2gg | 2.51 | 44.1 | 14.9 | p6 | p2 | 2.55 | 4.76 | - | 3.36 | 2.08 | 200 | 4.24E-04 |
| p4 | 3.31 | 48.1 | 21.0 | | | 3.35 | 3.19 | - | 5.01 | | 199 | |
| p4mm | 4.00 | 55.4 | 31.2 | | | 4.02 | 2.29 | - | 6.99 | | 198 | |
| p4gm | 4.00 | 55.4 | 31.2 | | | 4.02 | 2.29 | - | 6.99 | | 198 | |
| **p3** | **0.612** | **28.2** | **8.51** | | | **0.667** | **12.2** | **30.2** | **1.31** | | **197** | |
| p3m1 | 2.09 | 36.1 | 12.8 | | | 2.12 | 5.89 | - | 2.71 | | 200 | |
| p31m | 5.01 | 36.1 | 36.0 | | | 5.04 | 1.37 | - | 11.7 | | 200 | |
| **p6** | **0.631** | **28.2** | **12.0** | | | **0.659** | **12.3** | **30.3** | **1.30** | | **197** | |
| p6mm | 2.11 | 36.1 | 22.2 | | | 2.13 | 5.89 | - | 2.72 | | 200 | |

Table A18: Results for **1024-pixel** diameter **circular** selection of image with **10-pixel spread noise**

| Geometric Model | $J_{FC}$ | $F_{res}$ | $\phi_{res}$ (°) | CRISP's Suggestion | Estimated K-L-Best | G-AIC | G-AW full set (%) | G-AW p2, p3, p6 (%) | $E_{best, j}$ | $E_{p6, p3}$ | N | $\hat{\epsilon}^2$ |
|---|---|---|---|---|---|---|---|---|---|---|---|---|
| **p2** | **7.99E-03** | - | **2.43** | | | **0.0240** | **16.4** | **40.6** | - | | **189** | |
| p1m1 | 3.09 | 45.0 | 14.2 | | | 3.11 | 3.50 | - | 4.68 | | 189 | |
| p11m | 2.34 | 45.0 | 7.88 | | | 2.35 | 5.11 | - | 3.20 | | 189 | |
| p1g1 | 2.33 | 45.0 | 8.19 | | | 2.34 | 5.14 | - | 3.19 | | 189 | |
| p11g | 2.34 | 45.0 | 5.56 | | | 2.35 | 5.11 | - | 3.20 | | 189 | |
| p2mm | 2.42 | 45.0 | 11.3 | | | 2.43 | 4.92 | - | 3.33 | | 189 | |
| p2mg | 2.34 | 45.0 | 10.8 | | | 2.35 | 5.12 | - | 3.20 | | 189 | |
| p2gm | 2.34 | 45.0 | 10.8 | | | 2.35 | 5.12 | - | 3.20 | | 189 | |
| p2gg | 2.42 | 45.0 | 11.3 | p6 | p2 | 2.43 | 4.92 | - | 3.33 | 2.10 | 189 | 8.45E-05 |
| p4 | 3.22 | 48.3 | 18.3 | | | 3.23 | 3.29 | - | 4.97 | | 188 | |
| p4mm | 3.96 | 55.8 | 28.8 | | | 3.96 | 2.28 | - | 7.17 | | 185 | |
| p4gm | 3.96 | 55.8 | 28.8 | | | 3.96 | 2.28 | - | 7.17 | | 185 | |
| **p3** | **0.641** | **28.0** | **6.57** | | | **0.651** | **12.0** | **29.7** | **1.37** | | **186** | |
| p3m1 | 2.13 | 36.0 | 10.6 | | | 2.13 | 5.71 | - | 2.87 | | 186 | |
| p31m | 4.90 | 36.0 | 34.8 | | | 4.90 | 1.43 | - | 11.5 | | 186 | |
| **p6** | **0.643** | **28.0** | **10.7** | | | **0.648** | **12.0** | **29.7** | **1.37** | | **186** | |
| p6mm | 2.13 | 36.0 | 20.1 | | | 2.13 | 5.72 | - | 2.87 | | 186 | |

Table A19: Results for **2048-pixel square** selection of image with **10-pixel spread noise**

| Geometric Model | $J_{FC}$ | $F_{res}$ | $\phi_{res}$ (°) | CRISP's Suggestion | Estimated K-L-Best | G-AIC | G-AW full set (%) | G-AW p2, p3, p6 (%) | $E_{best, j}$ | $E_{p6, p3}$ | N | $\hat{\epsilon}^2$ |
|---|---|---|---|---|---|---|---|---|---|---|---|---|
| **p2** | **0.0131** | - | **2.48** | | | **0.0393** | **19.2** | **42.0** | - | | **195** | |
| p1m1 | 3.45 | 43.2 | 16.4 | | | 3.47 | 3.45 | - | 5.56 | | 195 | |
| p11m | 2.71 | 43.2 | 5.80 | | | 2.73 | 4.99 | - | 3.84 | | 195 | |
| p1g1 | 2.70 | 43.2 | 6.10 | | | 2.72 | 5.01 | - | 3.83 | | 195 | |
| p11g | 3.70 | 43.2 | 15.6 | | | 3.72 | 3.04 | - | 6.30 | | 195 | |
| p2mm | 2.82 | 43.2 | 12.5 | | | 2.83 | 4.74 | - | 4.04 | | 195 | |
| p2mg | 2.71 | 43.2 | 11.9 | | | 2.73 | 5.00 | - | 3.84 | | 195 | |
| p2gm | 2.71 | 43.2 | 11.9 | | | 2.73 | 5.00 | - | 3.84 | | 195 | |
| p2gg | 2.82 | 43.2 | 12.6 | p6 | p2 | 2.83 | 4.74 | - | 4.05 | 2.42 | 195 | 1.34E-04 |
| p4 | 3.91 | 46.6 | 19.9 | | | 3.92 | 2.75 | - | 6.97 | | 195 | |
| p4mm | 4.79 | 53.4 | 31.6 | | | 4.80 | 1.78 | - | 10.8 | | 194 | |
| p4gm | 4.79 | 53.4 | 31.6 | | | 4.80 | 1.78 | - | 10.8 | | 194 | |
| **p3** | **0.767** | **27.7** | **7.84** | | | **0.785** | **13.2** | **28.9** | **1.45** | | **195** | |
| p3m1 | 2.54 | 34.6 | 11.7 | | | 2.55 | 5.47 | - | 3.51 | | 195 | |
| p31m | 5.65 | 34.6 | 36.0 | | | 5.66 | 1.15 | - | 16.6 | | 195 | |
| **p6** | **0.769** | **27.7** | **11.5** | | | **0.778** | **13.3** | **29.0** | **1.45** | | **195** | |
| p6mm | 2.54 | 34.6 | 21.4 | | | 2.54 | 5.48 | - | 3.50 | | 195 | |

Table A20: Results for **2048-pixel** diameter **circular** selection of image with **10-pixel spread noise**

| Geometric Model | $J_{FC}$ | $F_{res}$ | $\phi_{res}$ (°) | CRISP's Suggestion | Estimated K-L-Best | G-AIC | G-AW full set (%) | G-AW p2, p3, p6 (%) | $E_{best, j}$ | $E_{p6, p3}$ | N | $\hat{\epsilon}^2$ |
|---|---|---|---|---|---|---|---|---|---|---|---|---|
| **p2** | **1.94E-03** | **-** | **1.41** | | | **5.81E-03** | **13.9** | **39.6** | **-** | | **189** | |
| p1m1 | 1.90 | 38.9 | 6.89 | | | 1.90 | 5.37 | - | 2.58 | | 189 | |
| p11m | 1.90 | 38.9 | 6.47 | | | 1.90 | 5.37 | - | 2.58 | | 189 | |
| p1g1 | 1.90 | 38.9 | 6.23 | | | 1.91 | 5.36 | - | 2.59 | | 189 | |
| p11g | 1.94 | 38.9 | 6.58 | | | 1.94 | 5.26 | - | 2.64 | | 189 | |
| p2mm | 1.90 | 38.9 | 12.0 | | | 1.90 | 5.37 | - | 2.58 | | 189 | |
| p2mg | 1.94 | 38.9 | 12.6 | | | 1.94 | 5.26 | - | 2.64 | | 189 | |
| p2gm | 1.94 | 38.9 | 12.6 | | | 1.94 | 5.26 | - | 2.64 | | 189 | |
| p2gg | 1.90 | 38.9 | 12.0 | p6 | p2 | 1.90 | 5.37 | - | 2.58 | 1.79 | 189 | 2.05E-05 |
| p4 | 3.08 | 39.8 | 22.1 | | | 3.08 | 2.99 | - | 4.64 | | 189 | |
| p4mm | 3.27 | 48.7 | 29.1 | | | 3.27 | 2.71 | - | 5.11 | | 186 | |
| p4gm | 3.27 | 48.7 | 29.0 | | | 3.27 | 2.71 | - | 5.11 | | 186 | |
| **p3** | **0.543** | **29.0** | **7.65** | | | **0.546** | **10.6** | **30.2** | **1.31** | | **189** | |
| p3m1 | 1.71 | 37.8 | 10.7 | | | 1.71 | 5.92 | - | 2.34 | | 189 | |
| p31m | 3.81 | 37.8 | 36.4 | | | 3.82 | 2.06 | - | 6.72 | | 189 | |
| **p6** | **0.545** | **29.0** | **13.7** | | | **0.546** | **10.6** | **30.2** | **1.31** | | **189** | |
| p6mm | 1.71 | 37.8 | 22.2 | | | 1.71 | 5.92 | - | 2.34 | | 189 | |

Table A21: Results for **1024-pixel square** selection of image with **RGB noise level 0.25 and 10-pixel spread noise**

| Geometric Model | $J_{FC}$ | $F_{res}$ | $\phi_{res}$ (°) | CRISP's Suggestion | Estimated K-L-Best | G-AIC | G-AW full set (%) | G-AW p2, p3, p6 (%) | $E_{best, j}$ | $E_{p6, p3}$ | N | $\hat{\epsilon}^2$ |
|---|---|---|---|---|---|---|---|---|---|---|---|---|
| **p2** | **0.0389** | **-** | **5.05** | | | **0.117** | **16.4** | **40.0** | **-** | | **215** | |
| p1m1 | 3.38 | 43.1 | 16.6 | | | 3.46 | 3.08 | - | 5.33 | | 215 | |
| p11m | 2.24 | 43.1 | 9.56 | | | 2.32 | 5.44 | - | 3.01 | | 215 | |
| p1g1 | 2.27 | 43.1 | 10.2 | | | 2.35 | 5.37 | - | 3.05 | | 215 | |
| p11g | 3.25 | 43.1 | 16.3 | | | 3.32 | 3.30 | - | 4.97 | | 215 | |
| p2mm | 2.52 | 43.1 | 15.4 | | | 2.55 | 4.84 | - | 3.38 | | 215 | |
| p2mg | 2.28 | 43.1 | 14.1 | | | 2.32 | 5.44 | - | 3.01 | | 215 | |
| p2gm | 2.28 | 43.1 | 14.1 | | | 2.32 | 5.44 | - | 3.01 | | 215 | |
| p2gg | 2.52 | 43.1 | 15.4 | p6 | p2 | 2.55 | 4.84 | - | 3.38 | 2.05 | 215 | 3.62E-04 |
| p4 | 3.32 | 47.7 | 21.1 | | | 3.36 | 3.24 | - | 5.06 | | 215 | |
| p4mm | 4.03 | 55.3 | 31.5 | | | 4.05 | 2.30 | - | 7.13 | | 213 | |
| p4gm | 4.03 | 55.3 | 31.5 | | | 4.05 | 2.30 | - | 7.13 | | 213 | |
| **p3** | **0.639** | **28.6** | **9.65** | | | **0.690** | **12.3** | **30.0** | **1.33** | | **213** | |
| p3m1 | 2.10 | 36.6 | 12.7 | | | 2.13 | 5.99 | - | 2.73 | | 209 | |
| p31m | 4.99 | 36.6 | 35.9 | | | 5.02 | 1.41 | - | 11.6 | | 209 | |
| **p6** | **0.663** | **28.6** | **12.8** | | | **0.689** | **12.3** | **30.0** | **1.33** | | **213** | |
| p6mm | 2.11 | 36.6 | 21.7 | | | 2.13 | 6.00 | - | 2.73 | | 209 | |

Table A22: Results for **1024-pixel** diameter **circular** selection of image with **RGB noise level 0.25 and 10-pixel spread noise**

| Geometric Model | $J_{FC}$ | $F_{res}$ | $\phi_{res}$ (°) | CRISP's Suggestion | Estimated K-L-Best | G-AIC | G-AW full set (%) | G-AW $p2, p3, p6$ (%) | $E_{best, j}$ | $E_{p6, p3}$ | $N$ | $\hat{\epsilon}^2$ |
|---|---|---|---|---|---|---|---|---|---|---|---|---|
| **p2** | **0.0101** | **-** | **3.11** | | | **0.0304** | **17.0** | **40.8** | **-** | | **211** | |
| p1m1 | 3.28 | 44.1 | 14.9 | | | 3.30 | 3.31 | - | 5.13 | | 211 | |
| p11m | 2.37 | 44.1 | 10.5 | | | 2.39 | 5.21 | - | 3.25 | | 211 | |
| p1g1 | 2.35 | 44.1 | 8.94 | | | 2.37 | 5.28 | - | 3.21 | | 211 | |
| p11g | 3.30 | 44.1 | 15.7 | | | 3.32 | 3.28 | - | 5.17 | | 211 | |
| p2mm | 2.48 | 44.1 | 12.5 | | | 2.49 | 4.95 | - | 3.43 | | 211 | |
| p2mg | 2.36 | 44.1 | 11.9 | | | 2.37 | 5.26 | - | 3.23 | | 211 | |
| p2gm | 2.36 | 44.1 | 11.9 | | | 2.37 | 5.26 | - | 3.23 | | 211 | |
| p2gg | 2.48 | 44.1 | 12.5 | p6 | p2 | 2.49 | 4.95 | - | 3.43 | 2.10 | 211 | 9.60E-05 |
| p4 | 3.36 | 48.1 | 19.1 | | | 3.37 | 3.20 | - | 5.30 | | 210 | |
| p4mm | 4.08 | 54.8 | 29.7 | | | 4.09 | 2.23 | - | 7.60 | | 207 | |
| p4gm | 4.08 | 54.8 | 29.7 | | | 4.09 | 2.23 | - | 7.60 | | 207 | |
| **p3** | **0.662** | **28.3** | **7.77** | | | **0.675** | **12.3** | **29.6** | **1.38** | | **208** | |
| p3m1 | 2.15 | 36.1 | 10.7 | | | 2.15 | 5.87 | - | 2.89 | | 207 | |
| p31m | 4.90 | 36.1 | 34.8 | | | 4.90 | 1.49 | - | 11.4 | | 207 | |
| **p6** | **0.661** | **28.3** | **11.2** | | | **0.668** | **12.3** | **29.7** | **1.38** | | **208** | |
| p6mm | 2.15 | 36.1 | 20.3 | | | 2.15 | 5.88 | - | 2.89 | | 207 | |

Table A23: Results for **2048-pixel square** selection of image with **RGB noise level 0.25 and 10-pixel spread noise**

| Geometric Model | $J_{FC}$ | $F_{res}$ | $\phi_{res}$ (°) | CRISP's Suggestion | Estimated K-L-Best | G-AIC | G-AW full set (%) | G-AW $p2, p3, p6$ (%) | $E_{best, j}$ | $E_{p6, p3}$ | $N$ | $\hat{\epsilon}^2$ |
|---|---|---|---|---|---|---|---|---|---|---|---|---|
| **p2** | **0.0134** | **-** | **2.59** | | | **0.0401** | **18.9** | **41.9** | **-** | | **190** | |
| p1m1 | 3.38 | 42.7 | 16.3 | | | 3.40 | 3.51 | - | 5.38 | | 190 | |
| p11m | 2.63 | 42.7 | 5.54 | | | 2.66 | 5.09 | - | 3.71 | | 190 | |
| p1g1 | 2.63 | 42.7 | 5.95 | | | 2.65 | 5.11 | - | 3.70 | | 190 | |
| p11g | 3.65 | 42.7 | 15.6 | | | 3.67 | 3.07 | - | 6.15 | | 190 | |
| p2mm | 2.79 | 42.7 | 12.7 | | | 2.81 | 4.73 | - | 3.99 | | 190 | |
| p2mg | 2.64 | 42.7 | 11.9 | | | 2.66 | 5.10 | - | 3.70 | | 190 | |
| p2gm | 2.64 | 42.7 | 11.9 | | | 2.66 | 5.10 | - | 3.70 | | 190 | |
| p2gg | 2.79 | 42.7 | 12.7 | p6 | p2 | 2.81 | 4.73 | - | 3.99 | 2.41 | 190 | 1.41E-04 |
| p4 | 3.86 | 46.5 | 19.8 | | | 3.87 | 2.78 | - | 6.78 | | 189 | |
| p4mm | 4.75 | 53.3 | 31.3 | | | 4.75 | 1.79 | - | 10.5 | | 189 | |
| p4gm | 4.75 | 53.3 | 31.3 | | | 4.75 | 1.79 | - | 10.5 | | 189 | |
| **p3** | **0.759** | **27.7** | **7.49** | | | **0.776** | **13.1** | **29.0** | **1.45** | | **189** | |
| p3m1 | 2.51 | 34.7 | 11.2 | | | 2.52 | 5.46 | - | 3.46 | | 190 | |
| p31m | 5.59 | 34.7 | 35.4 | | | 5.60 | 1.17 | - | 16.1 | | 190 | |
| **p6** | **0.754** | **27.7** | **10.7** | | | **0.763** | **13.2** | **29.2** | **1.44** | | **189** | |
| p6mm | 2.51 | 34.7 | 20.7 | | | 2.52 | 5.47 | - | 3.45 | | 190 | |

Table A24: Results for **2048-pixel** diameter **circular** selection of image with **RGB noise level 0.25 and 10-pixel spread noise**

| Geometric Model | $J_{FC}$ | $F_{res}$ | $\phi_{res}$ (°) | CRISP's Suggestion | Estimated K-L-Best | G-AIC | G-AW full set (%) | G-AW p2, p3, p6 (%) | $E_{best, j}$ | $E_{p6, p3}$ | $N$ | $\hat{\epsilon}^2$ |
|---|---|---|---|---|---|---|---|---|---|---|---|---|
| **p2** | **2.56E-03** | - | **1.53** | | | **7.69E-03** | **13.8** | **39.6** | - | | **186** | |
| p1m1 | 1.89 | 38.5 | 6.97 | | | 1.89 | 5.38 | - | 2.57 | | 186 | |
| p11m | 1.89 | 38.5 | 6.79 | | | 1.89 | 5.38 | - | 2.57 | | 186 | |
| p1g1 | 1.89 | 38.5 | 5.51 | | | 1.89 | 5.38 | - | 2.57 | | 186 | |
| p11g | 1.93 | 38.5 | 6.73 | | | 1.93 | 5.27 | - | 2.62 | | 186 | |
| p2mm | 1.89 | 38.5 | 12.3 | | | 1.89 | 5.39 | - | 2.56 | | 186 | |
| p2mg | 1.93 | 38.5 | 12.9 | | | 1.93 | 5.27 | - | 2.62 | | 186 | |
| p2gm | 1.93 | 38.5 | 12.9 | | | 1.93 | 5.27 | - | 2.62 | | 186 | |
| p2gg | 1.89 | 38.5 | 12.3 | p6 | p2 | 1.89 | 5.39 | - | 2.56 | 1.79 | 186 | 2.76E-05 |
| p4 | 3.06 | 39.8 | 22.4 | | | 3.06 | 3.00 | - | 4.60 | | 186 | |
| p4mm | 3.25 | 48.7 | 29.4 | | | 3.25 | 2.73 | - | 5.06 | | 183 | |
| p4gm | 3.25 | 48.7 | 29.4 | | | 3.25 | 2.73 | - | 5.06 | | 183 | |
| **p3** | **0.546** | **29.0** | **7.61** | | | **0.549** | **10.5** | **30.2** | **1.31** | | **186** | |
| p3m1 | 1.71 | 38.1 | 10.6 | | | 1.71 | 5.88 | - | 2.35 | | 186 | |
| p31m | 3.74 | 38.1 | 36.2 | | | 3.74 | 2.14 | - | 6.47 | | 186 | |
| **p6** | **0.546** | **29.0** | **13.5** | | | **0.547** | **10.5** | **30.2** | **1.31** | | **186** | |
| p6mm | 1.71 | 38.1 | 22.7 | | | 1.71 | 5.88 | - | 2.35 | | 186 | |

Table A25: Results for **1024-pixel square** selection of image with **RGB noise 0.50 and 10-pixel spread noise**

| Geometric Model | $J_{FC}$ | $F_{res}$ | $\phi_{res}$ (°) | CRISP's Suggestion | Estimated K-L-Best | G-AIC | G-AW full set (%) | G-AW p2, p3, p6 (%) | $E_{best, j}$ | $E_{p6, p3}$ | $N$ | $\hat{\epsilon}^2$ |
|---|---|---|---|---|---|---|---|---|---|---|---|---|
| **p2** | **0.0462** | - | **6.35** | | | **0.139** | **15.8** | **39.7** | - | | **214** | |
| p1m1 | 3.21 | 42.9 | 16.1 | | | 3.30 | 3.27 | - | 4.85 | | 214 | |
| p11m | 2.17 | 42.9 | 8.66 | | | 2.26 | 5.49 | - | 2.89 | | 214 | |
| p1g1 | 2.18 | 42.9 | 9.76 | | | 2.27 | 5.45 | - | 2.91 | | 214 | |
| p11g | 3.15 | 42.9 | 15.7 | | | 3.24 | 3.36 | - | 4.72 | | 214 | |
| p2mm | 2.41 | 42.9 | 15.8 | | | 2.45 | 4.98 | - | 3.18 | | 214 | |
| p2mg | 2.21 | 42.9 | 14.3 | | | 2.26 | 5.50 | - | 2.88 | | 214 | |
| p2gm | 2.21 | 42.9 | 14.3 | | | 2.26 | 5.50 | - | 2.88 | | 214 | |
| p2gg | 2.41 | 42.9 | 15.8 | p6 | p2 | 2.45 | 4.98 | - | 3.18 | 2.04 | 214 | 4.32E-04 |
| p4 | 3.20 | 47.6 | 21.5 | | | 3.25 | 3.35 | - | 4.73 | | 214 | |
| p4mm | 3.87 | 54.7 | 31.5 | | | 3.90 | 2.42 | - | 6.55 | | 211 | |
| p4gm | 3.87 | 54.7 | 31.5 | | | 3.90 | 2.42 | - | 6.55 | | 211 | |
| **p3** | **0.632** | **28.5** | **9.46** | | | **0.692** | **12.0** | **30.1** | **1.32** | | **207** | |
| p3m1 | 2.07 | 36.4 | 13.5 | | | 2.10 | 5.93 | - | 2.67 | | 208 | |
| p31m | 4.87 | 36.4 | 35.7 | | | 4.90 | 1.46 | - | 10.8 | | 208 | |
| **p6** | **0.652** | **28.5** | **13.0** | | | **0.682** | **12.1** | **30.2** | **1.31** | | **207** | |
| p6mm | 2.09 | 36.4 | 22.7 | | | 2.10 | 5.93 | - | 2.67 | | 208 | |

Table A26: Results for **1024-pixel** diameter **circular** selection of image with **RGB noise level 0.50 and 10-pixel spread noise**

| Geometric Model | $J_{FC}$ | $F_{res}$ | $\phi_{res}$ (°) | CRISP's Suggestion | Estimated K-L-Best | G-AIC | G-AW full set (%) | G-AW p2, p3, p6 (%) | $E_{best, j}$ | $E_{p6, p3}$ | N | $\hat{\epsilon}^2$ |
|---|---|---|---|---|---|---|---|---|---|---|---|---|
| **p2** | **0.0175** | - | **4.35** | | | **0.0526** | **15.7** | **40.3** | - | | **211** | |
| p1m1 | 3.06 | 44.1 | 13.9 | | | 3.10 | 3.43 | - | 4.59 | | 211 | |
| p11m | 2.21 | 44.1 | 7.67 | | | 2.25 | 5.25 | - | 3.00 | | 211 | |
| p1g1 | 2.20 | 44.1 | 8.16 | | | 2.24 | 5.28 | - | 2.98 | | 211 | |
| p11g | 2.21 | 44.1 | 6.36 | | | 2.25 | 5.25 | - | 3.00 | | 211 | |
| p2mm | 2.30 | 44.1 | 13.2 | | | 2.32 | 5.07 | - | 3.10 | | 211 | |
| p2mg | 2.23 | 44.1 | 12.0 | | | 2.24 | 5.26 | - | 2.99 | | 211 | |
| p2gm | 2.23 | 44.1 | 12.0 | | | 2.24 | 5.26 | - | 2.99 | | 211 | |
| p2gg | 2.30 | 44.1 | 13.0 | p6 | p2 | 2.32 | 5.07 | - | 3.10 | 2.05 | 211 | 1.66E-04 |
| p4 | 3.12 | 48.0 | 19.2 | | | 3.13 | 3.37 | - | 4.67 | | 211 | |
| p4mm | 3.78 | 55.3 | 29.4 | | | 3.79 | 2.43 | - | 6.48 | | 206 | |
| p4gm | 3.78 | 55.3 | 29.3 | | | 3.79 | 2.43 | - | 6.48 | | 206 | |
| **p3** | **0.638** | **27.9** | **7.23** | | | **0.661** | **11.6** | **29.8** | **1.36** | | **205** | |
| p3m1 | 2.08 | 35.3 | 11.0 | | | 2.09 | 5.69 | - | 2.76 | | 202 | |
| p31m | 4.72 | 35.3 | 34.5 | | | 4.74 | 1.51 | - | 10.4 | | 202 | |
| **p6** | **0.641** | **27.9** | **11.5** | | | **0.653** | **11.7** | **29.9** | **1.35** | | **205** | |
| p6mm | 2.08 | 35.3 | 21.4 | | | 2.08 | 5.70 | - | 2.76 | | 202 | |

Table A27: Results **for 2048-pixel square** selection of image with **RGB noise level 0.50 and 10-pixel spread noise**

| Geometric Model | $J_{FC}$ | $F_{res}$ | $\phi_{res}$ (°) | CRISP's Suggestion | Estimated K-L-Best | G-AIC | G-AW full set (%) | G-AW p2, p3, p6 (%) | $E_{best, j}$ | $E_{p6, p3}$ | N | $\hat{\epsilon}^2$ |
|---|---|---|---|---|---|---|---|---|---|---|---|---|
| **p2** | **0.0148** | - | **2.85** | | | **0.0443** | **18.8** | **41.8** | - | | **196** | |
| p1m1 | 3.40 | 43.2 | 16.2 | | | 3.43 | 3.46 | - | 5.43 | | 196 | |
| p11m | 2.63 | 43.2 | 5.67 | | | 2.66 | 5.08 | - | 3.70 | | 196 | |
| p1g1 | 2.62 | 43.2 | 6.12 | | | 2.65 | 5.10 | - | 3.69 | | 196 | |
| p11g | 3.63 | 43.2 | 15.6 | | | 3.66 | 3.07 | - | 6.11 | | 196 | |
| p2mm | 2.77 | 43.2 | 12.8 | | | 2.79 | 4.76 | - | 3.94 | | 196 | |
| p2mg | 2.64 | 43.2 | 12.0 | | | 2.66 | 5.09 | - | 3.69 | | 196 | |
| p2gm | 2.64 | 43.2 | 12.2 | | | 2.66 | 5.08 | - | 3.70 | | 196 | |
| p2gg | 2.77 | 43.2 | 13.0 | p6 | p2 | 2.79 | 4.76 | - | 3.94 | 2.40 | 196 | 1.51E-04 |
| p4 | 3.83 | 46.9 | 19.7 | | | 3.84 | 2.82 | - | 6.67 | | 196 | |
| p4mm | 4.70 | 53.7 | 31.1 | | | 4.70 | 1.83 | - | 10.3 | | 196 | |
| p4gm | 4.70 | 53.7 | 30.8 | | | 4.70 | 1.83 | - | 10.3 | | 196 | |
| **p3** | **0.757** | **27.9** | **7.74** | | | **0.777** | **13.0** | **29.0** | **1.44** | | **194** | |
| p3m1 | 2.51 | 34.8 | 11.8 | | | 2.52 | 5.45 | - | 3.45 | | 194 | |
| p31m | 5.42 | 34.8 | 35.6 | | | 5.43 | 1.27 | - | 14.7 | | 194 | |
| **p6** | **0.755** | **27.9** | **11.0** | | | **0.764** | **13.1** | **29.2** | **1.43** | | **194** | |
| p6mm | 2.51 | 34.8 | 21.2 | | | 2.51 | 5.47 | - | 3.44 | | 194 | |

Table A28: Results for **2048-pixel** diameter **circular** selection of image with **RGB noise level 0.50 and 10-pixel spread noise**

| Geometric Model | $J_{FC}$ | $F_{res}$ | $\phi_{res}$ (°) | CRISP's Suggestion | Estimated K-L-Best | G-AIC | G-AW full set (%) | G-AW p2, p3, p6 (%) | $E_{best, j}$ | $E_{p6, p3}$ | $N$ | $\hat{\epsilon}^2$ |
|---|---|---|---|---|---|---|---|---|---|---|---|---|
| **p2** | **3.14E-03** | - | **1.90** | | | **9.43E-03** | **13.3** | **39.4** | - | | **184** | |
| p1m1 | 1.78 | 36.3 | 6.90 | | | 1.79 | 5.48 | - | 2.43 | | 184 | |
| p11m | 1.78 | 36.3 | 6.77 | | | 1.79 | 5.48 | - | 2.43 | | 184 | |
| p1g1 | 1.78 | 36.3 | 6.41 | | | 1.79 | 5.48 | - | 2.44 | | 184 | |
| p11g | 1.83 | 36.3 | 7.13 | | | 1.84 | 5.34 | - | 2.50 | | 184 | |
| p2mm | 1.78 | 36.3 | 12.7 | | | 1.79 | 5.48 | - | 2.43 | | 184 | |
| p2mg | 1.84 | 36.3 | 13.2 | | | 1.84 | 5.34 | - | 2.50 | | 184 | |
| p2gm | 1.84 | 36.3 | 13.2 | | | 1.84 | 5.34 | - | 2.50 | | 184 | |
| p2gg | 1.78 | 36.3 | 12.7 | p6 | p2 | 1.79 | 5.48 | - | 2.43 | 1.73 | 184 | 3.42E-05 |
| p4 | 2.97 | 38.4 | 22.6 | | | 2.98 | 3.03 | - | 4.41 | | 184 | |
| p4mm | 3.11 | 46.7 | 29.4 | | | 3.12 | 2.82 | - | 4.73 | | 181 | |
| p4gm | 3.11 | 46.7 | 29.3 | | | 3.12 | 2.82 | - | 4.73 | | 181 | |
| **p3** | **0.530** | **28.6** | **7.46** | | | **0.534** | **10.3** | **30.3** | **1.30** | | **184** | |
| p3m1 | 1.63 | 36.9 | 10.6 | | | 1.63 | 5.93 | - | 2.25 | | 184 | |
| p31m | 3.63 | 36.9 | 36.1 | | | 3.64 | 2.18 | - | 6.13 | | 184 | |
| **p6** | **0.530** | **28.6** | **13.5** | | | **0.532** | **10.3** | **30.3** | **1.30** | | **184** | |
| p6mm | 1.63 | 36.9 | 22.6 | | | 1.63 | 5.92 | - | 2.25 | | 184 | |

Table A29: Results for **1024-pixel square** selection of image with **RGB noise level 0.75 and 10-pixel spread noise**

| Geometric Model | $J_{FC}$ | $F_{res}$ | $\phi_{res}$ (°) | CRISP's Suggestion | Estimated K-L-Best | G-AIC | G-AW full set (%) | G-AW p2, p3, p6 (%) | $E_{best, j}$ | $E_{p6, p3}$ | $N$ | $\hat{\epsilon}^2$ |
|---|---|---|---|---|---|---|---|---|---|---|---|---|
| **p2** | **0.0628** | - | **8.03** | | | **0.188** | **16.5** | **39.4** | - | | **237** | |
| p1m1 | 3.49 | 42.3 | 17.0 | | | 3.62 | 2.98 | - | 5.55 | | 237 | |
| p11m | 2.33 | 42.3 | 9.63 | | | 2.46 | 5.31 | - | 3.11 | | 237 | |
| p1g1 | 2.36 | 42.3 | 10.1 | | | 2.49 | 5.24 | - | 3.15 | | 237 | |
| p11g | 3.30 | 42.3 | 15.8 | | | 3.43 | 3.27 | - | 5.05 | | 237 | |
| p2mm | 2.64 | 42.3 | 17.3 | | | 2.70 | 4.71 | - | 3.51 | | 237 | |
| p2mg | 2.40 | 42.3 | 16.9 | | | 2.46 | 5.30 | - | 3.12 | | 237 | |
| p2gm | 2.40 | 42.3 | 16.8 | | | 2.46 | 5.30 | - | 3.12 | | 237 | |
| p2gg | 2.64 | 42.3 | 17.3 | p6 | p2 | 2.70 | 4.71 | - | 3.51 | 2.07 | 237 | 5.30E-04 |
| p4 | 3.51 | 46.3 | 24.0 | | | 3.57 | 3.04 | - | 5.43 | | 237 | |
| p4mm | 4.15 | 54.5 | 32.4 | | | 4.18 | 2.25 | - | 7.35 | | 235 | |
| p4gm | 4.15 | 54.5 | 32.4 | | | 4.18 | 2.25 | - | 7.35 | | 235 | |
| **p3** | **0.639** | **29.0** | **9.14** | | | **0.721** | **12.7** | **30.2** | **1.30** | | **230** | |
| p3m1 | 2.12 | 36.8 | 14.2 | | | 2.16 | 6.16 | - | 2.68 | | 227 | |
| p31m | 5.10 | 36.8 | 35.7 | | | 5.14 | 1.39 | - | 11.9 | | 227 | |
| **p6** | **0.665** | **29.0** | **14.3** | | | **0.706** | **12.8** | **30.4** | **1.30** | | **230** | |
| p6mm | 2.15 | 36.8 | 24.0 | | | 2.17 | 6.15 | - | 2.69 | | 227 | |

Table A30: Results for **1024-pixel** diameter **circular** selection of image with **RGB noise level 0.75 and 10-pixel spread noise**

| Geometric Model | $J_{FC}$ | $F_{res}$ | $\phi_{res}$ (°) | CRISP's Suggestion | Estimated K-L-Best | G-AIC | G-AW full set (%) | G-AW p2, p3, p6 (%) | $E_{best, j}$ | $E_{p6, p3}$ | N | $\hat{\epsilon}^2$ |
|---|---|---|---|---|---|---|---|---|---|---|---|---|
| **p2** | **0.0242** | - | **6.03** | | | **0.0727** | **13.6** | **38.7** | - | | **219** | |
| p1m1 | 1.84 | 36.4 | 7.26 | | | 1.89 | 5.49 | - | 2.48 | | 218 | |
| p11m | 1.83 | 36.4 | 7.46 | | | 1.88 | 5.53 | - | 2.46 | | 218 | |
| p1g1 | 2.13 | 36.4 | 14.3 | | | 2.18 | 4.76 | - | 2.86 | | 218 | |
| p11g | 1.90 | 36.4 | 7.60 | | | 1.95 | 5.33 | - | 2.56 | | 218 | |
| p2mm | 1.86 | 36.4 | 15.3 | | | 1.88 | 5.52 | - | 2.47 | | 218 | |
| p2mg | 1.92 | 36.4 | 15.7 | | | 1.94 | 5.35 | - | 2.55 | | 218 | |
| p2gm | 1.92 | 36.4 | 16.2 | | | 1.94 | 5.35 | - | 2.55 | | 218 | |
| p2gg | 1.86 | 36.4 | 15.3 | p6 | p2 | 1.88 | 5.52 | - | 2.47 | 1.87 | 218 | 2.21E-04 |
| p4 | 3.09 | 37.7 | 24.9 | | | 3.11 | 2.98 | - | 4.58 | | 218 | |
| p4mm | 3.31 | 47.8 | 31.4 | | | 3.32 | 2.68 | - | 5.08 | | 212 | |
| p4gm | 3.31 | 47.8 | 31.4 | | | 3.32 | 2.68 | - | 5.08 | | 212 | |
| **p3** | **0.516** | **29.3** | **8.84** | | | **0.547** | **10.7** | **30.6** | **1.27** | | **210** | |
| p3m1 | 1.77 | 39.2 | 13.3 | | | 1.78 | 5.79 | - | 2.35 | | 210 | |
| p31m | 3.85 | 39.2 | 35.7 | | | 3.87 | 2.04 | - | 6.67 | | 210 | |
| **p6** | **0.521** | **29.3** | **14.9** | | | **0.537** | **10.8** | **30.7** | **1.26** | | **210** | |
| p6mm | 1.77 | 39.2 | 24.2 | | | 1.78 | 5.80 | - | 2.35 | | 210 | |

Table A31: Results for **2048-pixel square** selection of image with **RGB noise level 0.75 and 10-pixel spread noise**

| Geometric Model | $J_{FC}$ | $F_{res}$ | $\phi_{res}$ (°) | CRISP's Suggestion | Estimated K-L-Best | G-AIC | G-AW full set (%) | G-AW p2, p3, p6 (%) | $E_{best, j}$ | $E_{p6, p3}$ | N | $\hat{\epsilon}^2$ |
|---|---|---|---|---|---|---|---|---|---|---|---|---|
| **p2** | **0.0164** | - | **3.13** | | | **0.0493** | **19.0** | **41.9** | - | | **188** | |
| p1m1 | 3.50 | 42.5 | 15.9 | | | 3.53 | 3.32 | - | 5.71 | | 188 | |
| p11m | 2.65 | 42.5 | 5.97 | | | 2.68 | 5.09 | - | 3.73 | | 188 | |
| p1g1 | 2.65 | 42.5 | 6.29 | | | 2.68 | 5.09 | - | 3.73 | | 188 | |
| p11g | 3.66 | 42.5 | 15.6 | | | 3.69 | 3.07 | - | 6.18 | | 188 | |
| p2mm | 2.83 | 42.5 | 13.3 | | | 2.85 | 4.69 | - | 4.05 | | 188 | |
| p2mg | 2.66 | 42.5 | 12.3 | | | 2.68 | 5.10 | - | 3.72 | | 188 | |
| p2gm | 2.66 | 42.5 | 12.3 | | | 2.68 | 5.10 | - | 3.72 | | 188 | |
| p2gg | 2.83 | 42.5 | 13.2 | p6 | p2 | 2.85 | 4.69 | - | 4.05 | 2.40 | 188 | 1.75E-04 |
| p4 | 3.88 | 46.3 | 20.3 | | | 3.90 | 2.77 | - | 6.85 | | 188 | |
| p4mm | 4.76 | 53.2 | 31.2 | | | 4.77 | 1.79 | - | 10.6 | | 187 | |
| p4gm | 4.76 | 53.2 | 31.2 | | | 4.77 | 1.79 | - | 10.6 | | 187 | |
| **p3** | **0.764** | **27.6** | **7.74** | | | **0.785** | **13.1** | **29.0** | **1.44** | | **185** | |
| p3m1 | 2.51 | 35.0 | 11.8 | | | 2.52 | 5.51 | - | 3.45 | | 185 | |
| p31m | 5.61 | 35.0 | 35.6 | | | 5.62 | 1.17 | - | 16.2 | | 185 | |
| **p6** | **0.767** | **27.6** | **11.1** | | | **0.778** | **13.2** | **29.1** | **1.44** | | **185** | |
| p6mm | 2.52 | 35.0 | 21.2 | | | 2.52 | 5.52 | - | 3.44 | | 185 | |

Table A32: Results for **2048-pixel** diameter **circular** selection of image with **RGB noise level 0.75 and 10-pixel spread noise**

| Geometric Model | $J_{FC}$ | $F_{res}$ | $\phi_{res}$ (°) | CRISP's Suggestion | Estimated K-L-Best | G-AIC | G-AW full set (%) | G-AW p2, p3, p6 (%) | $E_{best,j}$ | $E_{p6,p3}$ | N | $\hat{\epsilon}^2$ |
|---|---|---|---|---|---|---|---|---|---|---|---|---|
| **p2** | 4.15E-03 | - | 2.38 | | | 0.0124 | 13.8 | 39.6 | - | | 194 | |
| p1m1 | 1.89 | 38.3 | 7.76 | | | 1.89 | 5.38 | - | 2.56 | | 194 | |
| p11m | 1.89 | 38.3 | 7.40 | | | 1.89 | 5.38 | - | 2.56 | | 194 | |
| p1g1 | 1.91 | 38.3 | 7.65 | | | 1.91 | 5.32 | - | 2.59 | | 194 | |
| p11g | 1.91 | 38.3 | 7.14 | | | 1.91 | 5.32 | - | 2.59 | | 194 | |
| p2mm | 1.89 | 38.3 | 13.1 | | | 1.89 | 5.38 | - | 2.56 | | 194 | |
| p2mg | 1.91 | 38.3 | 13.2 | | | 1.91 | 5.33 | - | 2.59 | | 194 | |
| p2gm | 1.91 | 38.3 | 13.2 | | | 1.91 | 5.33 | - | 2.59 | | 194 | |
| p2gg | 1.89 | 38.3 | 13.1 | p6 | p2 | 1.89 | 5.38 | - | 2.56 | 1.78 | 194 | 4.28E-05 |
| p4 | 3.04 | 39.7 | 22.7 | | | 3.05 | 3.02 | - | 4.56 | | 194 | |
| p4mm | 3.23 | 48.8 | 29.2 | | | 3.24 | 2.75 | - | 5.01 | | 191 | |
| p4gm | 3.23 | 48.8 | 29.2 | | | 3.24 | 2.75 | - | 5.01 | | 191 | |
| **p3** | 0.550 | 29.5 | 8.07 | | | 0.555 | 10.5 | 30.2 | 1.31 | | 192 | |
| p3m1 | 1.71 | 39.0 | 10.7 | | | 1.71 | 5.89 | - | 2.34 | | 189 | |
| p31m | 3.78 | 39.0 | 36.1 | | | 3.79 | 2.09 | - | 6.60 | | 189 | |
| **p6** | 0.553 | 29.5 | 14.2 | | | 0.556 | 10.5 | 30.2 | 1.31 | | 192 | |
| p6mm | 1.71 | 39.0 | 22.3 | | | 1.71 | 5.89 | - | 2.34 | | 189 | |

Table A33: Results for **1024-pixel square** selection of image with **RGB noise level 1.00 and 10-pixel spread noise**

| Geometric Model | $J_{FC}$ | $F_{res}$ | $\phi_{res}$ (°) | CRISP's Suggestion | Estimated K-L-Best | G-AIC | G-AW full set (%) | G-AW p2, p3, p6 (%) | $E_{best,j}$ | $E_{p6,p3}$ | N | $\hat{\epsilon}^2$ |
|---|---|---|---|---|---|---|---|---|---|---|---|---|
| **p2** | 0.0799 | - | 9.27 | | | 0.240 | 15.3 | 38.4 | - | | 238 | |
| p1m1 | 3.27 | 41.3 | 16.3 | | | 3.43 | 3.11 | - | 4.92 | | 238 | |
| p11m | 2.15 | 41.3 | 10.0 | | | 2.31 | 5.44 | - | 2.82 | | 238 | |
| p1g1 | 2.18 | 41.3 | 10.5 | | | 2.34 | 5.37 | - | 2.85 | | 238 | |
| p11g | 3.09 | 41.3 | 16.0 | | | 3.25 | 3.41 | - | 4.50 | | 238 | |
| p2mm | 2.43 | 41.3 | 18.8 | | | 2.51 | 4.93 | - | 3.11 | | 238 | |
| p2mg | 2.22 | 41.3 | 17.2 | | | 2.30 | 5.46 | - | 2.80 | | 238 | |
| p2gm | 2.22 | 41.3 | 17.2 | | | 2.30 | 5.46 | - | 2.80 | | 238 | |
| p2gg | 2.43 | 41.3 | 18.7 | p6 | p2 | 2.51 | 4.93 | - | 3.11 | 2.03 | 238 | 6.72E-04 |
| p4 | 3.15 | 44.5 | 23.8 | | | 3.23 | 3.43 | - | 4.47 | | 238 | |
| p4mm | 3.83 | 53.3 | 32.5 | | | 3.87 | 2.49 | - | 6.14 | | 234 | |
| p4gm | 3.83 | 53.3 | 32.5 | | | 3.87 | 2.49 | - | 6.15 | | 234 | |
| **p3** | 0.584 | 28.0 | 9.52 | | | 0.688 | 12.2 | 30.7 | 1.25 | | 231 | |
| p3m1 | 2.05 | 37.0 | 14.3 | | | 2.10 | 6.06 | - | 2.53 | | 225 | |
| p31m | 4.83 | 37.0 | 35.0 | | | 4.88 | 1.50 | - | 10.2 | | 225 | |
| **p6** | 0.630 | 28.0 | 15.5 | | | 0.681 | 12.3 | 30.8 | 1.25 | | 231 | |
| p6mm | 2.07 | 37.0 | 23.7 | | | 2.10 | 6.06 | - | 2.53 | | 225 | |

Table A34: Results for **1024-pixel** diameter **circular** selection of image with **RGB noise level 1.00 and 10-pixel spread noise**

| Geometric Model | $J_{FC}$ | $F_{res}$ | $\phi_{res}$ (°) | CRISP's Suggestion | Estimated K-L-Best | G-AIC | G-AW full set (%) | G-AW p2, p3, p6 (%) | $E_{best, j}$ | $E_{p6, p3}$ | $N$ | $\hat{\epsilon}^2$ |
|---|---|---|---|---|---|---|---|---|---|---|---|---|
| **p2** | **0.0398** | **-** | **8.03** | | | **0.119** | **13.7** | **39.0** | **-** | | **231** | |
| p1m1 | 1.91 | 36.4 | 7.97 | | | 1.99 | 5.39 | - | 2.55 | | 231 | |
| p11m | 1.91 | 36.4 | 7.71 | | | 1.99 | 5.40 | - | 2.54 | | 231 | |
| p1g1 | 2.29 | 36.4 | 15.9 | | | 2.37 | 4.46 | - | 3.08 | | 231 | |
| p11g | 1.93 | 36.4 | 8.41 | | | 2.01 | 5.35 | - | 2.57 | | 231 | |
| p2mm | 1.94 | 36.4 | 17.4 | | | 1.98 | 5.42 | - | 2.54 | | 231 | |
| p2mg | 1.96 | 36.4 | 18.7 | | | 2.00 | 5.37 | - | 2.56 | | 231 | |
| p2gm | 1.96 | 36.4 | 18.7 | | | 2.00 | 5.37 | - | 2.56 | | 231 | |
| p2gg | 1.94 | 36.4 | 17.4 | p6 | p2 | 1.98 | 5.42 | - | 2.54 | 1.77 | 231 | 3.44E-04 |
| p4 | 3.20 | 37.2 | 27.2 | | | 3.24 | 2.89 | - | 4.76 | | 231 | |
| p4mm | 3.37 | 47.5 | 34.1 | | | 3.39 | 2.68 | - | 5.13 | | 228 | |
| p4gm | 3.37 | 47.5 | 33.7 | | | 3.38 | 2.69 | - | 5.12 | | 228 | |
| **p3** | **0.562** | **29.6** | **9.91** | | | **0.614** | **10.7** | **30.5** | **1.28** | | **226** | |
| p3m1 | 1.73 | 38.9 | 14.2 | | | 1.75 | 6.07 | - | 2.26 | | 222 | |
| p31m | 3.76 | 38.9 | 36.1 | | | 3.79 | 2.19 | - | 6.26 | | 222 | |
| **p6** | **0.585** | **29.6** | **18.6** | | | **0.611** | **10.7** | **30.5** | **1.28** | | **226** | |
| p6mm | 1.74 | 38.9 | 26.9 | | | 1.75 | 6.07 | - | 2.26 | | 222 | |

Table A35: Results for **2048-pixel square** selection of image with **RGB noise level 1.00 and 10-pixel spread noise**

| Geometric Model | $J_{FC}$ | $F_{res}$ | $\phi_{res}$ (°) | CRISP's Suggestion | Estimated K-L-Best | G-AIC | G-AW full set (%) | G-AW p2, p3, p6 (%) | $E_{best, j}$ | $E_{p6, p3}$ | $N$ | $\hat{\epsilon}^2$ |
|---|---|---|---|---|---|---|---|---|---|---|---|---|
| **p2** | **0.0240** | **-** | **4.32** | | | **0.0719** | **18.9** | **41.6** | **-** | | **211** | |
| p1m1 | 4.13 | 43.0 | 18.2 | | | 4.18 | 2.42 | - | 7.80 | | 211 | |
| p11m | 2.65 | 43.0 | 6.57 | | | 2.70 | 5.07 | - | 3.73 | | 211 | |
| p1g1 | 2.64 | 43.0 | 7.66 | | | 2.69 | 5.10 | - | 3.70 | | 211 | |
| p11g | 3.64 | 43.0 | 15.9 | | | 3.69 | 3.10 | - | 6.09 | | 211 | |
| p2mm | 2.77 | 43.0 | 14.3 | | | 2.80 | 4.83 | - | 3.90 | | 211 | |
| p2mg | 2.67 | 43.0 | 13.6 | | | 2.70 | 5.07 | - | 3.72 | | 211 | |
| p2gm | 2.68 | 43.0 | 13.9 | | | 2.70 | 5.07 | - | 3.72 | | 211 | |
| p2gg | 2.77 | 43.0 | 14.3 | p6 | p2 | 2.80 | 4.83 | - | 3.90 | 2.38 | 211 | 2.27E-04 |
| p4 | 3.81 | 46.6 | 20.7 | | | 3.83 | 2.88 | - | 6.56 | | 211 | |
| p4mm | 4.67 | 53.9 | 31.4 | | | 4.68 | 1.88 | - | 10.0 | | 206 | |
| p4gm | 4.67 | 53.9 | 31.4 | | | 4.68 | 1.88 | - | 10.0 | | 206 | |
| **p3** | **0.753** | **28.4** | **8.31** | | | **0.785** | **13.2** | **29.1** | **1.43** | | **206** | |
| p3m1 | 2.49 | 35.8 | 12.2 | | | 2.50 | 5.60 | - | 3.37 | | 204 | |
| p31m | 5.46 | 35.8 | 34.8 | | | 5.47 | 1.27 | - | 14.9 | | 204 | |
| **p6** | **0.755** | **28.4** | **11.7** | | | **0.771** | **13.3** | **29.3** | **1.42** | | **206** | |
| p6mm | 2.49 | 35.8 | 21.8 | | | 2.50 | 5.61 | - | 3.37 | | 204 | |

Table A36: Results for **2048-pixel** diameter **circular** selection of image with **RGB noise level 1.00 and 10-pixel spread noise**

| Geometric Model | $J_{FC}$ | $F_{res}$ | $\phi_{res}$ (°) | CRISP's Suggestion | Estimated K-L-Best | G-AIC | G-AW full set (%) | G-AW p2, p3, p6 (%) | $E_{best,\,j}$ | $E_{p6,\,p3}$ | $N$ | $\hat{\epsilon}^2$ |
|---|---|---|---|---|---|---|---|---|---|---|---|---|
| **p2** | **7.58E-03** | - | **3.37** | | | 0.0227 | 13.6 | 39.6 | - | | 207 | |
| p1m1 | 1.83 | 36.5 | 8.04 | | | 1.85 | 5.46 | - | 2.49 | | 207 | |
| p11m | 1.83 | 36.5 | 7.46 | | | 1.85 | 5.46 | - | 2.49 | | 207 | |
| p1g1 | 1.88 | 36.5 | 8.66 | | | 1.89 | 5.33 | - | 2.55 | | 207 | |
| p11g | 1.88 | 36.5 | 7.83 | | | 1.89 | 5.33 | - | 2.55 | | 207 | |
| p2mm | 1.84 | 36.5 | 14.3 | | | 1.84 | 5.46 | - | 2.49 | | 207 | |
| p2mg | 1.88 | 36.5 | 15.0 | | | 1.89 | 5.33 | - | 2.55 | | 207 | |
| p2gm | 1.88 | 36.5 | 15.0 | | | 1.89 | 5.33 | - | 2.55 | | 207 | |
| p2gg | 1.84 | 36.5 | 14.3 | p6 | p2 | 1.84 | 5.46 | - | 2.49 | 1.75 | 207 | 7.32E-05 |
| p4 | 3.04 | 38.3 | 23.8 | | | 3.05 | 2.99 | - | 4.54 | | 207 | |
| p4mm | 3.20 | 47.1 | 30.8 | | | 3.20 | 2.77 | - | 4.91 | | 204 | |
| p4gm | 3.20 | 47.1 | 31.0 | | | 3.21 | 2.77 | - | 4.91 | | 204 | |
| **p3** | **0.551** | **29.1** | **9.32** | | | 0.561 | 10.4 | 30.2 | 1.31 | | 202 | |
| p3m1 | 1.67 | 38.1 | 11.8 | | | 1.67 | 5.94 | - | 2.28 | | 200 | |
| p31m | 3.74 | 38.1 | 36.0 | | | 3.74 | 2.11 | - | 6.42 | | 200 | |
| **p6** | **0.556** | **29.1** | **16.0** | | | 0.561 | 10.4 | 30.2 | 1.31 | | 202 | |
| p6mm | 1.67 | 38.1 | 23.1 | | | 1.67 | 5.95 | - | 2.28 | | 200 | |

Table A37: Results for **1024-pixel square** selection of image **with RGB noise level 1.00 and 20-pixel spread noise**

| Geometric Model | $J_{FC}$ | $F_{res}$ | $\phi_{res}$ (°) | CRISP's Suggestion | Estimated K-L-Best | G-AIC | G-AW full set (%) | G-AW p2, p3, p6 (%) | $E_{best,\,j}$ | $E_{p6,\,p3}$ | $N$ | $\hat{\epsilon}^2$ |
|---|---|---|---|---|---|---|---|---|---|---|---|---|
| **p2** | **0.175** | - | **15.8** | | | 0.524 | 11.8 | 34.7 | - | | 214 | |
| p1m1 | 2.62 | 39.3 | 14.7 | | | 2.97 | 3.48 | - | 3.39 | | 214 | |
| p11m | 1.78 | 39.3 | 9.57 | | | 2.13 | 5.30 | - | 2.23 | | 214 | |
| p1g1 | 1.77 | 39.3 | 11.4 | | | 2.12 | 5.31 | - | 2.22 | | 214 | |
| p11g | 1.72 | 39.3 | 8.60 | | | 2.07 | 5.46 | - | 2.16 | | 214 | |
| p2mm | 1.84 | 39.3 | 21.7 | | | 2.02 | 5.60 | - | 2.11 | | 214 | |
| p2mg | 1.89 | 39.3 | 22.5 | | | 2.06 | 5.48 | - | 2.16 | | 214 | |
| p2gm | 1.89 | 39.3 | 22.6 | | | 2.06 | 5.47 | - | 2.16 | | 214 | |
| p2gg | 1.84 | 39.3 | 21.7 | p6 | p2 | 2.02 | 5.60 | - | 2.11 | 1.75 | 214 | 1.63E-03 |
| p4 | 2.68 | 44.3 | 25.2 | | | 2.86 | 3.68 | - | 3.21 | | 214 | |
| p4mm | 3.07 | 52.0 | 33.1 | | | 3.15 | 3.17 | - | 3.72 | | 208 | |
| p4gm | 3.07 | 52.0 | 33.0 | | | 3.15 | 3.17 | - | 3.72 | | 208 | |
| **p3** | **0.429** | **29.4** | **9.53** | | | 0.660 | 11.0 | 32.4 | 1.07 | | 212 | |
| p3m1 | 1.64 | 37.1 | 15.6 | | | 1.75 | 6.40 | - | 1.85 | | 207 | |
| p31m | 4.61 | 37.1 | 36.4 | | | 4.73 | 1.44 | - | 8.18 | | 207 | |
| **p6** | **0.511** | **29.4** | **20.0** | | | 0.627 | 11.2 | 32.9 | 1.05 | | 212 | |
| p6mm | 1.70 | 37.1 | 28.7 | | | 1.76 | 6.37 | - | 1.85 | | 207 | |

Table A38: Results for **1024-pixel** diameter **circular** selection of image with **RGB noise level 1.00 and 20-pixel spread noise**

| Geometric Model | $J_{FC}$ | $F_{res}$ | $\phi_{res}$ (°) | CRISP's Suggestion | Estimated K-L-Best | G-AIC | G-AW full set (%) | G-AW p2, p3, p6 (%) | $E_{best, j}$ | $E_{p6, p3}$ | $N$ | $\hat{\epsilon}^2$ |
|---|---|---|---|---|---|---|---|---|---|---|---|---|
| **p2** | **0.0852** | **-** | **14.9** | | | **0.256** | **11.5** | **34.4** | **-** | | **216** | |
| p1m1 | 1.66 | 36.2 | 7.97 | | | 1.83 | 5.22 | - | 2.19 | | 215 | |
| p11m | 1.65 | 36.2 | 8.51 | | | 1.82 | 5.24 | - | 2.19 | | 215 | |
| p1g1 | 1.66 | 36.2 | 7.33 | | | 1.83 | 5.23 | - | 2.19 | | 215 | |
| p11g | 1.68 | 36.2 | 9.14 | | | 1.85 | 5.17 | - | 2.22 | | 215 | |
| p2mm | 1.71 | 36.2 | 23.0 | | | 1.80 | 5.30 | - | 2.16 | | 215 | |
| p2mg | 1.74 | 36.2 | 24.1 | | | 1.83 | 5.23 | - | 2.19 | | 215 | |
| p2gm | 1.74 | 36.2 | 24.3 | | | 1.83 | 5.22 | - | 2.19 | | 215 | |
| p2gg | 1.71 | 36.2 | 23.0 | p6 | p2 | 1.80 | 5.30 | - | 2.16 | 1.70 | 215 | 7.89E-04 |
| p4 | 2.66 | 37.2 | 30.6 | | | 2.75 | 3.30 | - | 3.48 | | 216 | |
| p4mm | 2.85 | 48.5 | 35.1 | | | 2.89 | 3.07 | - | 3.74 | | 211 | |
| p4gm | 2.85 | 48.5 | 35.5 | | | 2.89 | 3.06 | - | 3.74 | | 211 | |
| **p3** | **0.254** | **27.7** | **9.55** | | | **0.366** | **10.8** | **32.6** | **1.06** | | **212** | |
| p3m1 | 1.34 | 40.0 | 15.7 | | | 1.40 | 6.47 | - | 1.77 | | 214 | |
| p31m | 3.31 | 40.0 | 35.7 | | | 3.37 | 2.42 | - | 4.73 | | 214 | |
| **p6** | **0.286** | **27.7** | **18.2** | | | **0.342** | **11.0** | **33.0** | **1.04** | | **212** | |
| p6mm | 1.37 | 40.0 | 28.9 | | | 1.40 | 6.48 | - | 1.77 | | 214 | |

Table A39: Results for **2048-pixel square** selection of image with **RGB noise level 1.00 and 20-pixel spread noise**

| Geometric Model | $J_{FC}$ | $F_{res}$ | $\phi_{res}$ (°) | CRISP's Suggestion | Estimated K-L-Best | G-AIC | G-AW full set (%) | G-AW p2, p3, p6 (%) | $E_{best, j}$ | $E_{p6, p3}$ | $N$ | $\hat{\epsilon}^2$ |
|---|---|---|---|---|---|---|---|---|---|---|---|---|
| **p2** | **0.0332** | **-** | **6.35** | | | **0.0996** | **14.9** | **36.8** | **-** | | **153** | |
| p1m1 | 2.89 | 43.1 | 16.1 | | | 2.95 | 3.57 | - | 4.16 | | 153 | |
| p11m | 2.17 | 43.1 | 7.45 | | | 2.24 | 5.11 | - | 2.91 | | 153 | |
| p1g1 | 2.17 | 43.1 | 6.88 | | | 2.23 | 5.12 | - | 2.91 | | 153 | |
| p11g | 3.20 | 43.1 | 16.5 | | | 3.27 | 3.05 | - | 4.88 | | 153 | |
| p2mm | 2.23 | 43.1 | 14.0 | | | 2.26 | 5.05 | - | 2.95 | | 153 | |
| p2mg | 2.19 | 43.1 | 14.5 | | | 2.23 | 5.13 | - | 2.90 | | 153 | |
| p2gm | 2.19 | 43.1 | 14.5 | | | 2.23 | 5.13 | - | 2.90 | | 153 | |
| p2gg | 2.23 | 43.1 | 13.9 | p6 | p2 | 2.26 | 5.05 | - | 2.95 | 2.04 | 153 | 4.34E-04 |
| p4 | 2.96 | 45.6 | 21.0 | | | 2.99 | 3.50 | - | 4.25 | | 153 | |
| p4mm | 3.66 | 50.8 | 29.7 | | | 3.67 | 2.49 | - | 5.97 | | 143 | |
| p4gm | 3.66 | 50.8 | 29.6 | | | 3.67 | 2.49 | - | 5.97 | | 143 | |
| **p3** | **0.366** | **23.0** | **6.02** | | | **0.406** | **12.8** | **31.6** | **1.17** | | **137** | |
| p3m1 | 1.81 | 29.8 | 10.2 | | | 1.83 | 6.27 | - | 2.37 | | 137 | |
| p31m | 4.90 | 29.8 | 35.4 | | | 4.92 | 1.33 | - | 11.1 | | 137 | |
| **p6** | **0.379** | **23.0** | **9.96** | | | **0.399** | **12.8** | **31.7** | **1.16** | | **137** | |
| p6mm | 1.81 | 29.8 | 19.5 | | | 1.82 | 6.28 | - | 2.37 | | 137 | |

Table A40: Results for **2048-pixel** diameter **circular** selection of image with **RGB noise level 1.00 and 20-pixel spread noise**

| Geometric Model | $J_{FC}$ | $F_{res}$ | $\phi_{res}$ (°) | CRISP's Suggestion | Estimated K-L-Best | G-AIC | G-AW full set (%) | G-AW p2, p3, p6 (%) | $E_{best,j}$ | $E_{p6,p3}$ | N | $\hat{\epsilon}^2$ |
|---|---|---|---|---|---|---|---|---|---|---|---|---|
| **p2** | **0.0226** | - | **5.99** | | | **0.0677** | **13.8** | **36.4** | - | | **153** | |
| p1m1 | 2.73 | 44.4 | 15.7 | | | 2.77 | 3.56 | - | 3.87 | | 153 | |
| p11m | 1.94 | 44.4 | 6.92 | | | 1.99 | 5.28 | - | 2.61 | | 153 | |
| p1g1 | 1.93 | 44.4 | 6.91 | | | 1.98 | 5.31 | - | 2.60 | | 153 | |
| p11g | 2.79 | 44.4 | 15.7 | | | 2.83 | 3.46 | - | 3.98 | | 153 | |
| p2mm | 2.01 | 44.4 | 13.2 | | | 2.04 | 5.15 | - | 2.68 | | 153 | |
| p2mg | 1.96 | 44.4 | 14.3 | | | 1.98 | 5.29 | - | 2.60 | | 153 | |
| p2gm | 1.96 | 44.4 | 14.3 | | | 1.98 | 5.29 | - | 2.60 | | 153 | |
| p2gg | 2.01 | 44.4 | 13.2 | p6 | p2 | 2.04 | 5.15 | - | 2.68 | 1.89 | 153 | 2.95E-04 |
| p4 | 2.66 | 47.6 | 20.2 | | | 2.68 | 3.73 | - | 3.69 | | 153 | |
| p4mm | 3.26 | 53.0 | 28.4 | | | 3.28 | 2.77 | - | 4.97 | | 142 | |
| p4gm | 3.26 | 53.0 | 28.4 | | | 3.28 | 2.77 | - | 4.97 | | 142 | |
| **p3** | **0.314** | **24.1** | **5.18** | | | **0.341** | **12.0** | **31.8** | **1.15** | | **139** | |
| p3m1 | 1.60 | 31.0 | 9.32 | | | 1.61 | 6.37 | - | 2.16 | | 137 | |
| p31m | 4.32 | 31.0 | 34.5 | | | 4.33 | 1.63 | - | 8.43 | | 137 | |
| **p6** | **0.323** | **24.1** | **9.32** | | | **0.337** | **12.0** | **31.8** | **1.14** | | **139** | |
| p6mm | 1.60 | 31.0 | 18.9 | | | 1.61 | 6.37 | - | 2.16 | | 137 | |

Table A41: Results for **1024-pixel square** selection of image with **RGB noise level 1.00 and 30-pixel spread noise**

| Geometric Model | $J_{FC}$ | $F_{res}$ | $\phi_{res}$ (°) | CRISP's Suggestion | Estimated K-L-Best | G-AIC | G-AW full set (%) | G-AW p2, p3, p6 (%) | $E_{best,j}$ | $E_{p6,p3}$ | N | $\hat{\epsilon}^2$ |
|---|---|---|---|---|---|---|---|---|---|---|---|---|
| **p2** | **0.144** | - | **17.8** | | | **0.432** | **11.2** | **33.5** | - | | **203** | |
| p1m1 | 1.54 | 38.7 | 8.72 | | | 1.83 | 5.57 | - | 2.01 | | 203 | |
| p11m | 1.48 | 38.7 | 7.85 | | | 1.77 | 5.73 | - | 1.95 | | 203 | |
| p1g1 | 1.52 | 38.7 | 9.79 | | | 1.80 | 5.63 | - | 1.99 | | 203 | |
| p11g | 1.55 | 38.7 | 7.65 | | | 1.83 | 5.55 | - | 2.02 | | 203 | |
| p2mm | 1.65 | 38.7 | 22.5 | | | 1.79 | 5.67 | - | 1.97 | | 203 | |
| p2mg | 1.60 | 38.7 | 23.2 | | | 1.75 | 5.79 | - | 1.93 | | 203 | |
| p2gm | 1.60 | 38.7 | 23.2 | | | 1.75 | 5.79 | - | 1.93 | | 203 | |
| p2gg | 1.65 | 38.7 | 22.5 | p3 | p2 | 1.79 | 5.67 | - | 1.97 | 1.67 | 203 | 1.42E-03 |
| p4 | 2.09 | 43.1 | 25.7 | | | 2.24 | 4.53 | - | 2.47 | | 203 | |
| p4mm | 2.57 | 52.4 | 32.2 | | | 2.65 | 3.70 | - | 3.02 | | 200 | |
| p4gm | 2.57 | 52.4 | 31.9 | | | 2.64 | 3.70 | - | 3.02 | | 200 | |
| **p3** | **0.278** | **28.4** | **9.46** | | | **0.465** | **11.0** | **33.0** | **1.02** | | **198** | |
| p3m1 | 1.36 | 37.4 | 16.6 | | | 1.45 | 6.71 | - | 1.67 | | 196 | |
| p31m | 4.17 | 37.4 | 37.9 | | | 4.26 | 1.65 | - | 6.80 | | 196 | |
| **p6** | **0.340** | **28.4** | **21.0** | | | **0.434** | **11.2** | **33.5** | **1.00** | | **198** | |
| p6mm | 5.34 | 37.4 | 53.6 | | | 5.38 | 0.940 | - | 11.9 | | 196 | |

Table A42: Results for **1024-pixel** diameter **circular** selection of image **with RGB noise level 1.00 and 30-pixel spread noise**

| Geometric Model | $J_{FC}$ | $F_{res}$ | $\phi_{res}$ (°) | CRISP's Suggestion | Estimated K-L-Best | G-AIC | G-AW full set (%) | G-AW p2, p3, p6 (%) | $E_{best, j}$ | $E_{p6, p3}$ | $N$ | $\hat{\epsilon}^2$ |
|---|---|---|---|---|---|---|---|---|---|---|---|---|
| **p2** | **0.113** | **-** | **15.4** | | | **0.338** | **11.8** | **34.1** | **-** | | **211** | |
| p1m1 | 1.61 | 40.9 | 7.83 | | | 1.83 | 5.58 | - | 2.11 | | 211 | |
| p11m | 1.65 | 40.9 | 8.99 | | | 1.88 | 5.46 | - | 2.16 | | 211 | |
| p1g1 | 1.64 | 40.9 | 9.28 | | | 1.87 | 5.48 | - | 2.15 | | 211 | |
| p11g | 1.61 | 40.9 | 6.20 | | | 1.83 | 5.58 | - | 2.11 | | 211 | |
| p2mm | 1.68 | 40.9 | 20.3 | | | 1.80 | 5.69 | - | 2.07 | | 211 | |
| p2mg | 1.73 | 40.9 | 22.5 | | | 1.84 | 5.56 | - | 2.12 | | 211 | |
| p2gm | 1.73 | 40.9 | 22.5 | | | 1.84 | 5.56 | - | 2.12 | | 211 | |
| p2gg | 1.68 | 40.9 | 20.4 | p3 | p2 | 1.80 | 5.68 | - | 2.07 | 1.74 | 211 | 1.07E-03 |
| p4 | 2.17 | 43.9 | 24.7 | | | 2.28 | 4.46 | - | 2.64 | | 211 | |
| p4mm | 2.70 | 53.2 | 30.9 | | | 2.75 | 3.53 | - | 3.34 | | 207 | |
| p4gm | 2.70 | 53.2 | 31.0 | | | 2.75 | 3.52 | - | 3.34 | | 207 | |
| **p3** | **0.276** | **28.2** | **9.07** | | | **0.423** | **11.3** | **32.7** | **1.04** | | **206** | |
| p3m1 | 1.43 | 38.1 | 15.8 | | | 1.50 | 6.58 | - | 1.79 | | 207 | |
| p31m | 4.03 | 38.1 | 35.6 | | | 4.10 | 1.79 | - | 6.57 | | 207 | |
| **p6** | **0.323** | **28.2** | **18.5** | | | **0.396** | **11.4** | **33.2** | **1.03** | | **206** | |
| p6mm | 5.26 | 38.1 | 50.4 | | | 5.29 | 0.989 | - | 11.9 | | 207 | |

Table A43: Results for **2048-pixel square** selection of image with **RGB noise level 1.00 and 30-pixel spread noise**

| Geometric Model | $J_{FC}$ | $F_{res}$ | $\phi_{res}$ (°) | CRISP's Suggestion | Estimated K-L-Best | G-AIC | G-AW full set (%) | G-AW p2, p3, p6 (%) | $E_{best, j}$ | $E_{p6, p3}$ | $N$ | $\hat{\epsilon}^2$ |
|---|---|---|---|---|---|---|---|---|---|---|---|---|
| **p2** | **0.0287** | **-** | **6.28** | | | **0.0862** | **13.1** | **35.4** | **-** | | **130** | |
| p1m1 | 3.22 | 41.9 | 18.1 | | | 3.28 | 2.65 | - | 4.94 | | 130 | |
| p11m | 1.85 | 41.9 | 6.44 | | | 1.91 | 5.26 | - | 2.49 | | 130 | |
| p1g1 | 1.84 | 41.9 | 6.24 | | | 1.89 | 5.31 | - | 2.47 | | 130 | |
| p11g | 1.90 | 41.9 | 8.17 | | | 1.96 | 5.13 | - | 2.55 | | 130 | |
| p2mm | 1.89 | 41.9 | 13.0 | | | 1.92 | 5.24 | - | 2.50 | | 130 | |
| p2mg | 1.87 | 41.9 | 13.9 | | | 1.90 | 5.30 | - | 2.47 | | 130 | |
| p2gm | 1.87 | 41.9 | 13.9 | | | 1.90 | 5.30 | - | 2.47 | | 130 | |
| p2gg | 1.89 | 41.9 | 13.0 | p6 | p2 | 1.92 | 5.24 | - | 2.50 | 1.91 | 130 | 4.42E-04 |
| p4 | 2.45 | 45.8 | 17.5 | | | 2.48 | 3.96 | - | 3.31 | | 130 | |
| p4mm | 3.13 | 49.2 | 26.6 | | | 3.14 | 2.85 | - | 4.60 | | 119 | |
| p4gm | 3.13 | 49.2 | 26.7 | | | 3.14 | 2.85 | - | 4.60 | | 119 | |
| **p3** | **0.236** | **20.8** | **5.09** | | | **0.272** | **11.9** | **32.2** | **1.10** | | **121** | |
| p3m1 | 1.54 | 26.6 | 9.15 | | | 1.56 | 6.28 | - | 2.09 | | 113 | |
| p31m | 4.71 | 26.6 | 34.7 | | | 4.73 | 1.29 | - | 10.2 | | 113 | |
| **p6** | **0.244** | **20.8** | **8.28** | | | **0.261** | **12.0** | **32.4** | **1.09** | | **121** | |
| p6mm | 1.55 | 26.6 | 16.0 | | | 1.55 | 6.29 | - | 2.08 | | 113 | |

Table A44: Results for **2048-pixel** diameter **circular** selection of image **with RGB noise level 1.00 and 30-pixel spread noise**

| Geometric Model | $J_{FC}$ | $F_{res}$ | $\phi_{res}$ (°) | CRISP's Suggestion | Estimated K-L-Best | G-AIC | G-AW full set (%) | G-AW p2, p3, p6 (%) | $E_{best, j}$ | $E_{p6, p3}$ | $N$ | $\hat{\epsilon}^2$ |
|---|---|---|---|---|---|---|---|---|---|---|---|---|
| **p2** | **0.0127** | - | **5.69** | | | **0.0381** | **11.0** | **34.8** | - | | **131** | |
| p1m1 | 1.51 | 40.5 | 6.47 | | | 1.54 | 5.19 | - | 2.12 | | 131 | |
| p11m | 1.51 | 40.5 | 4.42 | | | 1.54 | 5.20 | - | 2.12 | | 131 | |
| p1g1 | 1.51 | 40.5 | 3.91 | | | 1.54 | 5.19 | - | 2.12 | | 131 | |
| p11g | 1.38 | 40.5 | 6.52 | | | 1.41 | 5.55 | - | 1.98 | | 131 | |
| p2mm | 1.52 | 40.5 | 13.0 | | | 1.53 | 5.20 | - | 2.11 | | 131 | |
| p2mg | 1.39 | 40.5 | 14.3 | | | 1.40 | 5.55 | - | 1.98 | | 131 | |
| p2gm | 1.39 | 40.5 | 14.3 | | | 1.40 | 5.55 | - | 1.98 | | 131 | |
| p2gg | 1.52 | 40.5 | 13.0 | p3 | p2 | 1.53 | 5.20 | - | 2.11 | 1.57 | 131 | 1.94E-04 |
| p4 | 2.22 | 41.3 | 21.7 | | | 2.23 | 3.68 | - | 2.99 | | 131 | |
| p4mm | 2.42 | 50.7 | 24.9 | | | 2.43 | 3.33 | - | 3.30 | | 114 | |
| p4gm | 2.42 | 50.7 | 24.9 | | | 2.43 | 3.33 | - | 3.30 | | 114 | |
| **p3** | **0.156** | **22.8** | **4.00** | | | **0.171** | **10.3** | **32.6** | **1.07** | | **111** | |
| p3m1 | 1.07 | 33.6 | 8.37 | | | 1.07 | 6.55 | - | 1.68 | | 105 | |
| p31m | 3.15 | 33.6 | 35.5 | | | 3.15 | 2.32 | - | 4.75 | | 105 | |
| **p6** | **0.161** | **22.8** | **7.02** | | | **0.168** | **10.3** | **32.6** | **1.07** | | **111** | |
| p6mm | 1.07 | 33.6 | 16.3 | | | 1.07 | 6.55 | - | 1.68 | | 105 | |

Table A45: Results for **1024-pixel square** selection of image with **RGB noise level 1.00 and 40-pixel spread noise**

| Geometric Model | $J_{FC}$ | $F_{res}$ | $\phi_{res}$ (°) | CRISP's Suggestion | Estimated K-L-Best | G-AIC | G-AW full set (%) | G-AW p2, p3, p6 (%) | $E_{best, j}$ | $E_{p6, p3}$ | $N$ | $\hat{\epsilon}^2$ |
|---|---|---|---|---|---|---|---|---|---|---|---|---|
| **p2** | **0.116** | - | **18.8** | | | **0.347** | **10.8** | **33.5** | - | | **219** | |
| p1m1 | 1.25 | 34.8 | 7.65 | | | 1.48 | 6.12 | - | 1.76 | | 219 | |
| p11m | 1.49 | 34.8 | 8.30 | | | 1.72 | 5.44 | - | 1.99 | | 219 | |
| p1g1 | 1.73 | 34.8 | 11.1 | | | 1.96 | 4.82 | - | 2.24 | | 219 | |
| p11g | 1.28 | 34.8 | 7.86 | | | 1.51 | 6.04 | - | 1.79 | | 219 | |
| p2mm | 1.59 | 34.8 | 25.1 | | | 1.71 | 5.47 | - | 1.98 | | 219 | |
| p2mg | 1.34 | 34.8 | 25.0 | | | 1.45 | 6.22 | - | 1.74 | | 219 | |
| p2gm | 1.34 | 34.8 | 25.2 | | | 1.45 | 6.21 | - | 1.74 | | 219 | |
| p2gg | 1.59 | 34.8 | 25.1 | p3 | p2 | 1.71 | 5.47 | - | 1.98 | 1.50 | 219 | 1.06E-03 |
| p4 | 2.18 | 35.8 | 30.5 | | | 2.29 | 4.09 | - | 2.64 | | 219 | |
| p4mm | 2.44 | 48.7 | 35.6 | | | 2.50 | 3.68 | - | 2.93 | | 213 | |
| p4gm | 2.45 | 48.7 | 35.8 | | | 2.50 | 3.68 | - | 2.94 | | 213 | |
| **p3** | **0.220** | **27.9** | **13.0** | | | **0.373** | **10.7** | **33.1** | **1.01** | | **217** | |
| p3m1 | 1.10 | 39.8 | 19.1 | | | 1.18 | 7.14 | - | 1.51 | | 215 | |
| p31m | 3.57 | 39.8 | 39.1 | | | 3.64 | 2.08 | - | 5.20 | | 215 | |
| **p6** | **0.283** | **27.9** | **24.7** | | | **0.359** | **10.7** | **33.3** | **1.01** | | **217** | |
| p6mm | 4.47 | 39.8 | 56.6 | | | 4.50 | 1.35 | - | 7.99 | | 215 | |

Table A46: Results for **1024-pixel** diameter **circular** selection of image with **RGB noise level 1.00 and 40-pixel spread noise**

| Geometric Model | $J_{FC}$ | $F_{res}$ | $\phi_{res}$ (°) | CRISP's Suggestion | Estimated K-L-Best | G-AIC | G-AW full set (%) | G-AW p2, p3, p6 (%) | $E_{best,\,j}$ | $E_{p6,\,p3}$ | $N$ | $\hat{\epsilon}^2$ |
|---|---|---|---|---|---|---|---|---|---|---|---|---|
| **p2**   | **0.107** | -    | **17.8** |    |    | 0.322 | 10.2 | 33.5 | -    |      | **202** |          |
| p1m1     | 1.48      | 36.2 | 7.79     |    |    | 1.70  | 5.11 | -    | 1.99 |      | 202     |          |
| p11m     | 1.48      | 36.2 | 6.84     |    |    | 1.70  | 5.11 | -    | 1.99 |      | 202     |          |
| p1g1     | 1.30      | 36.2 | 7.44     |    |    | 1.52  | 5.60 | -    | 1.82 |      | 202     |          |
| p11g     | 1.30      | 36.2 | 6.81     |    |    | 1.51  | 5.61 | -    | 1.81 |      | 202     |          |
| p2mm     | 1.56      | 36.2 | 23.2     |    |    | 1.67  | 5.19 | -    | 1.96 |      | 202     |          |
| p2mg     | 1.38      | 36.2 | 23.1     |    |    | 1.49  | 5.67 | -    | 1.79 |      | 202     |          |
| p2gm     | 1.38      | 36.2 | 23.2     |    |    | 1.49  | 5.67 | -    | 1.79 |      | 202     |          |
| p2gg     | 1.56      | 36.2 | 23.1     | p3 | p2 | 1.67  | 5.19 | -    | 1.96 | 1.53 | 202     | 1.06E-03 |
| p4       | 2.13      | 38.9 | 27.8     |    |    | 2.23  | 3.91 | -    | 2.60 |      | 202     |          |
| p4mm     | 2.39      | 52.4 | 31.6     |    |    | 2.45  | 3.52 | -    | 2.89 |      | 199     |          |
| p4gm     | 2.39      | 52.4 | 31.6     |    |    | 2.45  | 3.52 | -    | 2.89 |      | 199     |          |
| **p3**   | **0.213** | 28.5 | **10.4** |    |    | 0.352 | 10.0 | 33.0 | 1.02 |      | **196** |          |
| p3m1     | 1.11      | 40.6 | 16.9     |    |    | 1.18  | 6.63 | -    | 1.53 |      | 197     |          |
| p31m     | 3.27      | 40.6 | 36.6     |    |    | 3.34  | 2.25 | -    | 4.52 |      | 197     |          |
| **p6**   | **0.254** | 28.5 | **23.0** |    |    | 0.323 | 10.2 | 33.5 | 1.00 |      | **196** |          |
| p6mm     | 1.13      | 40.6 | 30.2     |    |    | 1.16  | 6.67 | -    | 1.52 |      | 197     |          |

Table A47: Results for **2048-pixel square** selection of image with **RGB noise level 1.00 and 40-pixel spread noise**

| Geometric Model | $J_{FC}$ | $F_{res}$ | $\phi_{res}$ (°) | CRISP's Suggestion | Estimated K-L-Best | G-AIC | G-AW full set (%) | G-AW p2, p3, p6 (%) | $E_{best,\,j}$ | $E_{p6,\,p3}$ | $N$ | $\hat{\epsilon}^2$ |
|---|---|---|---|---|---|---|---|---|---|---|---|---|
| **p2**   | **0.0256** | -    | **9.53** |    |    | 0.0768 | 10.4 | 34.2 | -    |      | **149** |          |
| p1m1     | 1.20       | 40.8 | 5.50     |    |    | 1.26   | 5.76 | -    | 1.80 |      | 148     |          |
| p11m     | 1.51       | 40.8 | 4.59     |    |    | 1.56   | 4.95 | -    | 2.10 |      | 148     |          |
| p1g1     | 1.51       | 40.8 | 4.09     |    |    | 1.56   | 4.94 | -    | 2.10 |      | 148     |          |
| p11g     | 1.21       | 40.8 | 6.10     |    |    | 1.26   | 5.76 | -    | 1.80 |      | 148     |          |
| p2mm     | 1.53       | 40.8 | 16.1     |    |    | 1.55   | 4.96 | -    | 2.09 |      | 148     |          |
| p2mg     | 1.22       | 40.8 | 16.2     |    |    | 1.25   | 5.78 | -    | 1.80 |      | 148     |          |
| p2gm     | 1.22       | 40.8 | 16.2     |    |    | 1.25   | 5.78 | -    | 1.80 |      | 148     |          |
| p2gg     | 1.53       | 40.8 | 15.9     | p3 | p2 | 1.55   | 4.96 | -    | 2.09 | 1.47 | 148     | 3.44E-04 |
| p4       | 2.02       | 42.1 | 22.6     |    |    | 2.05   | 3.88 | -    | 2.68 |      | 148     |          |
| p4mm     | 2.26       | 51.1 | 26.2     |    |    | 2.27   | 3.47 | -    | 2.99 |      | 135     |          |
| p4gm     | 2.26       | 51.1 | 26.3     |    |    | 2.27   | 3.47 | -    | 2.99 |      | 135     |          |
| **p3**   | **0.130**  | 22.6 | **5.11** |    |    | 0.161  | 9.96 | 32.8 | 1.04 |      | **136** |          |
| p3m1     | 0.907      | 33.6 | 10.6     |    |    | 0.922  | 6.81 | -    | 1.53 |      | 137     |          |
| p31m     | 3.04       | 33.6 | 36.5     |    |    | 3.05   | 2.35 | -    | 4.43 |      | 137     |          |
| **p6**   | **0.141**  | 22.6 | **10.9** |    |    | 0.156  | 9.98 | 32.9 | 1.04 |      | **136** |          |
| p6mm     | 0.914      | 33.6 | 19.8     |    |    | 0.922  | 6.81 | -    | 1.53 |      | 137     |          |

Table A48: Results for **2048-pixel** diameter **circular** selection of image with **RGB noise level 1.00 and 40-pixel spread noise**

| Geometric Model | $J_{FC}$ | $F_{res}$ | $\phi_{res}$ (°) | CRISP's Suggestion | Estimated K-L-Best | G-AIC | G-AW full set (%) | G-AW p2, p3, p6 (%) | $E_{best, j}$ | $E_{p6, p3}$ | N | $\hat{\epsilon}^2$ |
|---|---|---|---|---|---|---|---|---|---|---|---|---|
| **p2** | **0.0225** | - | **9.15** | | | **0.0674** | **10.5** | **34.3** | - | | **144** | |
| p1m1 | 1.25 | 40.8 | 5.00 | | | 1.29 | 5.72 | - | 1.84 | | 144 | |
| p11m | 1.53 | 40.8 | 4.01 | | | 1.58 | 4.96 | - | 2.13 | | 144 | |
| p1g1 | 1.53 | 40.8 | 3.64 | | | 1.58 | 4.96 | - | 2.13 | | 144 | |
| p11g | 1.25 | 40.8 | 5.72 | | | 1.29 | 5.71 | - | 1.85 | | 144 | |
| p2mm | 1.55 | 40.8 | 15.9 | | | 1.57 | 4.97 | - | 2.12 | | 144 | |
| p2mg | 1.26 | 40.8 | 16.1 | | | 1.29 | 5.74 | - | 1.84 | | 144 | |
| p2gm | 1.26 | 40.8 | 16.0 | | | 1.29 | 5.74 | - | 1.84 | | 144 | |
| p2gg | 1.55 | 40.8 | 15.6 | p3 | p2 | 1.57 | 4.98 | - | 2.12 | 1.49 | 144 | 3.12E-04 |
| p4 | 2.08 | 42.7 | 22.8 | | | 2.10 | 3.82 | - | 2.76 | | 144 | |
| p4mm | 2.31 | 51.8 | 26.2 | | | 2.32 | 3.42 | - | 3.08 | | 132 | |
| p4gm | 2.31 | 51.8 | 26.2 | | | 2.32 | 3.42 | - | 3.08 | | 132 | |
| **p3** | **0.129** | **22.9** | **4.81** | | | **0.156** | **10.1** | **32.8** | **1.05** | | **128** | |
| p3m1 | 0.939 | 34.4 | 10.1 | | | 0.952 | 6.78 | - | 1.56 | | 127 | |
| p31m | 3.14 | 34.4 | 37.2 | | | 3.15 | 2.26 | - | 4.68 | | 127 | |
| **p6** | **0.137** | **22.9** | **9.66** | | | **0.151** | **10.1** | **32.9** | **1.04** | | **128** | |
| p6mm | 0.946 | 34.4 | 19.2 | | | 0.952 | 6.78 | - | 1.56 | | 127 | |